\documentclass[11pt]{article}

\usepackage[final]{acl}

\usepackage{times}
\usepackage{latexsym}

\usepackage[T1]{fontenc}

\usepackage[utf8]{inputenc}

\usepackage{microtype}

\usepackage{graphicx}
\usepackage{amsmath}
\usepackage{amssymb}
\usepackage{booktabs}
\usepackage{multirow}
\usepackage{adjustbox}

\usepackage{subfigure}
\usepackage[most]{tcolorbox}
\usepackage{xcolor}
\usepackage{enumitem}
\usepackage[table]{xcolor}
\usepackage{array}

\tcbuselibrary{listings, breakable, skins}

\definecolor{PromptBlue}{HTML}{EAF3FF}
\definecolor{PromptBlueFrame}{HTML}{3B82F6}
\definecolor{PromptGreen}{HTML}{ECFDF5}
\definecolor{PromptGreenFrame}{HTML}{10B981}
\definecolor{PromptPurple}{HTML}{F5F3FF}
\definecolor{PromptPurpleFrame}{HTML}{8B5CF6}
\definecolor{PromptOrange}{HTML}{FFF7ED}
\definecolor{PromptOrangeFrame}{HTML}{F97316}
\definecolor{PromptGray}{HTML}{F8FAFC}
\definecolor{PromptGrayFrame}{HTML}{64748B}

\newtcblisting{promptbox}[2][]{
  enhanced,
  breakable,
  colback=#2!7,
  colframe=#2!80!black,
  coltitle=white,
  fonttitle=\bfseries,
  title=#1,
  boxrule=0.8pt,
  arc=2mm,
  left=2mm,
  right=2mm,
  top=1mm,
  bottom=1mm,
  listing only,
  listing options={
    basicstyle=\ttfamily\scriptsize,
    breaklines=true,
    breakatwhitespace=true,
    columns=fullflexible,
    keepspaces=true,
    showstringspaces=false,
    aboveskip=0pt,
    belowskip=0pt
  }
}

\newtcolorbox{formatbox}[2][]{
  enhanced,
  breakable,
  colback=#2!7,
  colframe=#2!80!black,
  coltitle=white,
  fonttitle=\bfseries,
  title=#1,
  boxrule=0.8pt,
  arc=2mm,
  left=2mm,
  right=2mm,
  top=1mm,
  bottom=1mm
}
\usepackage{fontawesome5}

\title{Human Label Variation as Stable Signal: Learning Annotator-Specific Explanation Behavior via Cross-Annotator Preference Optimization}

\author{
 \textbf{Beiduo Chen\textsuperscript{\faMountain\kern1pt\faRobot}} \quad
 \textbf{Pingjun Hong\textsuperscript{\faLaptopCode\kern1pt\faGraduationCap}} \quad
 \textbf{Ziyun Zhang\textsuperscript{\faMountain}} \quad
 \\
 \textbf{Benjamin Roth\textsuperscript{\faLaptopCode}} \quad
 \textbf{Anna Korhonen\textsuperscript{\faSchool}} \quad
 \textbf{Barbara Plank\textsuperscript{\faMountain\kern1pt\faRobot}}
\\
\textsuperscript{\faMountain}MaiNLP, Center for Information and Language Processing, LMU Munich, Germany \\
\textsuperscript{\faRobot}Munich Center for Machine Learning
\textsuperscript{\faSchool}LTL, University of Cambridge, United Kingdom \\
\textsuperscript{\faLaptopCode}Faculty and \textsuperscript{\faGraduationCap}UniVie Doctoral School of Computer Science, University of Vienna, Austria \\
\footnotesize{
\tt 
\href{mailto:beiduo.chen@lmu.de}{\textcolor{black}{beiduo.chen@lmu.de}},
\href{mailto:pingjun.hong@univie.ac.at}{\textcolor{black}{pingjun.hong@univie.ac.at}},
\href{mailto:alk23@cam.ac.uk}{\textcolor{black}{alk23@cam.ac.uk}},
\href{mailto:b.plank@lmu.de}{\textcolor{black}{b.plank@lmu.de}}}}

\begin{document}
\maketitle

\begin{abstract}
Free-text explanations extend human label variation (HLV) beyond label disagreement by revealing the reasoning and preferences behind annotators' decisions. 
We study whether large language models (LLMs) can learn and reproduce such annotator-specific label-explanation behavior.
Using two sentence-pair tasks with four annotators each---natural language inference and paraphrase judgment---we first analyze whether annotators exhibit stable individual patterns.
We find that such patterns are weak at the single-annotation level due to strong input-content effects, but become detectable after input-content reduction and annotator-level aggregation.
We then compare prompting and supervised fine-tuning (SFT) baselines and propose cross-annotator preference optimization (CAPO), which contrasts a target annotator's response with other valid but less target-specific annotations for the same input. 
Experiments show that prompting is limited and unstable, SFT better captures annotator-specific behavior, and CAPO further improves aggregation-aware imitation and judge-based attribution while preserving target-specific reasoning patterns under human validation. 
Overall, our results show that HLV can be learned as annotator-specific label-explanation behavior, suggesting a path toward scalable explanation-based annotation grounded in annotator histories rather than labels alone.
\end{abstract}

\section{Introduction}

Human label variation (HLV; \citealp{plank-2022-problem}) highlights that annotator disagreement often captures meaningful differences in interpretation, rather than noise to be collapsed into a majority label \citep{DBLP:journals/aim/AroyoW15,pavlick-kwiatkowski-2019-inherent,plank-etal-2014-learning,DBLP:journals/jair/UmaFHPPP21,DBLP:conf/aaai/CabitzaCB23}. 
Most work on HLV represents such disagreement as label distributions~\cite{nie-etal-2020-learn,zhou-etal-2022-distributed,chen-etal-2024-seeing,lu2026aligning,chen-etal-2026-decoupling}. 
However, labels alone provide only a partial view of annotator behavior: they indicate \textit{what} annotators choose, but not \textit{why}. 
Free-text explanations provide complementary evidence by revealing how annotators interpret an input, which evidence they attend to, and how they justify their decisions~\cite{DBLP:journals/tacl/JiangM22,jiang-etal-2023-ecologically,hong-etal-2026-agree}. 
Recent work has therefore used explanations to better understand disagreement and annotator reasoning in sentence-pair tasks such as natural language inference (NLI) and paraphrase judgment \citep{basile-etal-2021-need,weber-genzel-etal-2024-varierr,chen-etal-2025-rose,chen-etal-2025-threading,hong-etal-2025-litex}. 
This raises a practical question: how can large language models (LLMs) \textit{simulate annotator-specific label-explanation behavior}, rather than label alone, to produce interpretable, useful, and scalable HLV annotations?

Before modeling such behavior, we first ask whether annotator-specific behavior is observable. We examine two 4-annotator datasets with explanations, covering both NLI and paraphrase judgment settings \citep{weber-genzel-etal-2024-varierr,leonardelli-etal-2025-lewidi}. 
This diagnostic is non-trivial because explanations for the same input are content-dominated: annotators may mention similar semantic relations even when their broader reasoning styles differ. In NLI, for instance, two annotators may both note lexical overlap, but treat it differently---as sufficient evidence for entailment or as compatible with a neutral judgment due to unsupported information.
As a result, single explanations provide only weak evidence of annotator-specific patterns. 
We therefore remove input-content effects and aggregate annotations by annotator, revealing \textbf{stable annotator-specific patterns that are better observed in repeated behavior than in isolated responses.}

We then ask whether LLMs can learn such behavior. 
We compare prompting and supervised fine-tuning (SFT) as simulation baselines, covering in-context and profile prompting, and parameter adaptation~\citep{DBLP:conf/nips/BrownMRSKDNSSAA20,sorensen-etal-2025-value,DBLP:conf/iclr/HuSWALWWC22}. 
Beyond these baselines, we propose cross-annotator preference optimization (CAPO), which uses variation among annotators on the same input as contrastive supervision. 
For a target annotator, CAPO treats the target's label-explanation response as preferred and other annotators' responses as valid but less target-specific alternatives, rather than as incorrect outputs. 
Implemented with direct preference optimization~\citep{DBLP:conf/nips/RafailovSMMEF23}, CAPO learns to \textbf{prefer responses that better match the target annotator while still respecting the space of plausible human annotations.}

This simulation setting also requires evaluation beyond decision matching or single-reference explanation similarity. 
We therefore separate target-label prediction, reference explanation similarity, and recognizable annotator imitation. 
While ROUGE, BERTScore, and embedding cosine similarity~\citep{lin-2004-rouge,DBLP:conf/iclr/ZhangKWWA20,reimers-gurevych-2019-sentence} capture aspects of explanation overlap, they are difficult to interpret as measures of annotator-specificity and often lack sensitivity when comparing simulation methods. 
Motivated by our diagnostic analysis, we emphasize aggregation-aware metrics that evaluate groups of outputs for the same target annotator, complemented by LLM-as-judge attribution and human validation. 
Across Qwen3~\citep{qwen3technicalreport} and Llama3.2~\citep{DBLP:journals/corr/abs-2407-21783}, prompting is limited and unstable, SFT is a stronger baseline, and CAPO further improves aggregation-aware imitation and judge-based attribution. 
Overall, our results suggest that HLV can be modeled as annotator-specific label-explanation behavior, but that reliable evaluation of individual annotator style requires grouping outputs across instances.

We make three main contributions:

\begin{itemize}[leftmargin=*, noitemsep, topsep=0pt]
    \item We diagnose annotator-specific behavior, which is difficult to characterize at the single-instance level but becomes stable after input-content removal and annotator-level aggregation.
    \item We propose cross-annotator preference optimization (CAPO), using variation among valid human annotations as contrastive supervision for modeling target-specific label-explanation behavior.
    \item We evaluate prompting, SFT, and CAPO with decision matching, reference similarity, aggregation-aware imitation, and human validation, 
    showing that CAPO improves pattern-level recognizability of target annotator over baselines.
\end{itemize}

\section{Related Work}

\paragraph{Explanation-grounded human label variation.}
HLV and perspectivist NLP challenge the view of disagreement as annotation noise \citep{DBLP:journals/aim/AroyoW15,plank-etal-2014-learning,DBLP:journals/jair/UmaFHPPP21,plank-2022-problem,DBLP:conf/aaai/CabitzaCB23}, motivating label-distribution modeling and annotator-aware learning from disagreement \citep{nie-etal-2020-learn,zhou-etal-2022-distributed,wang-plank-2023-actor,leonardelli-etal-2025-lewidi,lu2026aligning,chen-etal-2026-decoupling}. In sentence-pair tasks such as NLI, such variation often reflects ambiguity, uncertainty, or alternative interpretations that are obscured by majority labels  \citep{pavlick-kwiatkowski-2019-inherent,chen-etal-2020-uncertain}. Yet label distributions alone reveal little about why annotators disagree or how they justify their decisions.
Free-text explanations address this gap by exposing annotators' reasoning, evidence use, and decision processes \citep{DBLP:journals/tacl/JiangM22,jiang-etal-2023-ecologically,weber-genzel-etal-2024-varierr,hong-etal-2025-litex,hong-etal-2026-agree}. Recent work has used explanations to analyze disagreement, identify plausible variation, and approximate human label distributions \citep{chen-etal-2024-seeing,chen-etal-2025-rose,chen-etal-2025-threading,DBLP:journals/corr/abs-2511-08949,panahi-etal-2026-llms}. Our work differs by treating explanations not only as diagnostic evidence or auxiliary rationales, but as part of annotator-specific behavior to be learned and evaluated.

\paragraph{LLM-based annotator simulation.}
LLMs are increasingly used to generate labels, rationales, and simulated human responses~\cite{DBLP:journals/corr/abs-2303-15056,DBLP:conf/uist/ParkOCMLB23}. Existing work often targets aggregate behavior, majority labels, or population-level variation, while persona- and profile-based prompting conditions models on demographic, social, or role descriptions~\cite{deshpande-etal-2023-toxicity,DBLP:conf/icml/SanturkarDLLLH23,DBLP:journals/corr/abs-2209-06899}. Such approaches can expose perspective sensitivity, but they are often weakly grounded in actual annotation histories and may reduce annotator behavior to coarse-grained or stereotype-prone profiles \citep{cheng-etal-2023-marked,orlikowski-etal-2023-ecological,orlikowski-etal-2025-beyond,sorensen-etal-2025-value}.
In contrast, we study data-grounded annotator behavior modeling under HLV, where target behavior is inferred from repeated labels and explanations by the same annotator rather than imposed through external personas or optimized only for aggregate agreement. 
This shifts evaluation from decision matching or surface similarity to whether generated outputs preserve recognizable annotator-specific patterns across instances.
Our setting can also be viewed as a constrained, data-grounded form of personalization.
We use the term \emph{annotator simulation} to emphasize the evaluation target: fidelity to an observed annotator's historical joint behavior over decisions and explanations, rather than optimization for satisfaction, usefulness, or a newly stated preference.
\begin{figure}[t]
\centering
    \subfigure[Pairwise label agreement]{
        \includegraphics[width=0.46\linewidth]{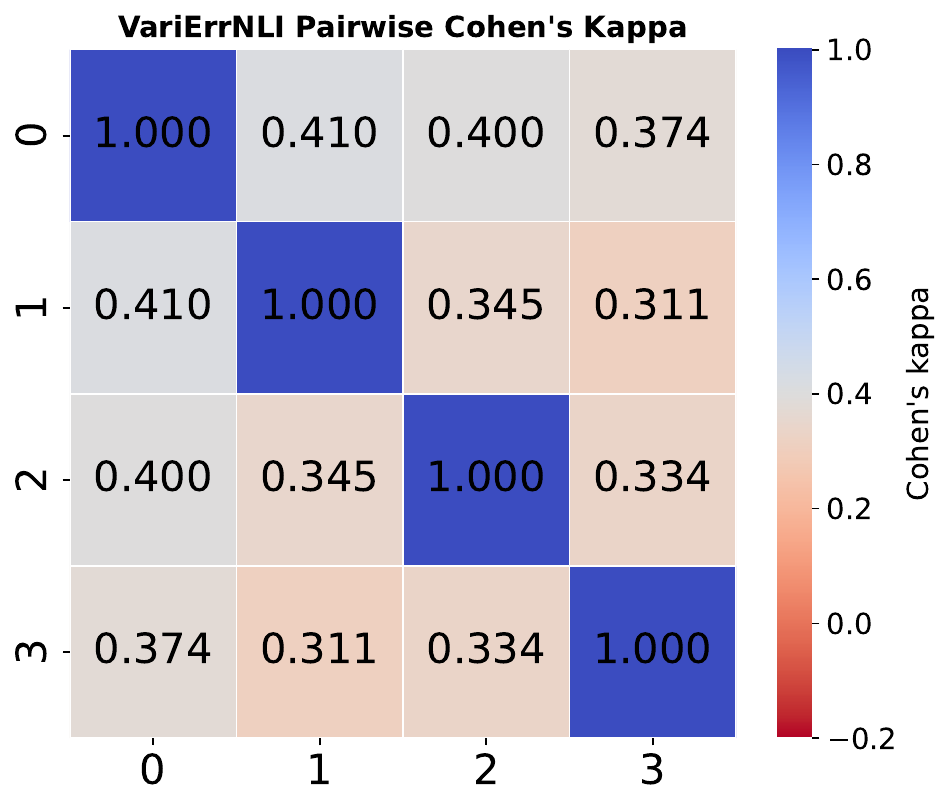}
    }
    \subfigure[Label proportions]{
        \includegraphics[width=0.46\linewidth]{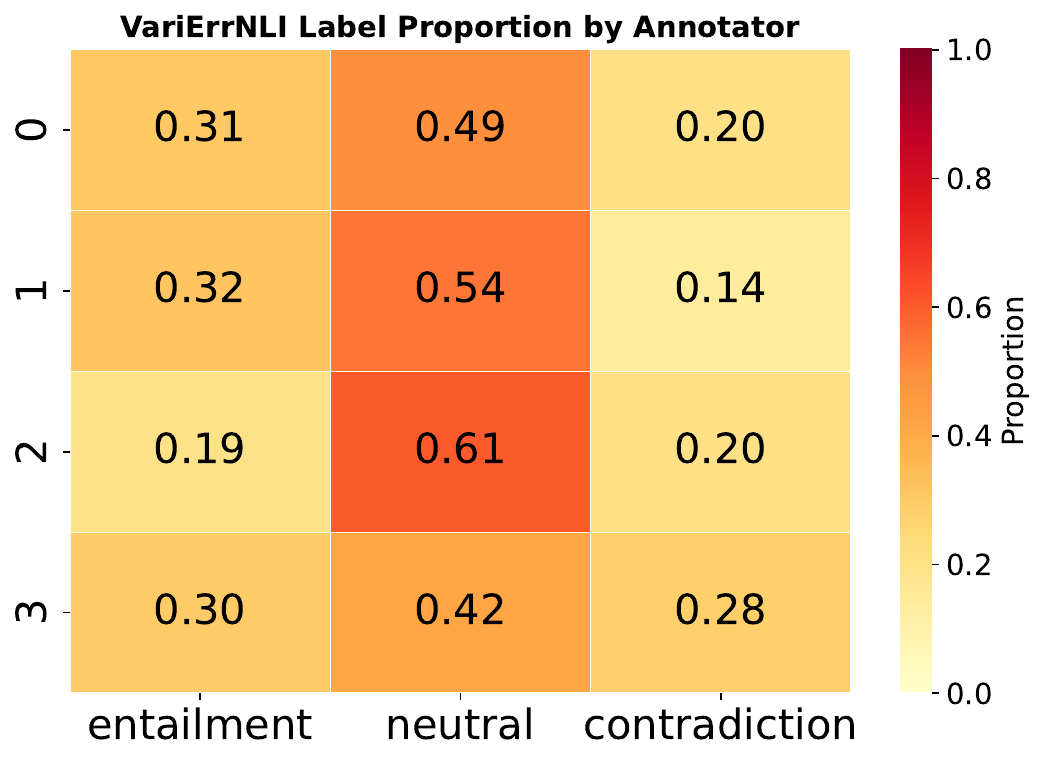}
    }
\caption{VariErr label variation across annotators. 
Annotators differ both in pairwise agreement and in their label priors, 
suggesting that HLV contains annotator-specific structure rather than only item-level ambiguity.}
\label{fig:varierr-label-variation}
\end{figure}

\section{Stability of Annotator Variation}
\label{sec:stable}

A prerequisite of our simulation is to assess \textit{whether stable annotator-specific behavior exists.} 
To do so, we analyze two four-annotator datasets with per-decision explanations: VariErr~\citep{weber-genzel-etal-2024-varierr}, which contains categorical NLI label-explanation judgments, and R2~\citep{leonardelli-etal-2025-lewidi}, which contains ordinal paraphrase-relatedness score-explanation judgments.
We here focus on VariErr NLI, the more conservative of our two datasets.\!\footnote{Dataset details are in Section~\ref{subsec:dataset} and Appendix~\ref{app:data-splits}} R2 paraphrase-judgment results follow the same pattern, with stronger separability, and are reported in Appendix~\ref{app:r2-stable}. In VariErr, each input is a premise-hypothesis pair $x_i=(p_i,h_i)$, and each annotator $a\in\{0,1,2,3\}$ provides an NLI label $y_{i,a}$ (for R2 it's a score) and a free-text explanation $e_{i,a}$. An annotation is therefore $z_{i,a}=(y_{i,a},e_{i,a})$.

\subsection{Annotators Differ in Label Use}
\label{sec:stable_labels}

We first examine label variation. Figure~\ref{fig:varierr-label-variation} shows pairwise Cohen's $\kappa$~\cite{cohen1960coefficient} and each annotator's label proportions over all VariErr annotations. Agreement is moderate: off-diagonal $\kappa$ values range from $0.31$ to $0.41$. The label marginals also differ. Some global annotator preferences emerge: for example, Annotator 2 assigns \textsc{neutral} most often and \textsc{entailment} least often, while Annotator 3 assigns most \textsc{contradiction}. Thus, annotators are not interchangeable even before explanations are considered.
This label-level structure is the basis of perspectivist and annotator-aware modeling approaches (e.g.,~\citealp{DBLP:journals/tacl/DavaniDP22,DBLP:journals/aim/AroyoW15}), but our question is whether such annotator-specific structure also extends to the joint label-explanation behavior.

\begin{figure}[t]
\centering
\includegraphics[width=0.91\columnwidth]{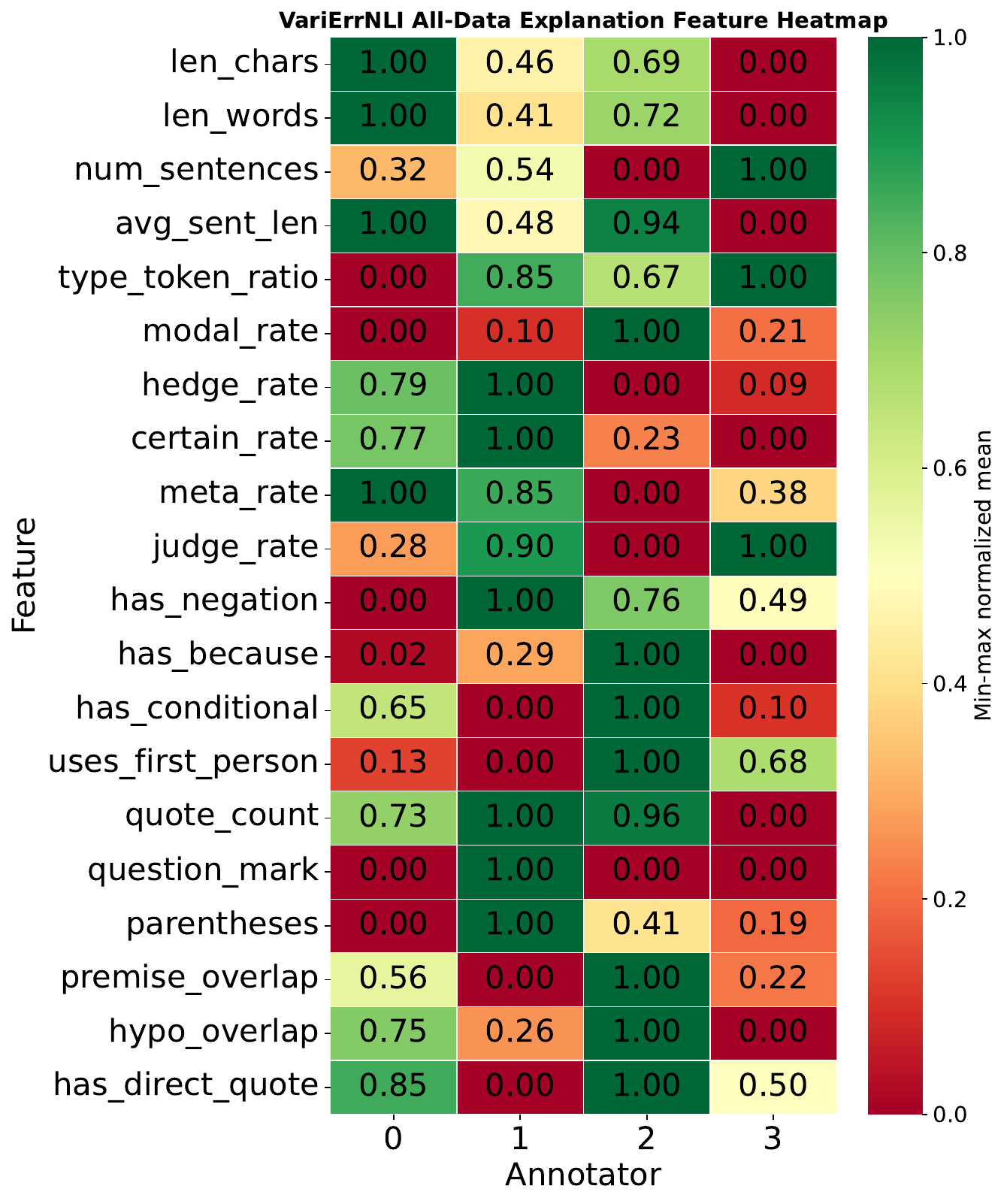}
\caption{VariErr explanation-style features averaged by annotator and min-max normalized within each feature. Feature differences provide interpretable evidence of repeated writing habits, such as length, modality, negation, and quotation; feature details are in Appendix~\ref{app:feature-inventory}.}
\label{fig:varierr-feature-heatmap}
\end{figure}

\begin{figure}[t]
\centering
\includegraphics[width=0.85\columnwidth]{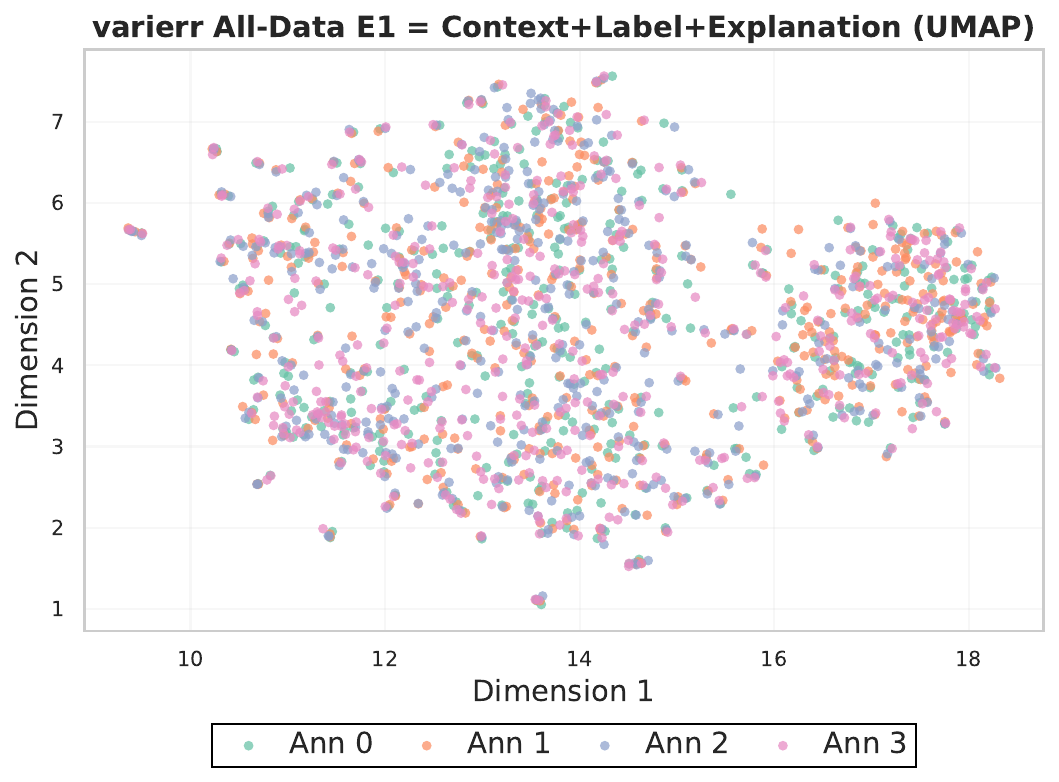}
\caption{UMAP of VariErr $E_1$ embeddings. Annotator colors are substantially mixed, indicating that raw semantic embeddings are dominated by item content.}
\label{fig:varierr-feature-umap}
\end{figure}

\begin{figure*}[t]
\centering
\includegraphics[width=0.95\textwidth]{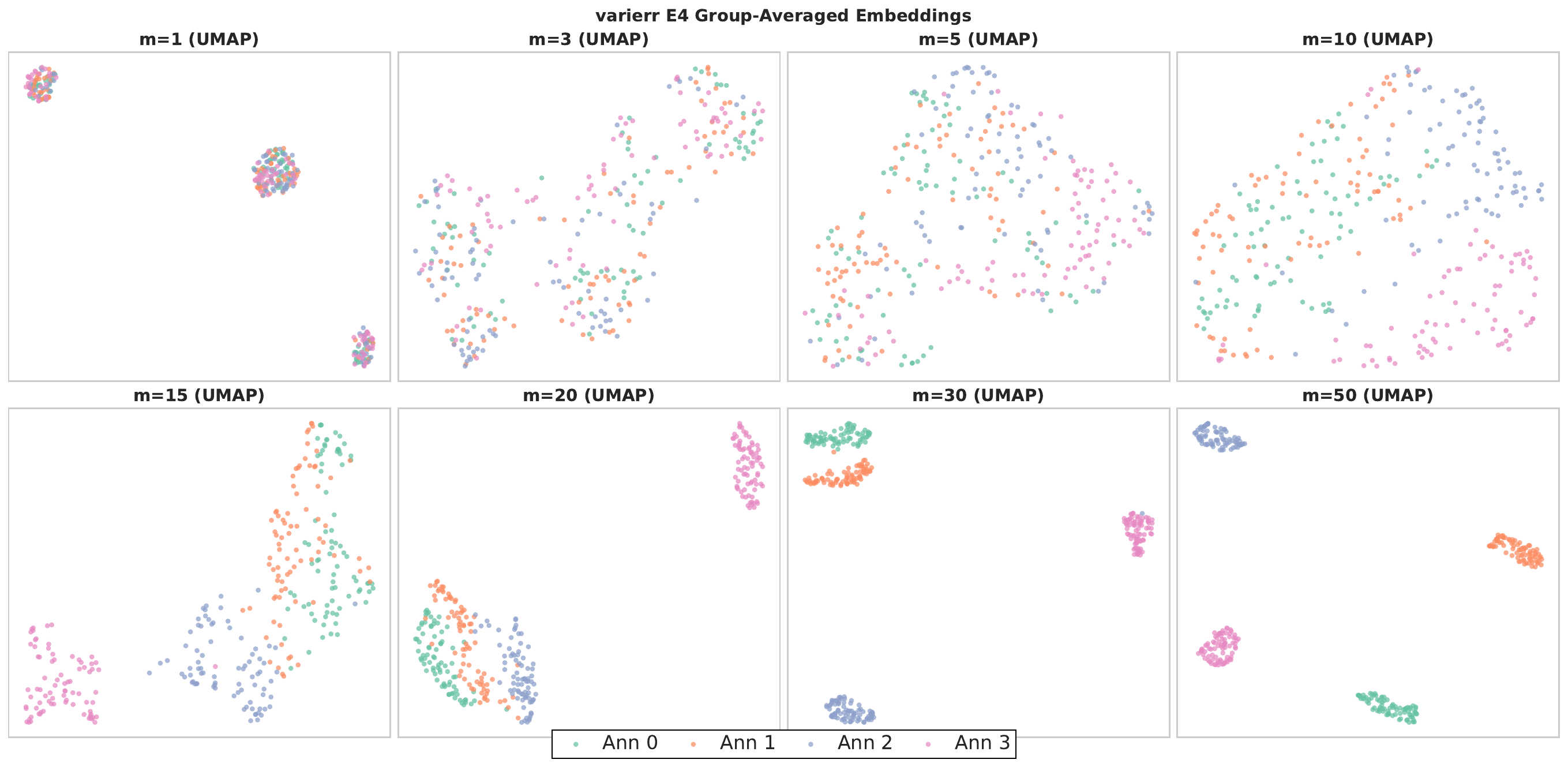}
\caption{VariErr group-averaged $E_4$ for increasing $m$. Each visualization samples 80 groups per annotator. Averaging annotations by annotator suppresses item-specific content and reveals stable annotator-specific structure.}
\label{fig:varierr-e4-group-umap}
\end{figure*}

\subsection{Explanation Style Is Visible, But Raw Embeddings Are Content-Dominated}
\label{sec:stable_explanations}

We next ask whether explanations carry annotator-specific signal and propose ways to quantify it. We compute two complementary views over all VariErr annotations. First, we extract handcrafted surface-level features to capture explanation style, including length, sentence count, type-token ratio, and linguistic features.
These features are designed to capture how an annotator writes.
Second, we embed the input-label-explanation as $E_1$:\footnote{sentence-transformers/all-mpnet-base-v2 on huggingface.}
\begin{equation}
E_1 = f(p_i,h_i,y_{i,a},e_{i,a}).
\end{equation}

The feature heatmap (Figure~\ref{fig:varierr-feature-heatmap}) reveals small but repeated surface-level tendencies,
e.g., Annotator 0 writes the longest explanations on average,
and Annotator 2 has the highest modal-word and first-person rates. 
In contrast, the $E_1$ UMAP (Figure~\ref{fig:varierr-feature-umap}) is heavily mixed across annotators. We hypothesize it is because $E_1$ contains the input,
making explanations for the same or similar inputs semantically similar regardless of annotator-specific habits.

\paragraph{Within-item permutation baseline.}
Figure~\ref{fig:varierr-feature-heatmap} is intended to capture a multivariate annotator profile rather than uniquely identifying annotators from individual features.
We therefore perform a within-item permutation test that keeps each item's four explanations and label/score composition fixed while randomly reassigning annotator identities, preserving item content but removing persistent identity.
Across 5,000 permutations, the observed mean feature effect size is substantially larger than the permutation null on both datasets (Appendix~\ref{app:permutation}), showing that the observed feature differences cannot be explained by random regroupings of content-matched explanations.

\subsection{Single-Annotation Classifiers Are Informative But Insufficient}
\label{sec:stable_single_classifier}

To quantify whether this variation is learnable, we train annotator 4-way classifiers\!\footnote{The feature classifier uses a \texttt{StandardScaler} followed by balanced logistic regression~\cite{DBLP:journals/jmlr/PedregosaVGMTGBPWDVPCBPD11}; embedding classifiers use balanced logistic regression on $E_1$.} on an item-level split: 250 items for training and 250 disjoint items for testing, with all annotator rows for an item kept on the same side.  
The handcrafted features reach $50.2\%$ test accuracy, confirming that explanation style is partially recoverable from a single annotation. However, raw $E_1$ reaches only $41.9\%$. This supports the qualitative UMAP result: semantic embeddings of individual explanations are strongly entangled with the question content. 
To test whether content removal improves annotator recoverability, we construct additional representations that separate the input from the label-explanation response:
\begin{equation}
\begin{aligned}
E_0 = f(p_i,h_i), E_2 = f(y_{i,a},e_{i,a}),
\end{aligned}
\end{equation}
While $E_2$ can still reflect the input, we define two residual embeddings after normalizing $E_0$ and $E_2$:
\begin{equation}
\begin{aligned}
E_3 &= \operatorname{norm}(E_2-E_0),\\
E_4 &= \operatorname{norm}\left(E_2-(E_2^\top E_0)E_0\right).
\end{aligned}
\end{equation}
These residuals are intended as diagnostics, not as perfect content-removal methods.
Among single-annotation embeddings, $E_4$ performs best, reaching $56.5\%$ test accuracy (Appendix~\ref{app:varierr-extra}), improving over $E_1$ but still falling short of reliable instance-level identification.
This suggests that individual explanations remain mixtures of annotator-specific tendencies and item-specific content, limiting their usefulness as standalone signals for training or evaluation.

\subsection{Aggregation Reveals Stable Behavior}
\label{sec:stable_group}

\begin{figure}[t]
\centering
\includegraphics[width=0.90\columnwidth]{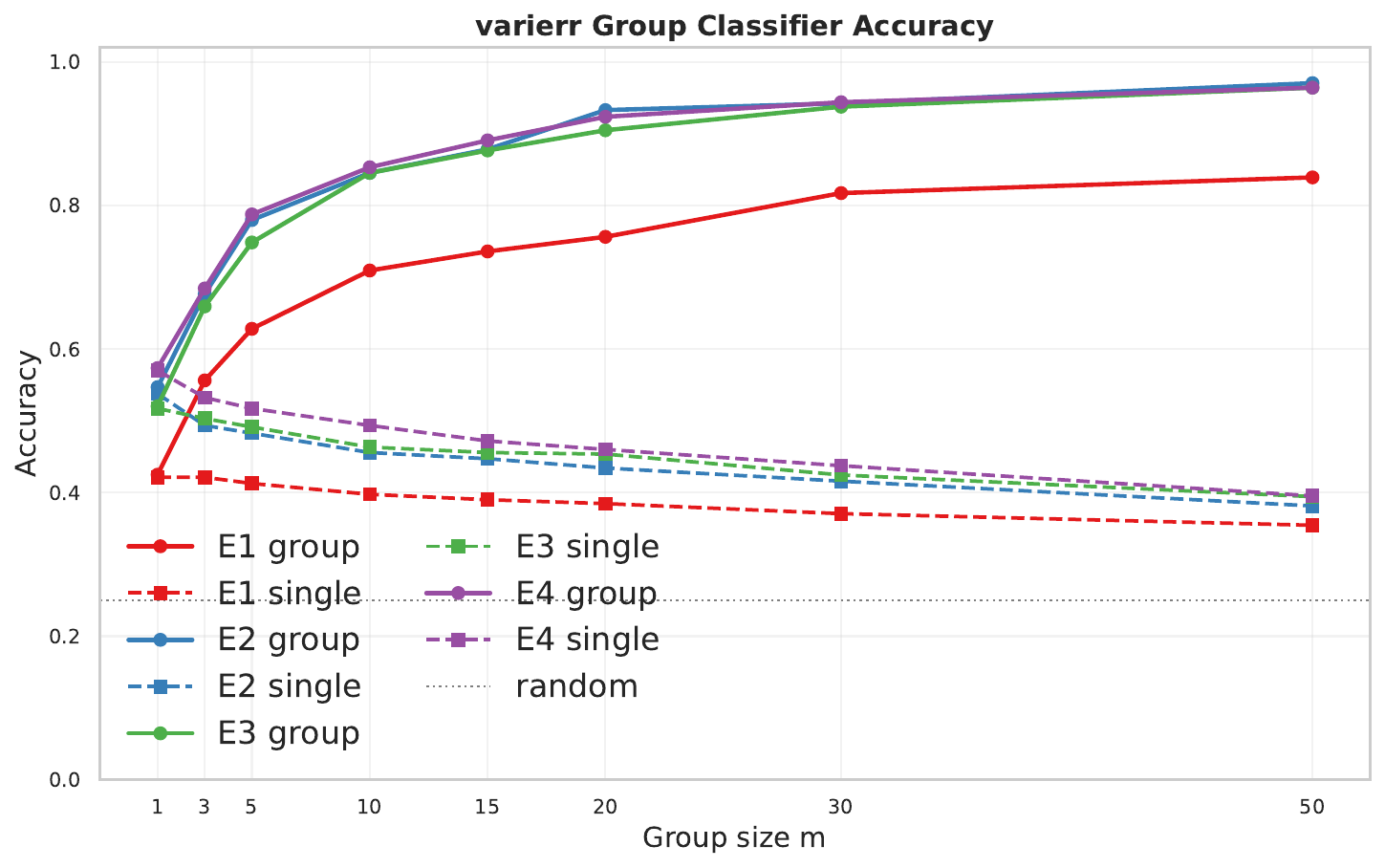}
\caption{VariErr annotator-classifier accuracy by representation and group size $m$ (240 train groups and 160 test groups per annotator). Solid lines evaluate grouped test annotations; dashed lines apply the same group-trained classifiers to individual test annotations. Aggregation sharply improves annotator recoverability, especially for content-reduced representations. Detailed group classifier sweep results are in Appendix~\ref{app:group-sweep}.}
\label{fig:varierr-group-curve}
\end{figure}

We therefore move from individual annotations to groups. The intuition is simple:\footnote{This intuition is related to averaging-based directions in activation steering~\cite{DBLP:conf/nips/ArditiOSPPGN24,turner2023steering}.} \textit{if we average many embeddings from the same annotator across different items, item-specific semantic directions should cancel more than repeated annotator-specific tendencies.} For each $E_1$--$E_4$ and group size $m$, we sample groups of $m$ annotations from the same annotator, average their normalized embeddings, and renormalize the group vector. Classifiers are trained on group vectors from train set and evaluated on group vectors from test set. 
Figure~\ref{fig:varierr-e4-group-umap} shows group-averaged $E_4$ embeddings. The clusters become increasingly annotator-specific as $m$ grows. Figure~\ref{fig:varierr-group-curve} quantifies the same effect: the VariErr $E_4$ group classifier rises from $57.3\%$ at $m=1$ to 
$96.4\%$ at $m=50$.\!\footnote{Due to random test groups, accuracy at $m=1$ slightly differs from single-annotation classifiers.}
By contrast, classifiers trained on larger group means are not reliable single-instance classifiers. This gap is the key diagnostic result: 
target-specific behavioral patterns are much clearer when observed as repeated behavior than as properties of isolated explanations.

We use $E_4$ as the main diagnostic representation because it is the strongest
single-annotation representation and has the clearest content-debiasing
interpretation; Appendix~\ref{app:linear-e0-recovery} further shows that it has
the lowest linear recoverability of the input.
At larger group sizes, however, $E_2$--$E_4$ perform similarly, suggesting that aggregation suppresses much of the item-level variation. 
Thus, our main claim is not that individual annotations are reliably classifiable, but that repeated label-explanation behavior forms a stable annotator-specific signal,
motivating the following aggregation-aware imitation metrics.

\section{Annotator Simulation and Evaluation}
\label{sec:methods}
\label{sec:experimental_setup}

We next study \textit{whether LLMs can simulate the annotator-specific label-explanation behavior}. 
We formulate this as conditional generation: given an input question and a target annotator ID, the model outputs both the task decision and an explanation.

\subsection{Methods}

We compare methods with increasing supervision strength, summarized in Table~\ref{tab:method-summary}: prompting, supervised fine-tuning (SFT), and cross-annotator preference optimization (CAPO). 
All methods condition on the same symbolic target annotator ID, but differ in whether and how they ground this ID in the annotator's past behavior; see details in Appendix~\ref{app:parameters}.

\paragraph{Prompting.} The {Base} setting provides only the input pair and target ID, without annotation history, demonstrations, or profiles; it therefore measures default behavior under the same output schema rather than true annotator conditioning. 
Prompt-based methods ground the target ID at inference time using $n$ examples from the target annotator. 
We consider in-context learning (ICL), which provides demonstrations directly; value profiles (VP;~\citealp{sorensen-etal-2025-value}), which summarize the annotator's label tendencies and explanation style into a natural-language profile; and VP-ICL, which combines both. 
In the main experiments we use $n=50$, with prompt in Appendix~\ref{app:prompts}, $n$ sweeps in Appendix~\ref{app:prompt-size-sweep}, and profiles in Appendix~\ref{app:value-profiles}.

\subsection{Task and Model}
\label{subsec:dataset}

We evaluate on two datasets with annotator-level labels and explanations: VariErr NLI~\cite{weber-genzel-etal-2024-varierr}, where outputs are entailment, neutral, or contradiction labels, and R2 paraphrase judgment~\cite{leonardelli-etal-2025-lewidi}, where outputs are integer relatedness scores from $-5$ to $5$. 
All methods use the same train/dev/test item splits, summarized in Table~\ref{tab:dataset-summary}. 
Experiments are conducted with Qwen3-4B-Instruct-2507 (Qwen3;~\citealt{qwen3technicalreport}) and Llama-3.2-3B-Instruct (Llama3.2;~\citealt{DBLP:journals/corr/abs-2407-21783}).

\begin{table}[t]
\centering
\small
\begin{adjustbox}{max width=\columnwidth}
\begin{tabular}{lllll}
\toprule
Dataset & Task & Output & Split & Task Metric \\
\midrule
VariErr & NLI & label + explanation & 300/100/100 & Label Acc \\
R2 & paraphrase judgment & score + explanation & 300/100/100 & Score Acc; MAE \\
\bottomrule
\end{tabular}
\end{adjustbox}
\caption{Dataset summary. Both datasets contain 4 annotators per item. Details are in Appendix~\ref{app:data-splits}.}
\label{tab:dataset-summary}
\end{table}

\begin{table}[t]
\centering
\small
\begin{adjustbox}{max width=\columnwidth}
\begin{tabular}{llll}
\toprule
Method & Annotator signal & Parameter update & Role \\
\midrule
Base & Symbolic ID only & No & format control \\
ICL & $n$ examples & No & example-only prompting \\
VP & profile from $n$ examples & No & global summary \\
VP-ICL & profile + $n$ examples & No & strongest prompt baseline \\
SFT & target annotations & Yes & learns target behavior \\
CAPO & cross-annotator preference pairs & Yes & sharpens target contrast \\
\bottomrule
\end{tabular}
\end{adjustbox}
\caption{Main annotator simulation methods. 
}
\label{tab:method-summary}
\end{table}

\paragraph{SFT.}
For each annotator $a$, we train an independent LoRA adapter~\cite{DBLP:conf/iclr/HuSWALWWC22} on that annotator's training annotations.\footnote{We use independent adapters because a shared annotator-ID-conditioned variant showed much weaker annotator separation despite using all annotators' data, suggesting that target behavior is better captured in separate adapted parameters than recovered from symbolic IDs; see Appendix~\ref{app:shared_sft}.}
Given the base model $\pi_0$, the adapted policy is denoted
$\pi_{\theta_a}^{\mathrm{SFT}}(z\mid x,a)$, where $z=(y,e)$.
We optimize the standard conditional language-modeling objective:
\begin{equation}
\mathcal{L}_{\mathrm{SFT}}(a)
= - \sum_{(x_i,z_{i,a}) \in \mathcal{D}_a}
\log \pi_{\theta_a}^{\mathrm{SFT}}(z_{i,a}\mid x_i,a).
\label{eq:sft-loss}
\end{equation}
Thus, SFT stores target-annotator behavior in independent adapted parameters rather than recovering it from a profile string or symbolic annotator ID.

\subsection{Cross-Annotator Preference Optimization}
\label{sec:capo}

We propose CAPO that continues training from the corresponding SFT adapter for annotator $a$ and uses cross-annotator variation as contrastive preference supervision. For an input $x_i$, the chosen response is the target annotator's annotation $z^+=z_{i,a}$ and the rejected response is another annotator's annotation $z^-=z_{i,b}$ for the same input, with $b\ne a$. The rejected response is not treated as wrong; it is a valid human alternative that is less target-specific under the goal of imitating annotator $a$.

Let $\pi_{\theta_a}$ be the trainable CAPO policy initialized from $\pi_{\theta_a}^{\mathrm{SFT}}$, and let $\pi_{\mathrm{ref},a}=\pi_{\theta_a}^{\mathrm{SFT}}$ be the frozen reference policy. Define:
\begin{equation}
\begin{aligned}
r_{\theta_a}(z;x_i,a)
&= \log \pi_{\theta_a}(z\mid x_i,a) \\
&\quad - \log \pi_{\mathrm{ref},a}(z\mid x_i,a),
\end{aligned}
\label{eq:capo-reward}
\end{equation}
CAPO optimizes a DPO-style loss~\cite{DBLP:conf/nips/RafailovSMMEF23}:
\begin{equation}
\begin{aligned}
\mathcal{L}_{\mathrm{CAPO}}
&= -\log \sigma\!\Big(
\beta\big[
r_{\theta_a}(z^+;x_i,a) \\
&\quad - r_{\theta_a}(z^-;x_i,a)
\big]\Big).
\end{aligned}
\label{eq:capo-loss}
\end{equation}
This turns variation among valid human annotations into supervision for target-specific adaptation, rather than into a quality ranking among annotators.

\paragraph{Pair construction.}
We use conservative pair policies so that CAPO cannot rely only on large label or score differences.
For VariErr, the main policy pairs annotations with the same NLI label, $y_{i,a}=y_{i,b}$, making the preference signal primarily explanation-based.
For R2, where exact equality is too sparse over the 11-point scale, we require $|s_{i,a}-s_{i,b}|\le 1$.
Pair counts and policy ablations are reported in Appendices~\ref{app:capo-pair-counts} and~\ref{app:capo-pair-policy-ablation}.

\paragraph{Checkpoint selection.}
Because CAPO targets recognizable annotator-specific behavior rather than likelihood alone, we select checkpoints with an aggregation-aware development criterion.
For each candidate checkpoint, we generate development outputs for each target annotator and choose the checkpoint whose output groups receive the highest target-annotator confidence from a group classifier trained only on train set, which follows Section~\ref{sec:stable} that annotator-specific behavior is more reliable at the group level than at the single-output level.
The selector does not use test outputs or classifiers trained on test annotations; implementation details and ablations are in Appendix~\ref{app:capo-checkpoint-ablation}.

\begin{table}[t]
\centering
\small
\begin{adjustbox}{max width=\columnwidth}
\begin{tabular}{llll}
\toprule
View & Evaluates & Main metrics & Main limitation \\
\midrule
Decision & task decision agreement & Acc; MAE & ignores reasoning style \\
Reference & explanation overlap & ROUGE-L; BERTScore; EmbSim & input content can dominate \\
Imitation & annotator-specific behavior & Feature KL; GC; Judge Acc & single instances are ambiguous \\
\bottomrule
\end{tabular}
\end{adjustbox}
\caption{Evaluation views. Our main claim concerns annotator imitation, but we interpret it alongside decision matching and reference-based explanation similarity.}
\label{tab:evaluation-views}
\end{table}

\begin{table*}[t]
\centering
\scriptsize
\setlength{\tabcolsep}{3.2pt}
\renewcommand{\arraystretch}{1.08}
\begin{adjustbox}{max width=\textwidth}
\begin{tabular}{llrrrrrr|rrrrrrrr}
\toprule
\multirow{2}{*}{Model} & \multirow{2}{*}{Method}
& \multicolumn{6}{c|}{VariErr}
& \multicolumn{7}{c}{R2} \\
\cmidrule(lr){3-8} \cmidrule(lr){9-15}
& 
& Label Acc $\uparrow$
& ROUGE-L $\uparrow$
& Feature KL $\downarrow$
& GC Conf $\uparrow$
& ImiScore $\uparrow$
& Judge Acc $\uparrow$
& Score Acc $\uparrow$
& Score MAE $\downarrow$
& ROUGE-L $\uparrow$
& Feature KL $\downarrow$
& GC Conf $\uparrow$
& ImiScore $\uparrow$
& Judge Acc $\uparrow$ \\
\midrule

\multirow{6}{*}{Qwen3}
& Base
& 0.472 & 0.159 & 2.394 & 0.239 & -0.034 & 0.237
& 0.292 & 2.453 & 0.071 & 5.188 & 0.248 & -0.004 & 0.237 \\
& ICL
& 0.535 & 0.186 & 0.692 & 0.344 & 0.179 & 0.265
& 0.407 & 1.702 & 0.197 & 0.587 & 0.809 & 0.810 & 0.463 \\
& VP
& 0.492 & 0.131 & 4.391 & 0.234 & -0.348 & 0.240
& 0.278 & 2.475 & 0.100 & 2.740 & 0.649 & 0.651 & 0.360 \\
& VP-ICL
& 0.545 & 0.201 & 0.476 & 0.494 & 0.528 & 0.282
& 0.415 & 1.667 & 0.276 & 0.246 & 0.947 & 0.956 & 0.510 \\
& SFT
& \textbf{0.638} & \textbf{0.283} & 0.084 & 0.845 & 0.859 & 0.300
& 0.450 & \textbf{1.478} & 0.307 & 0.130 & 0.929 & 0.936 & 0.512 \\
& \textbf{CAPO}
& 0.627 & 0.281 & \textbf{0.081} & \textbf{0.867} & \textbf{0.888} & \textbf{0.328}
& \textbf{0.455} & 1.490 & \textbf{0.316} & \textbf{0.112} & \textbf{0.964} & \textbf{0.973} & \textbf{0.560} \\

\midrule

\multirow{6}{*}{Llama3.2}
& Base
& 0.315 & 0.153 & 2.036 & 0.258 & 0.023 & 0.265
& 0.207 & 3.188 & 0.068 & 1.515 & 0.255 & 0.015 & 0.230 \\
& ICL
& 0.380 & 0.214 & 0.476 & 0.295 & 0.375 & 0.260
& 0.273 & 3.333 & 0.213 & 0.480 & 0.871 & 0.879 & 0.367 \\
& VP
& 0.338 & 0.148 & 1.907 & 0.358 & 0.325 & 0.253
& 0.180 & 4.185 & 0.098 & 0.913 & 0.359 & 0.175 & 0.260 \\
& VP-ICL
& 0.412 & 0.210 & 0.539 & 0.260 & 0.432 & 0.275
& 0.273 & 3.390 & 0.230 & 0.257 & 0.858 & 0.865 & 0.370 \\
& SFT
& \textbf{0.512} & \textbf{0.271} & \textbf{0.113} & 0.909 & 0.946 & 0.282
& 0.450 & 1.535 & 0.290 & \textbf{0.118} & 0.963 & 0.972 & 0.512 \\
& \textbf{CAPO}
& \textbf{0.512} & 0.262 & 0.121 & \textbf{0.924} & \textbf{0.964} & \textbf{0.297}
& \textbf{0.460} & \textbf{1.470} & \textbf{0.293} & 0.187 & \textbf{0.985} & \textbf{0.995} & \textbf{0.530} \\

\bottomrule
\end{tabular}
\end{adjustbox}
\caption{
Main results report overall averages across annotators, with ROUGE-L as the representative similarity metric since BERTScore and EmbSim show the same pattern. 
We interpret higher GC Conf and Judge Acc as stronger preservation of annotator-specific reasoning patterns.
CAPO generally improves imitation metrics while maintaining competitive decision accuracy; bootstrap significance tests in Appendix~\ref{app:significance} further show higher mean ImiScore than SFT in all settings.
Complete metrics and setting results are provided in Appendix~\ref{app:full_results}.
}
\label{tab:main-results}
\end{table*}

\subsection{Evaluation Metrics}
\label{sec:exp_metrics}

We evaluate simulation from three complementary views as in Table~\ref{tab:evaluation-views}: decision matching, explanation similarity, and annotator recognizability. 
Following Section~\ref{sec:stable}, we treat annotator-specificity as a repeated behavior, and therefore complement single-output metrics with aggregation-aware metrics.

\paragraph{Decision matching.}
For VariErr, we report label accuracy against the target annotator. 
For R2, we report exact score accuracy and score mean absolute error (MAE). 
These metrics test whether the model matches the annotator's task decision.

\paragraph{Reference-based explanation similarity.}
Generated explanations are compared with the target annotator's reference explanation by ROUGE-L~\citep{lin-2004-rouge}, BERTScore~\citep{DBLP:conf/iclr/ZhangKWWA20}, and embedding cosine similarity~\citep{reimers-gurevych-2019-sentence}, following~\citet{giulianelli-etal-2023-comes}. 
These metrics capture lexical and semantic overlap, but also reward input content shared across annotators.

\paragraph{Aggregation-aware evaluation.}
We use group-level metrics motivated by Section~\ref{sec:stable}. 
First, we compute mean KL divergence between generated and human distributions of handcrafted explanation-style features, measuring whether a method reproduces repeated stylistic tendencies such as length, lexical reuse, and uncertainty marking.

Second, we evaluate whether generated outputs are recognizable as the intended annotator using a held-out group classifier (GC). 
For each method and target annotator, we form bootstrapped groups of generated outputs, classify their aggregated representations, and report GC confidence: the mean probability assigned to the intended annotator. 
We also report a normalized imitation score:
\begin{equation}
\mathrm{ImiScore}(M,a) =
\frac{C_a(M_a)-C_a(M_{\neg a})}
     {C_a(H_a)-C_a(M_{\neg a})},
\end{equation}
where $C_a(\cdot)$ is the classifier probability assigned to annotator $a$, $M_a$ denotes outputs generated for annotator $a$, $M_{\neg a}$ denotes outputs from the same method targeting other annotators, and $H_a$ denotes human references from annotator $a$. 
ImiScore near 0 indicates no gain over non-target generated outputs, near 1 indicates human-level recognizability, and above 1 indicates amplified recognizability. 
See details and robustness checks in Appendix~\ref{app:group-size-sweep}.

\paragraph{LLM-as-judge.}
As an item-level complement to GC, we use DeepSeek V4 Pro as a single-instance attribution judge~\citep{deepseekai2026deepseekv4}. 
Given a generated decision and explanation, together with four shuffled human annotator candidates for the same item, the judge selects the candidate the output most resembles.\!\footnote{We additionally test candidate-order sensitivity by evaluating selected outputs under all four cyclic candidate orders, so that the intended annotator appears equally often in positions A--D.
The resulting method comparisons are stable across orderings, indicating that the reported Judge Accuracy differences are not driven by one fixed candidate arrangement; see Appendix~\ref{app:judge_order}.}
We report accuracy against the intended annotator, with details in Appendix~\ref{app:judge}.

\section{Results and Analyses}
\label{sec:results}

\subsection{Main Results}
\label{sec:main-results}

Table~\ref{tab:main-results} reports results on VariErr and R2.
Across both datasets and models, performance improves from frozen prompting to parameter-based simulation.
Prompting recovers limited annotator signal when demonstrations are included, with VP-ICL generally the strongest; profile-only prompting is less reliable, suggesting that compact summaries miss fine-grained label-explanation behavior.

SFT substantially outperforms prompting, showing that annotator-specific behavior is learnable from repeated histories. CAPO further improves imitation-oriented metrics while maintaining competitive decision accuracy. Since its preference pairs control for label or score differences, these gains are unlikely to come from decisions alone, but instead reflect sharper explanation-grounded target specificity. The divergence across decision, reference-similarity, and imitation metrics shows that annotator simulation requires behavior-level evaluation beyond label matching or single-reference explanation overlap.

A complementary decomposition further shows that CAPO generally combines stronger target alignment with reduced non-target leakage, rather than improving recognizability solely by suppressing alternative annotators (Appendix~\ref{app:target_leakage}).

\begin{figure}[t]
    \centering
    \includegraphics[width=\columnwidth]{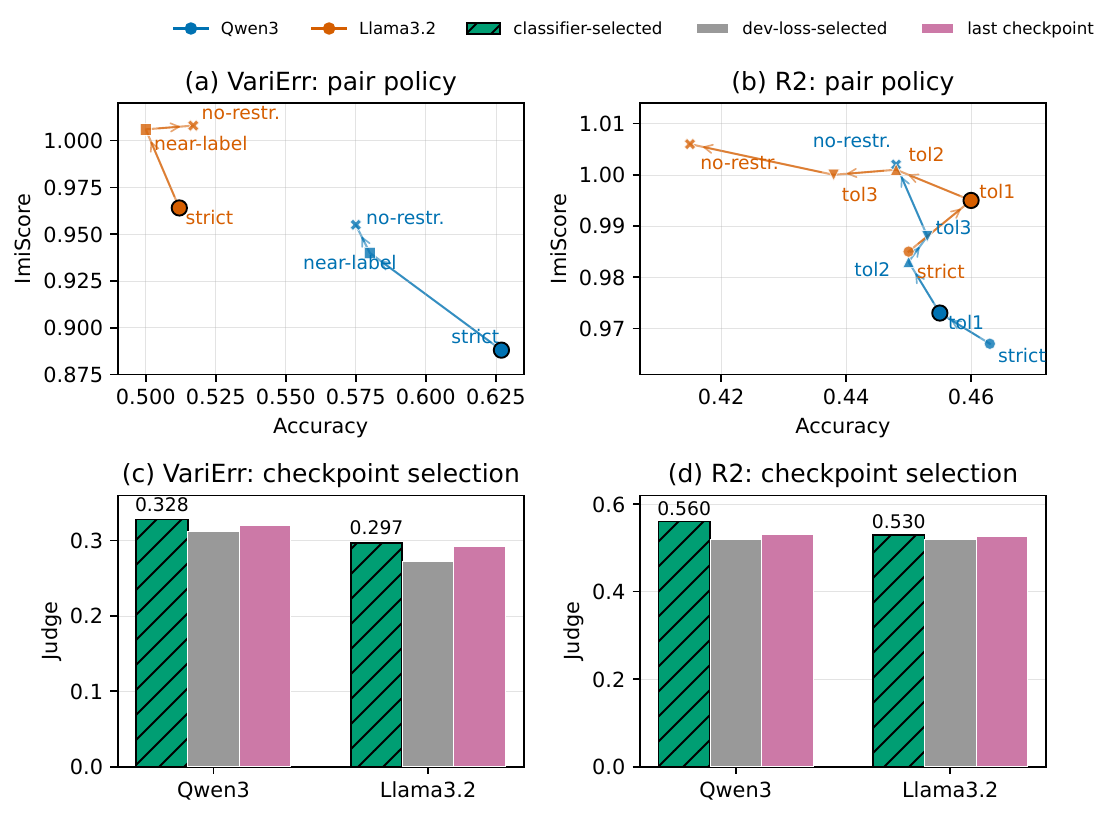}
    \caption{
    CAPO policy and checkpoint effects. Panels (a) and (b) show how pair-policy
    looseness affects the accuracy-imitation trade-off. 
    Panels (c) and (d) compare checkpoint-selection
    strategies using judge-based imitation, which is not the selection criterion.
    }
    \label{fig:capo-policy-selection}
\end{figure}

\subsection{Policy and Selection Effects}

Figure~\ref{fig:capo-policy-selection} summarizes CAPO pair-policy and checkpoint-selection ablations, with full results in Appendix~\ref{app:capo-policy-ablation}. CAPO constructs preference pairs within each input instance, using the target annotator’s label-explanation pair as chosen and another annotator’s pair as rejected. Pair policies control decision similarity: VariErr uses same-label pairs under the strict policy, while looser variants allow near-label or unrestricted pairs; R2 uses same-score pairs under the strict policy and tolerance-$k$ pairs otherwise. Although looser policies may improve recognizability by creating stronger contrasts, they can introduce label or score confounds. We therefore adopt conservative settings in the main experiments: strict same-label pairs for VariErr and tolerance-1 pairs for R2, emphasizing explanation framing and target-specific reasoning.

Checkpoint-selection results provide a sanity check.
Although CAPO selects checkpoints using a group classifier trained only on human training annotations, the selected checkpoints remain competitive under Judge Acc, which is not used for selection.
This suggests that CAPO's gains are not merely artifacts of optimizing the group-classifier-style metric.
The ablations also show that CAPO moves models along an accuracy--imitation frontier rather than uniformly improving all metrics.

\begin{table}[t]
\centering
\small
\begin{adjustbox}{max width=\columnwidth}
\begin{tabular}{llcc}
\toprule
Setting & Feature & SFT KL & CAPO KL \\
\midrule
Qwen-VariErr & \texttt{avg\_sent\_len} & 0.1127 & 0.0233 \\
Qwen-VariErr & \texttt{modal\_rate} & 0.1795 & 0.1274 \\
Qwen-R2 & \texttt{len\_chars} & 0.1691 & 0.0256 \\
Qwen-R2 & \texttt{type\_token\_ratio} & 0.1394 & 0.1111 \\
\bottomrule
\end{tabular}
\end{adjustbox}
\caption{
Representative feature-wise KL. 
}
\label{tab:feature-kl-breakdown}
\end{table}

\subsection{Feature Shifts}
\label{sec:feature-case-study}

To interpret what changes under CAPO, we report representative feature-wise KL shifts in Table~\ref{tab:feature-kl-breakdown}, with full results in Appendix~\ref{app:feature-shift}.
CAPO reduces distribution shift across several interpretable dimensions, including length, sentence structure, modal expressions, and lexical diversity.
This suggests that CAPO improves alignment with multiple aspects of annotator-specific style rather than amplifying a single marker.
Because the effect is not uniform across all settings, we treat Feature KL as a diagnostic rather than an optimization target.

\begin{table*}[t]
\centering
\small
\renewcommand{\arraystretch}{1.25}
\begin{adjustbox}{max width=\textwidth}
\begin{tabular}{p{0.02\textwidth} p{0.14\textwidth} p{0.25\textwidth} p{0.27\textwidth} p{0.27\textwidth} p{0.25\textwidth}}
\toprule
Case & Premise & Hypothesis & Gold Explanation & SFT Explanation & CAPO Explanation \\
\midrule

\textbf{1}
& I guess history repeats itself, Jane.
& I truly think the prior situation shows history repeats itself.
& The context doesn't specify what shows that history repeats itself.
& Both the context and the statement suggest that history repeats itself.
& The context doesn't mention what the speaker truly thinks. \\

\multicolumn{6}{p{1.3\textwidth}}{
\textcolor{gray}{\textit{\textbf{Observation:} SFT mainly follows lexical overlap between the premise and hypothesis, whereas CAPO moves closer to the gold rationale by identifying unsupported information introduced by the hypothesis.}}
} \\
\midrule

\textbf{2}
& He's a bad lot.
& He's a dishonest person.
& Bad people can be both honest and dishonest.
& He can be a dishonest person, but he can also be a good person.
& He's a bad lot, but he can be honest or dishonest. \\

\multicolumn{6}{p{1.3\textwidth}}{
\textcolor{gray}{\textit{\textbf{Observation:} Both systems produce neutral-style explanations, but CAPO preserves the relevant distinction between being ``bad'' and being specifically dishonest, while SFT introduces the irrelevant alternative of being good.}}
} \\

\bottomrule
\end{tabular}
\end{adjustbox}
\caption{
Case study comparing SFT and CAPO on VariErr. 
Case study for R2 is in Appendix~\ref{app:case_study_r2}.
}
\label{tab:case-study}
\end{table*}

\subsection{Human Validation}
\label{sec:human-validation}

We validate whether aggregation-aware gains correspond to human-recognizable explanation quality on 50 VariErr samples. Two annotators evaluate coherence, input grounding, gold-explanation compatibility, and similarity to the target annotator's style and perspective. Both produce the same system ranking, CAPO $>$ SFT $>$ prompting, with $82.8\%$ agreement; see Appendix~\ref{app:human-validation-protocol}. This suggests that CAPO improves not only automatic imitation metrics, but also perceived explanation validity and target specificity.

\paragraph{Case study.}
Table~\ref{tab:case-study} illustrates how these gains arise.
CAPO better preserves the gold annotator's reasoning focus, while SFT more often relies on shallow lexical overlap or introduces irrelevant contrasts.
Thus, CAPO's higher GC Conf and ImiScore reflect more recognizable target-annotator behavior, rather than merely higher surface similarity.

\paragraph{Profile study.}
To interpret what annotator-level variation captures, we compare LLM-generated profiles in Appendix~\ref{app:value-profiles} with manual behavioral characterizations in Appendix~\ref{app:profile_study_details}.
We do not attribute these differences to demographics, education, or psychological traits.
Instead, the profiles reveal recurring task interpretations and explanation strategies, such as evidence checking, uncertainty handling, underspecification sensitivity, aspect-level comparison, and answer-equivalence reasoning.
This helps explain why features such as length, modality, first-person use, and input reuse form stable signals rather than random noise.

\section{Discussion and Future Work}

\paragraph{Robustness with larger prompting models.}
To assess whether our prompting results depend on the 3--4B backbones used in the main experiments, we additionally evaluate Llama-3.1-70B and Qwen3-235B-A22B-Instruct-2507-FP8 using the same prompting protocols on both VariErr and R2.
Increasing model scale substantially strengthens prompting and narrows the gap with parameter-based adaptation on several metrics, particularly task-level performance, but does not uniformly improve annotator-specific imitation.
These results provide a more complete and scale-robust view of prompting while leaving the controlled same-backbone comparison between SFT and CAPO unchanged.
Full results are reported in Appendix~\ref{app:large_prompting}.

\paragraph{Robustness to explicit stylistic markers.}
Aggregated representations may encode explicit stylistic cues, so we additionally test whether annotator recoverability remains after removing embedding components linearly predictable from handcrafted surface and explanation-style features.
Human annotator identity remains almost perfectly recoverable after residualization, although aggressive residualization weakens some differences among generated outputs.
We therefore interpret GC as measuring repeated joint label--explanation behavior, including both semantic framing and recurring writing tendencies, rather than a purely content-independent ``reasoning style.''
Full results are reported in Appendix~\ref{app:marker_residual}.

\paragraph{Broader populations and behavioral representations.}
Extending our framework to larger annotator populations remains an important direction.
Suitable public resources are scarce because our setting requires persistent identities, repeated responses to shared inputs, individual decisions, and per-decision free-text explanations.
Larger personalization datasets such as LiteraryTaste~\citep{DBLP:journals/corr/abs-2511-09310} could support partial evaluation of the preference-learning component, while larger explanation-grounded datasets would enable a fuller test of population-level scalability.
Our $E_3/E_4$ representations also reduce, rather than fully remove, item-related information; complementary style-specialized representations such as those of \citet{DBLP:conf/rep4nlp/WegmannSN22} may therefore provide additional views of recurring annotator behavior.
Exploring larger populations together with alternative behavioral representations is a promising direction for future work.

\section{Conclusion}
We studied whether human label variation can be learned as annotator-specific label-explanation behavior rather than only as label disagreement. We showed that such behavior is weak and content-entangled at the single-instance level, but becomes a stable and recoverable signal when aggregated across items. Building on this finding, we proposed CAPO, which improves aggregation-aware annotator-specific modeling and attribution metrics while maintaining competitive decision accuracy across two sentence-pair tasks, showing that variation among annotators can be used as supervision for scalable, explanation-based HLV generation. 
More broadly, our results call for annotator simulation to move beyond single-reference matching toward behavior-level modeling and evaluation, while guarding against over-amplifying recognizable annotator markers.

\clearpage
\section*{Limitations}

\paragraph{Empirical scope and annotator-pool size.}
Our evidence is limited to two English sentence-pair datasets with four recurring annotators each, and we do not assume that annotator separability or CAPO's gains transfer unchanged to larger populations, domains, or tasks.
CAPO may also depend on annotator-pool composition: with more annotators, easy negatives may dilute fine-grained learning, while near-duplicate annotators may produce overly ambiguous contrastive pairs.
Thus, performance need not improve monotonically with pool size.
Difficulty-aware negative selection, such as choosing same-label responses at intermediate explanation-embedding distances, is a promising direction for larger populations.

\paragraph{Data regime and scaling.}
Our experiments use hundreds of training annotations per annotator.
Future work should examine lower-resource regimes to determine when SFT is sufficient and when cross-annotator preference signals provide the largest benefit, as well as larger-scale settings with more annotators, domains, and valid contrastive pairs.
For prompting-based methods, retrieval-, diversity-, or profile-guided selection of annotation histories may further improve few-shot personalization without requiring target-specific adaptation.

\paragraph{Generalization to new annotators.}
Our setup assumes sufficient label--explanation history to train an annotator-specific adapter.
New annotators with only a few examples instead create a cold-start problem.
Although LoRA is parameter-efficient, maintaining a separate adapter for each target causes storage and adaptation costs to grow with the annotator population.
Our shared annotator-ID-conditioned SFT ablation reduces this duplication but yields substantially weaker annotator separation, suggesting that scalable parameter sharing remains an open challenge.
Promising extensions include retrieving or interpolating existing adapters, learning reusable behavioral representations, or combining profile prompting with lightweight parameter updates.
Such approaches could reduce the cost of adapting to evolving annotator pools.

\section*{Ethical Considerations}

Our goal is to model human label variation for analysis, not to impersonate individuals.
Annotator identities should remain anonymized, and annotator-specific models should only be used with appropriate consent and data governance.
Because repeated behavior may reflect sensitive demographic, cultural, ideological, or other personal patterns, these models should not be used to infer private attributes or replace human judgment in high-stakes settings.
Preference optimization may also amplify artifacts; imitation scores above the human-reference scale should therefore be interpreted as increased recognizability rather than more authentic human behavior.
A stronger recognizability evaluation could involve people familiar with the target annotator, following the motivation of IMPersona-style evaluation~\citep{DBLP:journals/corr/abs-2504-04332}.
Such protocols could test whether unseen human and generated responses are recognizable as belonging to the same individual, but would introduce substantially stronger consent and privacy concerns.
We therefore leave them to future work under explicit participant consent and appropriate safeguards.

\paragraph{Reproducibility and responsible release.}
We release all code, prompts, evaluation scripts, and model-training configurations on \href{https://github.com/mainlp/CAPO}{https://github.com/mainlp/CAPO} to support reproducibility. Any released artifacts will preserve annotator anonymity and exclude private or non-redistributable annotation content.

\paragraph{Use of AI Assistants.}
The authors acknowledge the use of ChatGPT solely for correcting grammatical errors, enhancing the coherence of the final manuscript.

\section*{Acknowledgements}
We thank the members of the MaiNLP lab for their insightful feedback on earlier drafts of this paper. 
We specifically appreciate the suggestions of Siyao Peng, Monica Riedler, Shijia Zhou and Xinpeng Wang. 
We are also grateful to the anonymous reviewers for their constructive feedback.
BC acknowledges his membership in the European Laboratory for Learning and Intelligent Systems (ELLIS) PhD program.
PH and BR are funded by the Vienna Science and Technology Fund (WWTF) 10.47379/VRG19008 "Knowledgeinfused Deep Learning for Natural Language Processing".
This research is in parts supported by ERC Consolidator Grant DIALECT 101043235.
AK is supported by the UK Research and Innovation (UKRI) Frontier Research Grant EP/Y031350/1 EQUATE (the UK government's funding guarantee for ERC Advanced Grants).

\appendix

\section{Prompt and Output Formats}
\label{app:prompts}

This appendix records the prompt templates implemented in the experiment scripts.
The prompt-based baselines use the same task templates as the fine-tuned models,
with optional demonstrations or profiles prepended. In the base prompt,
\texttt{Annotator\_\{a\}} is an ungrounded symbolic ID; no examples or profile
are provided.

\subsection{VariErr Task Prompt}

\begin{promptbox}[VariErr task prompt]{PromptBlueFrame}
You are Annotator_{a}.
Follow this annotator's specific
language style and reasoning habits.

Premise:
{premise}

Hypothesis:
{hypothesis}

Please provide the NLI label and
a concise explanation.

Output format:
Label: Entailment, Neutral, or
Contradiction
Explanation: <concise explanation>
\end{promptbox}

For independent SFT and independent CAPO, the target annotator is represented by
the adapter rather than by an explicit ID. The corresponding instruction replaces
the first sentence with:

\begin{promptbox}[Adapter-based VariErr instruction]{PromptGrayFrame}
You are a human NLI annotator.
Follow your learned language style
and reasoning habits.
\end{promptbox}

\subsection{R2 Task Prompt}

\begin{promptbox}[R2 task prompt]{PromptGreenFrame}
You are Annotator_{a}.
Follow this annotator's specific
language style and reasoning habits.

Question 1:
{question1}

Question 2:
{question2}

Assign a Likert score from -5 to 5
indicating how strongly the two
questions are paraphrases of one
another, then provide a concise
explanation.
-5 means definitely not paraphrases,
0 means uncertain or mixed, and
5 means definitely paraphrases.

Output format:
Score: <integer from -5 to 5>
Explanation: <concise explanation>
\end{promptbox}

\subsection{ICL Demonstrations}

ICL prepends 50 target-annotator examples sampled from the train split. Each
example contains the task input without the output-format block, followed by the
target annotator's label or score and explanation.

\begin{promptbox}[ICL demonstration format]{PromptPurpleFrame}
Here are examples of this annotator's
previous labels and explanations.

Example k
Input:
...

Annotator_{a} answer:
Label/Score: ...
Explanation: ...

Now annotate the following item.
...
\end{promptbox}

\subsection{Value Profile Construction}

The direct profile baseline asks the same model family to summarize 50
target-annotator training annotations.

\begin{promptbox}[Value profile construction prompt]{PromptOrangeFrame}
Summarize Annotator_{a}'s annotation
behavior from the following examples.
Focus on label/score bias, value
judgment criteria, evidence usage,
explanation style, wording,
strictness, uncertainty, and quirks.
Write a compact profile that can guide
another LLM to imitate this annotator.
Do not predict labels for the samples.
\end{promptbox}

At inference time, VP and VP-ICL prepend:

\begin{promptbox}[Value profile inference prefix]{PromptOrangeFrame}
Use the following value/style profile
for this annotator when deciding the
label/score and writing the explanation.

Annotator_{a} profile:
{profile}
\end{promptbox}

\subsection{CAPO Format}

\begin{formatbox}[CAPO preference-pair format]{PromptGrayFrame}
CAPO uses the same task prompt as SFT. For each target annotator $a$, the chosen
response is the target annotator's formatted label/score and explanation. The
rejected response is another annotator's formatted label/score and explanation
for the same item.
\end{formatbox}

\section{Implementation Details}
\label{app:implementation}

\subsection{Data Splits}
\label{app:data-splits}

\begin{table}[ht]
\centering
\small
\begin{adjustbox}{max width=\columnwidth}
\begin{tabular}{llrrr}
\toprule
Dataset & Split & Items & Explanations & Complete Items \\
\midrule
\multirow{3}{*}{VariErr}
& train & 300 & 1,055 & 194 \\
& dev & 100 & 400 & 100 \\
& test & 100 & 400 & 100 \\
\midrule
\multirow{3}{*}{R2}
& train & 300 & 1,200 & 300 \\
& dev & 100 & 400 & 100 \\
& test & 100 & 400 & 100 \\
\bottomrule
\end{tabular}
\end{adjustbox}
\caption{Dataset splits used by all methods. VariErr train contains missing annotator explanations for some items; dev and test are complete four-annotator splits.}
\label{tab:split-counts}
\end{table}

Statistics of the two datasets are listed in Table~\ref{tab:split-counts}.
For independent SFT, the number of VariErr train examples per annotator is 257, 265, 251, and 282 for annotators 0--3. R2 has 300 train examples for each annotator. Dev contains 100 examples per annotator for both datasets. VariErr has 1,055 usable training judgments because some train items are missing annotator rows, while dev and test contain four judgments per item. R2 contains 1,200/400/400 train/dev/test judgments. Both datasets are in English.

\subsection{Models and Training Hyperparameters}
\label{app:parameters}

All the hyperparameters are elaborated in Table~\ref{tab:implementation}. Training is conducted on a single NVIDIA A100-SXM4-80GB GPU. SFT training takes an average of 30 minutes per setting, while CAPO training takes an average of 50 minutes per setting.

\begin{table*}[ht]
\centering
\small
\begin{adjustbox}{max width=\textwidth}
\begin{tabular}{ll}
\toprule
Component & Setting \\
\midrule
Base models & \texttt{Qwen3-4B-Instruct-2507}; \texttt{Llama-3.2-3B-Instruct} \\
Prompting decode & temperature 0.0, top-$p$ 1.0, max new tokens 180, max input length 8192 \\
SFT decode & temperature 0.0, top-$p$ 1.0, max new tokens 160, max input length 1536 \\
SFT adapters & LoRA rank 32, alpha 64, dropout 0.05, target modules q/k/v/o/gate/up/down projections \\
SFT optimization & 5 epochs, learning rate $2\times10^{-5}$, batch size 1, gradient accumulation 16, cosine schedule, warmup 0.03 \\
CAPO initialization & Continue training from the corresponding independent SFT adapter \\
CAPO objective & DPO-style loss with $\beta=0.1$, one epoch, learning rate $1\times10^{-6}$ for independent adapters \\
CAPO optimization & batch size 1, gradient accumulation 8, max prompt length 1024, max length 1536, eval/save every 10 steps \\
Checkpoint selection & Main CAPO results use train-only group-classifier selection; dev-loss and last checkpoints are ablations \\
Precision & bf16 with gradient checkpointing; 4-bit loading disabled in the selected runs \\
\bottomrule
\end{tabular}
\end{adjustbox}
\caption{Implementation settings reflected in the training and generation scripts.}
\label{tab:implementation}
\end{table*}

\subsection{CAPO Pair Construction and Pair Counts}
\label{app:capo-pair-counts}

CAPO pairs are ordered same-item pairs. For a target annotator $a$, the chosen response is $a$'s annotation and the rejected response is another annotator's annotation for the same input. Pair builders balance the final training set by target annotator. Table~\ref{tab:capo-pair-counts} reports the number of train pairs after balancing; candidate counts before balancing are included to show how restrictive each policy is.

\begin{table}[ht]
\centering
\small
\begin{adjustbox}{max width=\columnwidth}
\begin{tabular}{llrrrr}
\toprule
Dataset & Policy & Cand. & Train & Min/ann. & Max/ann. \\
\midrule
VariErr & strict & 1,736 & 1,700 & 425 & 425 \\
VariErr & near-label & 2,718 & 2,668 & 667 & 667 \\
VariErr & no restriction & 2,816 & 2,740 & 685 & 685 \\
R2 & strict & 992 & 840 & 210 & 210 \\
R2 & tol1 & 1,874 & 1,476 & 369 & 369 \\
R2 & tol2 & 2,256 & 1,816 & 454 & 454 \\
R2 & tol3 & 2,426 & 2,000 & 500 & 500 \\
R2 & no restriction & 3,600 & 3,600 & 900 & 900 \\
\bottomrule
\end{tabular}
\end{adjustbox}
\caption{CAPO training pair counts for independent pair files. VariErr strict and R2 tol1 are the main policies.}
\label{tab:capo-pair-counts}
\end{table}

VariErr strict requires identical NLI labels; near-label adds entailment--neutral and neutral--contradiction pairs; no restriction allows any cross-annotator label pair. R2 strict requires identical scores, while tol1--tol3 require score distance at most 1--3. All the ablation experimental results are in Appendix~\ref{app:capo-policy-ablation}.

\subsubsection{Theoretical Sensitivity of CAPO to Annotator Pool Size}
\label{app:capo_pool_size}

CAPO constructs preferences from cross-annotator variation on the same input. 
For a target annotator $a$, the chosen response is the target annotator's annotation
$z^+ = z_{i,a}$, while the rejected response is another annotator's annotation
$z^- = z_{i,b}$ for $b \neq a$. Following the DPO-style objective used in the
main method, we write the reward difference relative to the SFT reference model as
\begin{equation}
\begin{aligned}
r_{\theta_a}(z;x_i,a)
&= \log \pi_{\theta_a}(z\mid x_i,a) \\
&\quad - \log \pi_{\mathrm{ref},a}(z\mid x_i,a).
\end{aligned}
\end{equation}

The CAPO loss for annotator $a$ is then
\begin{equation}
\label{eq:capo_pool_loss}
\begin{aligned}
\mathcal{L}_{\mathrm{CAPO}}(a)
= - \mathbb{E}_{(x_i,z^+,z^-)}
\Big[
\log \sigma
\big(
\beta \Delta r_{\theta_a}
\big)
\Big],
\end{aligned}
\end{equation}
where
\begin{equation}
\begin{aligned}
\Delta r_{\theta_a}
&= r_{\theta_a}(z^+;x_i,a) \\
&\quad - r_{\theta_a}(z^-;x_i,a).
\end{aligned}
\end{equation}

When the annotator pool size is $N$, each input provides $N-1$ possible rejected
responses for each target annotator. The empirical CAPO objective can therefore
be approximated as
\begin{equation}
\label{eq:capo_pool_average}
\mathcal{L}_{\mathrm{CAPO}}(a)
\approx
\frac{1}{|\mathcal{D}_a|}
\sum_{x_i\in\mathcal{D}_a}
\ell_i(a),
\end{equation}
with the per-input loss
\begin{equation}
\label{eq:capo_item_loss}
\begin{aligned}
\ell_i(a)
= - \frac{1}{N-1}
& \sum_{b\neq a}
\log \sigma
\Big(
\beta \big(
r_{\theta_a}(z_{i,a};x_i,a) \\
& - r_{\theta_a}(z_{i,b};x_i,a)
\big)
\Big).
\end{aligned}
\end{equation}

This formulation exposes three ways in which CAPO depends on the annotator pool
size.

\paragraph{Variance and bias of the contrastive signal.}
When $N$ is very small, each target annotation has only a few possible rejected
responses. In the extreme case of $N=2$, each input provides only one rejected
candidate. The resulting training signal may have high variance: if the only
available rejected annotator happens to produce an annotation that is very close
to the target annotator on a particular item, the pair provides little useful
discriminative signal. As $N$ increases, the rejected pool becomes larger and
the empirical estimate in Eq.~\ref{eq:capo_item_loss} becomes more stable.
However, increasing $N$ also changes the distribution of rejected candidates,
which affects the strength and type of contrastive signal.

\paragraph{Annotator similarity and contrastive margin.}
Let the reward margin for a target annotator $a$ against another annotator $b$ be
\begin{equation}
\begin{aligned}
m_{i,a,b}
&= r_{\theta_a}(z_{i,a};x_i,a) \\
&\quad - r_{\theta_a}(z_{i,b};x_i,a).
\end{aligned}
\end{equation}
CAPO is most informative when rejected responses are different enough from the
target response to create a meaningful preference, but not so different that the
comparison becomes trivial. With a small curated annotator pool, such as the
$N=4$ setting used in our experiments, annotators may share substantial task
understanding and explanation conventions. This can make the margin smaller,
requiring CAPO to learn fine-grained target-specific differences. With a larger
pool, the rejected set is more diverse and may contain easier negatives whose
style is clearly different from the target annotator. These easy negatives can
increase the apparent margin, but they may contribute less to learning subtle
explanation behavior. Conversely, a very large pool may also contain
near-duplicate annotators whose explanations are almost indistinguishable from
the target annotator's explanations. Treating these near-duplicates as rejected
responses may introduce noisy or overly strict penalties against reasonable
human variants.

These effects suggest a non-monotonic relationship between $N$ and the relative
benefit of CAPO over SFT. With extremely small pools, the signal may be too
high-variance. With moderately sized pools, CAPO can obtain useful contrasts
while still focusing on target-specific behavior. With very large pools, easy
negatives and near-duplicate negatives may respectively dilute or corrupt the
contrastive signal unless the pair construction strategy is adapted.

\paragraph{Scalability of strict pair construction.}
For VariErr, our main CAPO setting uses a strict same-label policy:
\begin{equation}
y_{i,a} = y_{i,b}.
\end{equation}
This policy reduces the risk that CAPO learns only label-level differences,
making the contrast more explanation-focused. If label categories are denoted by
$c\in\mathcal{Y}$, the expected fraction of same-label annotator pairs is
\begin{equation}
p_{\mathrm{same}}
=
\sum_{c\in\mathcal{Y}} P(c)^2.
\end{equation}
Thus, the expected number of same-label pairs for an item is
\begin{equation}
\mathbb{E}\!\left[
\#\mathrm{pairs}_{\mathrm{same}}
\right]
=
\binom{N}{2}
p_{\mathrm{same}}.
\end{equation}
Using the approximate VariErr label proportions from Figure~\ref{fig:varierr-label-variation}(b), this fraction
is
\begin{equation}
\begin{aligned}
p_{\mathrm{same}}
&\approx 0.31^2 + 0.52^2 + 0.17^2 \\
&\approx 0.40.
\end{aligned}
\end{equation}
Therefore, about $40\%$ of cross-annotator pairs satisfy the same-label
constraint under this approximation. As $N$ grows, the total number of
same-label pairs grows quadratically in $N$, while the number of available
same-label rejected candidates per target annotation grows approximately
linearly in $N$. This means that the strict policy itself is not a scalability
bottleneck.

However, larger pools also increase the chance that same-label rejected
responses include either overly easy negatives or near-duplicate negatives.
A natural extension is therefore to introduce difficulty-aware rejected-candidate
sampling. Instead of sampling uniformly from all valid rejected responses, CAPO
could prioritize same-label responses whose explanation embeddings are at an
intermediate distance from the target response. This would avoid contrasts that
are either too easy to be useful or too similar to be treated as reliable
negative examples. Such a strategy is analogous to curriculum learning or
hard-negative mining, and we leave it to future work.

\subsection{Group Classifier Details}
\label{app:group-size-sweep}

Group classifiers are used for two separate purposes: CAPO checkpoint selection and final imitation evaluation. These uses are kept separate to avoid test leakage. For each method and target annotator, we form bootstrapped groups of generated outputs and feed their aggregated representations to the classifier. We report GC confidence, the mean probability assigned to the intended annotator, which captures the strength of the attribution and is more sensitive when several methods already achieve high accuracy.

\paragraph{Checkpoint-selection classifier.}
For CAPO checkpoint selection, we train the group classifier only on human training annotations. For each candidate CAPO checkpoint, we generate outputs on the development set, construct groups by target annotator, and select the checkpoint with the highest mean target-annotator confidence. This classifier is used only for model selection and never sees test annotations or test generations.

\paragraph{Evaluation classifier.}
For final reporting, we train a separate group classifier on human annotations from train, development, and test splits. This classifier is used only as an evaluation instrument for generated test outputs. It is not used to select CAPO checkpoints. We report both group classifier accuracy and confidence. Accuracy is the percentage of bootstrapped groups whose predicted annotator matches the intended target annotator. Confidence is the average probability assigned to the intended annotator, regardless of whether it is the top prediction.

\paragraph{Classifier configuration.}
Unless otherwise stated, group classifiers use group size 20. We use fixed logistic-regression regularization $C=21.544347$ so that confidence values are comparable across group sizes and settings. The group-size sweep shows that group size 20 is large enough to make human annotator identity reliably recoverable on both datasets, while remaining small enough to support stable bootstrapping on generated outputs. 

\paragraph{Group-size sweep.}
Table~\ref{tab:gc-selection-sweep} and~\ref{tab:gc-final-sweep} report the group-size sweep for the train-only and train+dev+test classifiers. The main paper uses group size 20 because it yields near-perfect human group classification on both datasets, supporting the aggregation-aware evaluation protocol.

\begin{table}[ht]
\centering
\small
\begin{adjustbox}{max width=\columnwidth}
\begin{tabular}{lrrrr}
\toprule
Dataset & Group & Test Acc & Test Conf & Test Groups \\
\midrule
\multirow{4}{*}{VariErr}
& 5 & 0.838 & 0.780 & 400 \\
& 10 & 0.950 & 0.834 & 200 \\
& 20 & 0.990 & 0.889 & 100 \\
& 30 & 0.988 & 0.915 & 80 \\
\midrule
\multirow{4}{*}{R2}
& 5 & 0.998 & 0.978 & 400 \\
& 10 & 1.000 & 0.985 & 200 \\
& 20 & 1.000 & 0.990 & 100 \\
& 30 & 1.000 & 0.990 & 80 \\
\bottomrule
\end{tabular}
\end{adjustbox}
\caption{Train-only group-classifier sweep used for CAPO checkpoint selection. The classifier is trained on human training annotations and evaluated on held-out human test annotations.}
\label{tab:gc-selection-sweep}
\end{table}

\begin{table}[ht]
\centering
\small
\begin{adjustbox}{max width=\columnwidth}
\begin{tabular}{lrrrr}
\toprule
Dataset & Group & Test Acc & Test Conf & Test Groups \\
\midrule
\multirow{4}{*}{VariErr}
& 5 & 0.950 & 0.871 & 400 \\
& 10 & 0.985 & 0.921 & 200 \\
& 20 & 1.000 & 0.951 & 100 \\
& 30 & 1.000 & 0.961 & 80 \\
\midrule
\multirow{4}{*}{R2}
& 5 & 1.000 & 0.982 & 400 \\
& 10 & 1.000 & 0.987 & 200 \\
& 20 & 1.000 & 0.991 & 100 \\
& 30 & 1.000 & 0.992 & 80 \\
\bottomrule
\end{tabular}
\end{adjustbox}
\caption{Train+dev+test group-classifier sweep. The group-size-20 classifiers from this setting are used only for final reported GC confidence and imitation score.}
\label{tab:gc-final-sweep}
\end{table}

\subsubsection{Robustness to Explicit Surface Markers}
\label{app:marker_residual}

The group classifier operates on aggregated $E_4$ representations rather than directly on the handcrafted features.
Nevertheless, $E_4$ may encode explicit signals such as explanation length, punctuation, quotation, or first-person usage.
We therefore test whether annotator recoverability remains after removing components of $E_4$ that are linearly predictable from these observable markers.

Using only human training annotations, we fit a multi-output ridge regression
\begin{equation}
    \widehat{E}_4 = g_{\phi}(\mathbf{s}),
\end{equation}
where $\mathbf{s}$ denotes the corresponding handcrafted feature vector.
For held-out human annotations and generated outputs, we then compute
\begin{equation}
    E_{4,\mathrm{res}}
    =
    E_4 - \widehat{E}_4(\mathbf{s}).
\end{equation}

We consider two variants.
\textbf{Surface-marker residualization} removes components predictable from length, sentence structure, type--token ratio, punctuation, quotation, first-person usage, and simple negation/causal/conditional markers.
\textbf{All-feature residualization} additionally removes components predictable from modality, hedging, certainty, metadiscourse, judgment/contrast markers, and input-overlap features.

\begin{table}[t]
\centering
\small
\begin{adjustbox}{max width=\columnwidth}
\begin{tabular}{lccc}
\toprule
Dataset &
Original $E_4$ &
Surface residual &
All-feature residual \\
\midrule
VariErr & 0.98 & 0.98 & 0.98 \\
R2      & 1.00 & 1.00 & 1.00 \\
\bottomrule
\end{tabular}
\end{adjustbox}
\caption{
Human group-classifier accuracy under explicit-marker residualization.
Annotator identity remains almost perfectly recoverable even after removing embedding components linearly predictable from the handcrafted features.
}
\label{tab:marker_residual}
\end{table}

Results are in Table \ref{tab:marker_residual}.
The stability of human group-classifier accuracy shows that aggregate annotator identity is not reducible to the explicit handcrafted markers.
For generated outputs, some CAPO--SFT differences weaken under the all-feature residualization, particularly on R2.
This result is also informative: explicit stylistic regularities constitute part of the joint behavior being modeled.
Accordingly, we use GC as a diagnostic of repeated label--explanation behavior that combines semantic framing and recurring writing tendencies, rather than interpreting it as a pure measure of deep or content-independent reasoning style.

\subsection{LLM-as-Judge Details}
\label{app:judge}

The external judge is DeepSeek V4 Pro, called with temperature 0.0 and JSON-object output. For each generated judgment, the judge receives the original item, the generated label/score and explanation, and the four human gold candidates for the same item in a deterministic shuffled order. The main metric is 1-vs-4 attribution accuracy against the intended target annotator. The parse rate is 1.0.

The judge is deliberately instance-level, whereas the group classifier is aggregation-level. This makes the judge a useful but stricter complement: on VariErr, even strong fine-tuned systems remain only moderately above chance because the four candidates often discuss the same premise-hypothesis evidence; on R2, the scoring and explanation habits are more directly recognizable.

\paragraph{LLM-as-Judge Prompt}
The judge sees the item, four anonymized shuffled human candidates for the same
item, and the generated output.

\begin{promptbox}[LLM-as-judge prompt]{PromptBlueFrame}
Four original gold annotations,
shown in random anonymous order:
Option A:
Label/Score: ...
Explanation: ...
...

Predicted output:
Label/Score: ...
Explanation: ...

Task:
Choose which anonymous gold option the
predicted output is most similar to.
Use label agreement, explanation
meaning, reasoning style, wording,
specificity, and annotation habits.

Return JSON only:
{"choice": "A" | "B" | "C" | "D"}.
\end{promptbox}

\subsubsection{Candidate-Order Robustness}
\label{app:judge_order}

LLM-as-judge attribution can in principle be affected by candidate position.
We therefore conduct a balanced cyclic-order robustness analysis.
Each selected generated output is evaluated under all four cyclic permutations of the four human candidates, placing the intended annotator equally often in positions A, B, C, and D.
The analysis covers both datasets, both 3--4B model families, and both SFT and CAPO.

\begin{table}[t]
\centering
\small
\begin{adjustbox}{max width=\columnwidth}
\begin{tabular}{lr}
\toprule
Statistic & Result \\
\midrule
Successfully parsed judge calls & $3{,}200/3{,}200$ \\
Order-specific accuracy range & $0.01$--$0.08$ \\
Typical range & $0.02$--$0.06$ \\
CAPO $\geq$ SFT comparisons & $14/16$ \\
Small SFT-over-CAPO reversals & $2/16$ \\
Magnitude of the two reversals & $0.01$--$0.02$ \\
VariErr--Llama mean CAPO--SFT difference & $+0.0475$ \\
95\% CI for VariErr--Llama difference & $[0.0225,\,0.0750]$ \\
Selected-position association, Cramer's $V$ & $0.050$ \\
\bottomrule
\end{tabular}
\end{adjustbox}
\caption{
Balanced cyclic-order robustness of LLM-as-judge attribution.
Candidate order has only a small effect and does not materially change the SFT--CAPO comparison.
}
\label{tab:judge_order}
\end{table}

Results are in Table \ref{tab:judge_order}.
Candidate-order variation is small relative to the qualitative method comparison.
CAPO is at least tied with SFT in 14 of the 16 dataset--model--order comparisons; the only two reversals are R2 differences of only $0.01$--$0.02$.
For VariErr--Llama, CAPO is higher under all four orders, with a paired mean difference of $+0.0475$ and a 95\% confidence interval of $[0.0225,0.0750]$.
Although the aggregate selected-letter distribution is not perfectly uniform, its association with position is very small ($V=0.050$).
We therefore retain Judge Accuracy as an independent item-level complement to the aggregation-aware metrics, rather than as a metric carrying the main conclusion by itself.
\section{Additional Stable Variation Analysis}
\label{app:stable}

This appendix gives the full diagnostic details behind Section~\ref{sec:stable}: the feature inventory, single-annotation classifier results with confusion matrices, group-level classifier sweeps, and the parallel R2 analysis. All classifier splits are item-level splits, so no test item appears in classifier training.

\subsection{Explanation-Style Feature Inventory}
\label{app:feature-inventory}

Table~\ref{tab:feature-inventory} lists the handcrafted explanation-style features used in the heatmaps and feature classifiers. These features are intentionally shallow: they are meant to capture repeated writing and justification habits that may be washed out in semantic embeddings.

\begin{table*}[t]
\centering
\small
\begin{adjustbox}{max width=\textwidth}
\begin{tabular}{p{0.18\textwidth}p{0.30\textwidth}p{0.44\textwidth}}
\toprule
Feature family & Feature names & How the feature is computed \\
\midrule
Length and structure &
\texttt{len\_chars}, \texttt{len\_words}, \texttt{num\_sentences}, \texttt{avg\_sent\_len}, \texttt{type\_token\_ratio} &
The explanation is lowercased and tokenized with the regex \texttt{\textbackslash b\textbackslash w+\textbackslash b}. Characters are counted on the raw explanation; words are the regex tokens; sentences are non-empty spans split by \texttt{[.!?]+}. \texttt{avg\_sent\_len} is tokens divided by sentences, and \texttt{type\_token\_ratio} is unique tokens divided by tokens. Empty denominators are clipped to 1. \\
Lexical stance markers &
\texttt{modal\_rate}, \texttt{hedge\_rate}, \texttt{certain\_rate}, \texttt{meta\_rate}, \texttt{judge\_rate} &
Each value is a dictionary-hit rate over explanation tokens. Modal words are \texttt{must}, \texttt{should}, \texttt{could}, \texttt{would}, \texttt{might}, \texttt{may}, \texttt{can}, \texttt{need}, \texttt{shall}, \texttt{will}; hedge words include \texttt{possibly}, \texttt{perhaps}, \texttt{probably}, \texttt{likely}, \texttt{seems}, \texttt{appears}, \texttt{suggests}, \texttt{implies}, \texttt{unclear}, \texttt{ambiguous}; certainty words include \texttt{clearly}, \texttt{obviously}, \texttt{definitely}, \texttt{certainly}, \texttt{necessarily}, \texttt{always}, \texttt{never}, \texttt{impossible}. \texttt{meta\_rate} and \texttt{judge\_rate} use dataset-specific dictionaries: VariErr uses premise/hypothesis/NLI vocabulary, while R2 uses question/paraphrase vocabulary. \\
Reasoning-connective rates &
\texttt{has\_negation}, \texttt{has\_because}, \texttt{has\_conditional}, \texttt{uses\_first\_person} &
Binary indicators. \texttt{has\_negation} checks for \texttt{not}, \texttt{no}, \texttt{never}, \texttt{neither}, \texttt{nor}; \texttt{has\_because} checks for any causal word; \texttt{has\_conditional} checks for \texttt{if}, \texttt{unless}, \texttt{only}; \texttt{uses\_first\_person} checks for \texttt{i}, \texttt{my}, \texttt{me}. \\
Punctuation and quotation &
\texttt{quote\_count}, \texttt{question\_mark}, \texttt{parentheses}, \texttt{has\_direct\_quote} &
\texttt{quote\_count} is the number of single plus double quote characters in the raw explanation. \texttt{question\_mark} and \texttt{parentheses} are binary indicators for the presence of \texttt{?} and either parenthesis. \texttt{has\_direct\_quote} is 1 if any 15-character lowercase substring from the concatenated input texts, sampled every 5 characters, appears in the explanation. \\
Input reuse &
\texttt{text\_a\_overlap}, \texttt{text\_b\_overlap}; these correspond to premise/hypothesis for VariErr and question 1/question 2 for R2 &
After removing a small stopword list from the input and explanation token sets, each feature is the fraction of non-stopword tokens from the corresponding input side that also appear in the explanation: $|\texttt{expl}\cap\texttt{text\_a}|/\max(|\texttt{text\_a}|,1)$ and analogously for \texttt{text\_b}. \\
\bottomrule
\end{tabular}
\end{adjustbox}
\caption{Handcrafted explanation-style features.}
\label{tab:feature-inventory}
\end{table*}

\subsubsection{Literature Grounding for Handcrafted Explanation Features}
\label{app:feature_sources}

The handcrafted feature set used in our Feature KL analysis is intended to capture repeated, interpretable aspects of how annotators write explanations, rather than the item content alone. Although the exact combination of features and the Feature KL aggregation are specific to our setting, the individual feature families are grounded in prior work on distributional text evaluation, stylometry, hedging and uncertainty, NLI artifacts, and explanation-based analyses of human label variation.

At the metric level, Feature KL follows the general idea of distribution-level comparison between generated and human text. MAUVE compares human and machine text distributions using information divergences in a quantized embedding space \citep{pillutla-etal-2021-mauve}. Our Feature KL can be viewed as an interpretable counterpart: instead of comparing distributions in a learned embedding space, we compare distributions over linguistically meaningful handcrafted feature bins. This makes the resulting shift easier to inspect feature by feature.

Table~\ref{tab:feature_source_grounding} summarizes the provenance of each feature. Length and lexical-diversity features are standard stylometric cues \citep{stamatatos2009survey,koppel2009computational} and have also been used in recent comparisons of human and LLM-generated text, where human text was found to exhibit more scattered sentence length distributions and richer vocabulary \citep{munozortiz2024contrasting}. Type-token ratio is further connected to lexical diversity and entropy in large corpora \citep{rosillo-rodes-etal-2024-entropy}. Modal, hedge, certainty, meta-reasoning, and judgment markers are motivated by work on uncertainty and disagreement: the CoNLL-2010 shared task formalized hedge cue detection \citep{farkas-etal-2010-conll}, and NLI disagreement work shows that annotators use uncertainty, presupposition, and pragmatic reasoning to justify different interpretations \citep{DBLP:journals/tacl/JiangM22}. Recent explanation taxonomies for NLI also explicitly model uncertainty-preserving and reasoning-type distinctions \citep{hong-etal-2025-litex}.

Reasoning-connective and negation features are grounded in NLI-specific work. Negation and vagueness are known annotation artifacts in NLI \citep{gururangan-etal-2018-annotation}, and lexical overlap has been shown to induce strong NLI heuristics \citep{mccoy-etal-2019-right,rajaee-etal-2022-looking}. Explanation datasets such as LiveNLI and VariErr further motivate studying how annotators express causal, conditional, and evidence-based reasoning in free-text explanations \citep{jiang-etal-2023-ecologically,weber-genzel-etal-2024-varierr}. Input-word reuse features similarly follow the NLI overlap-bias literature and explanation-based analyses of whether annotators copy or rephrase premise and hypothesis evidence. Finally, punctuation, quotation, parenthesis use, and first-person markers are classic surface-style and authorship-profiling cues \citep{stamatatos2009survey,koppel2009computational,argamon-etal-2009-profiling,preotiuc-pietro-etal-2017-beyond}. We therefore use these features not as arbitrary correlates, but as a compact set of interpretable cues with precedent in stylometry, NLI bias analysis, and explanation-grounded HLV research.

\begin{table*}[t]
\centering
\small
\resizebox{0.8\textwidth}{!}{
\begin{tabular}{p{0.20\linewidth} p{0.28\linewidth} p{0.44\linewidth}}
\toprule
\textbf{Feature group} & \textbf{Features} & \textbf{Literature grounding} \\
\midrule
Length and lexical diversity &
\texttt{len\_chars}, \texttt{len\_words}, \texttt{num\_sentences}, \texttt{avg\_sent\_len}, \texttt{type\_token\_ratio} &
Sentence length, text length, and lexical diversity are standard stylometric features \citep{stamatatos2009survey,koppel2009computational}. Human--LLM comparison work reports systematic differences in sentence-length distributions and vocabulary diversity \citep{munozortiz2024contrasting}. TTR is also a standard lexical-diversity measure connected to entropy in large corpora \citep{rosillo-rodes-etal-2024-entropy}. \\
\midrule
Epistemic stance and judgment markers &
\texttt{modal\_rate}, \texttt{hedge\_rate}, \texttt{certain\_rate}, \texttt{meta\_rate}, \texttt{judge\_rate} &
Hedge and uncertainty cues have a long precedent in hedge detection, including the CoNLL-2010 shared task \citep{farkas-etal-2010-conll}. NLI disagreement analyses show that uncertainty, pragmatic interpretation, and annotator reasoning strategies are central to label variation \citep{DBLP:journals/tacl/JiangM22}. LiTEx further formalizes explanation categories for NLI reasoning variation \citep{hong-etal-2025-litex}. \\
\midrule
Reasoning connectives and negation &
\texttt{has\_because}, \texttt{has\_conditional}, \texttt{has\_negation} &
Causal and conditional markers capture explicit reasoning structure in explanations, as studied in LiveNLI and explanation-based HLV work \citep{jiang-etal-2023-ecologically,chen-etal-2025-threading}. Negation is a known NLI artifact and surface cue correlated with inference labels \citep{gururangan-etal-2018-annotation}; related NLI challenge-set work studies lexical and syntactic heuristics \citep{mccoy-etal-2019-right}. \\
\midrule
Input-word reuse and quotation from input &
\texttt{premise\_overlap}, \texttt{hypo\_overlap}, \texttt{has\_direct\_quote} &
Premise--hypothesis word overlap is a central cue in NLI bias and heuristic analyses \citep{mccoy-etal-2019-right,rajaee-etal-2022-looking}. VariErr motivates studying explanation text together with labels to separate plausible variation from error \citep{weber-genzel-etal-2024-varierr}. The direct-quote feature operationalizes a more literal form of input reuse; its exact 15-character matching rule is heuristic and should be treated as a feature-engineering choice rather than a previously standardized definition. \\
\midrule
Punctuation and parenthetical style &
\texttt{quote\_count}, \texttt{question\_mark}, \texttt{parentheses} &
Punctuation and symbol-use patterns are classic stylometric and authorship-attribution features \citep{stamatatos2009survey,koppel2009computational}. Human--LLM linguistic comparison work also reports differences in symbol and punctuation-related usage \citep{munozortiz2024contrasting}. \\
\midrule
First-person use &
\texttt{uses\_first\_person} &
Pronoun and self-reference features are common in authorship profiling and social-media style analysis \citep{argamon-etal-2009-profiling,preotiuc-pietro-etal-2017-beyond}. In explanation settings, first-person use can indicate how explicitly an annotator frames the judgment as their own interpretation. \\
\bottomrule
\end{tabular}}
\caption{Literature grounding for the 20 handcrafted explanation-style features used in Feature KL. The feature set combines standard stylometric cues, NLI-specific overlap and negation cues, and explanation-oriented markers of uncertainty and reasoning style.}
\label{tab:feature_source_grounding}
\end{table*}

The table also clarifies the limits of the feature design. Most features have direct precedent as stylometric, uncertainty, NLI-bias, or explanation-analysis cues. The main exception is \texttt{has\_direct\_quote}: while it is motivated by input reuse and quotation behavior, the 15-character matching threshold is a heuristic operationalization. We therefore treat Feature KL as a diagnostic measure rather than a fully validated psycholinguistic scale, and future work could replace the unweighted average with a reliability- or discriminativeness-weighted aggregation, for example using feature-wise annotator-separation statistics.

\subsubsection{Within-Item Permutation Baseline}
\label{app:permutation}

We test whether the annotator-level feature differences in Section~\ref{sec:stable_explanations} exceed variation expected from item composition alone.
For each item, we keep the observed explanations and labels/scores fixed and randomly permute annotator identities among the available responses.
The procedure therefore preserves the input, explanation texts, and within-item decision composition while destroying persistent correspondence between an annotator identity and repeated behavior across items.
For every permutation, we recompute the annotator effect for each explanation feature and summarize the resulting effect sizes.
We use 5,000 permutations.
Table~\ref{tab:permutation} compares the observed mean effect size with the resulting permutation null.

\begin{table*}[t]
\centering
\small
\begin{tabular}{lccc}
\toprule
Dataset &
Observed mean $\eta^2$ &
Permutation-null mean &
Empirical $p$ \\
\midrule
VariErr & 0.0385 & 0.00645 & 0.00020 \\
R2      & 0.1291 & 0.00745 & 0.00020 \\
\bottomrule
\end{tabular}
\caption{
Within-item permutation test for annotator-specific explanation features.
Random reassignment preserves item content and the observed responses while removing persistent annotator identity.
Observed feature-level variation is substantially larger than the permutation null on both datasets.
}
\label{tab:permutation}
\end{table*}

The result provides a null-model interpretation for the feature heatmaps:
the effects of individual features are generally small, but their systematic annotator-level structure is substantially stronger than expected from content-matched random groupings.
We therefore interpret the handcrafted features as evidence of repeated observable behavioral and stylistic regularities, rather than as individually identifying markers.
After Benjamini--Hochberg correction across the feature-wise tests,
15/20 features on VariErr and 18/20 on R2 remain above the within-item null.

\subsection{Additional VariErr Embedding Figures and Results}
\label{app:varierr-extra}

\begin{figure*}[t]
\centering
\begin{minipage}{0.32\textwidth}
\centering
\includegraphics[width=\linewidth]{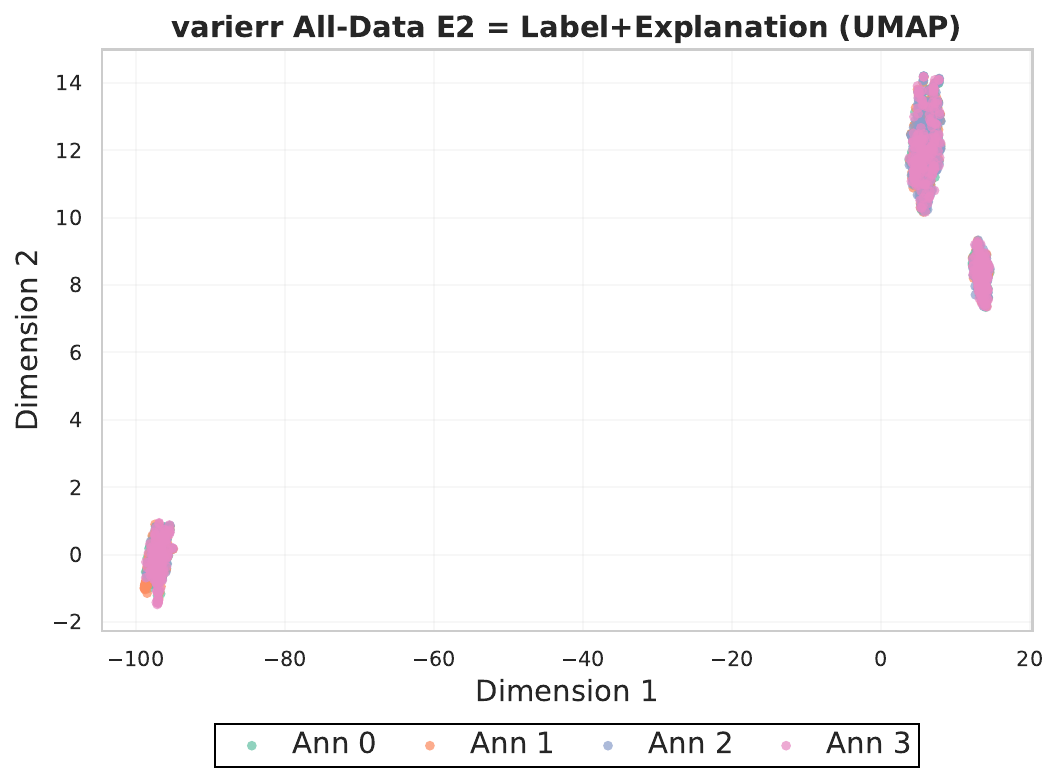}
{\footnotesize (a) $E_2$}
\end{minipage}
\hfill
\begin{minipage}{0.32\textwidth}
\centering
\includegraphics[width=\linewidth]{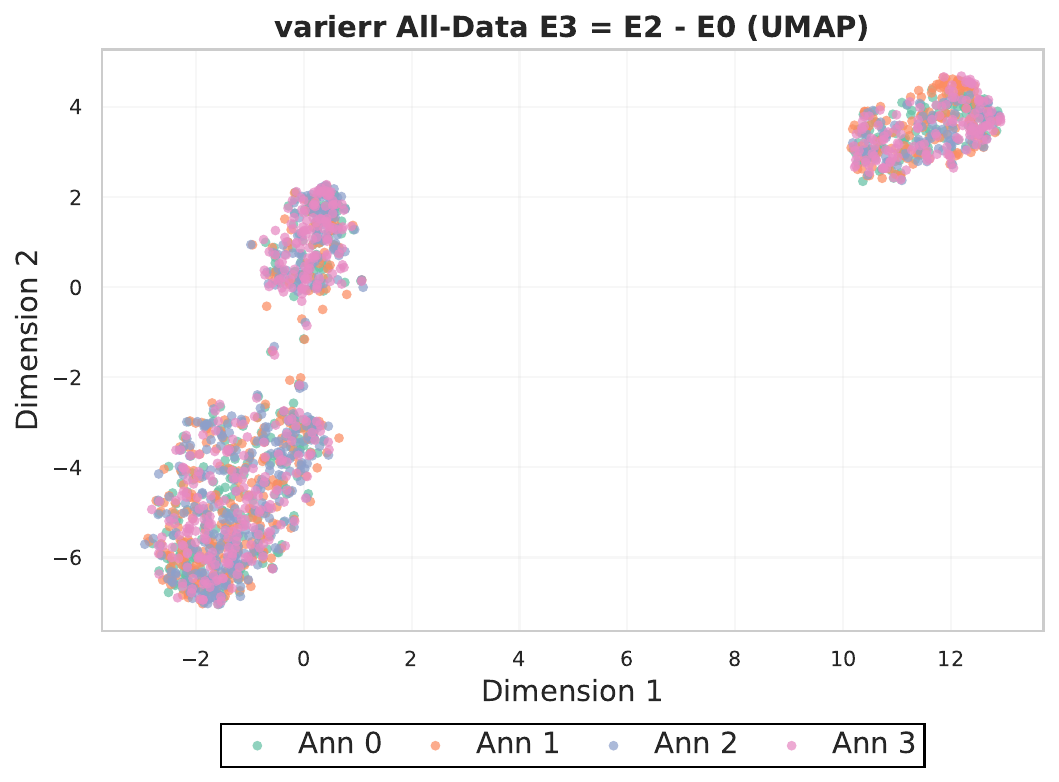}
{\footnotesize (b) $E_3$}
\end{minipage}
\hfill
\begin{minipage}{0.32\textwidth}
\centering
\includegraphics[width=\linewidth]{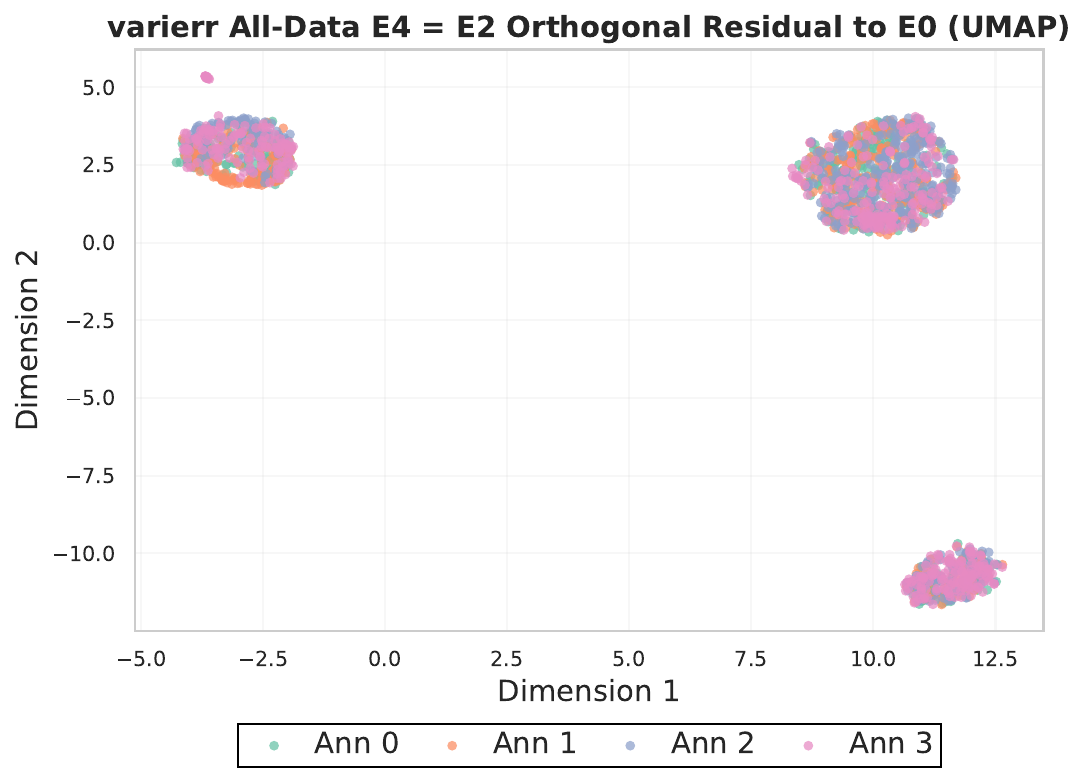}
{\footnotesize (c) $E_4$}
\end{minipage}
\caption{VariErr individual-annotation UMAPs for content-reduced explanation representations.}
\label{fig:app-varierr-single-umaps}
\end{figure*}

\begin{figure*}[t]
\centering
\includegraphics[width=\textwidth]{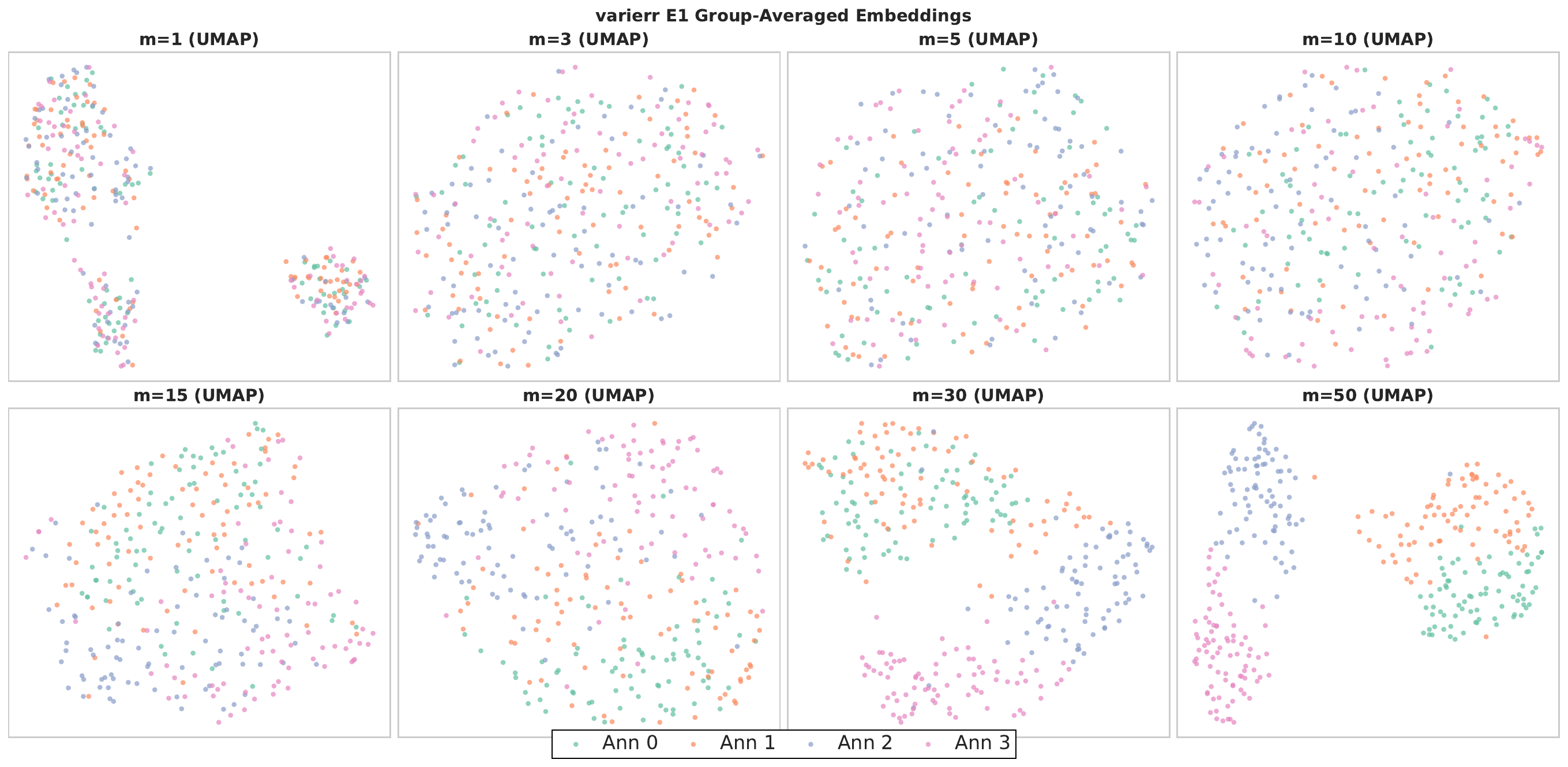}
\caption{VariErr group-averaged $E_1$ embeddings. Raw context-containing embeddings become more separable with aggregation, but remain weaker than content-reduced representations.}
\label{fig:app-varierr-e1-group-umap}
\end{figure*}

\begin{figure*}[t]
\centering
\includegraphics[width=\textwidth]{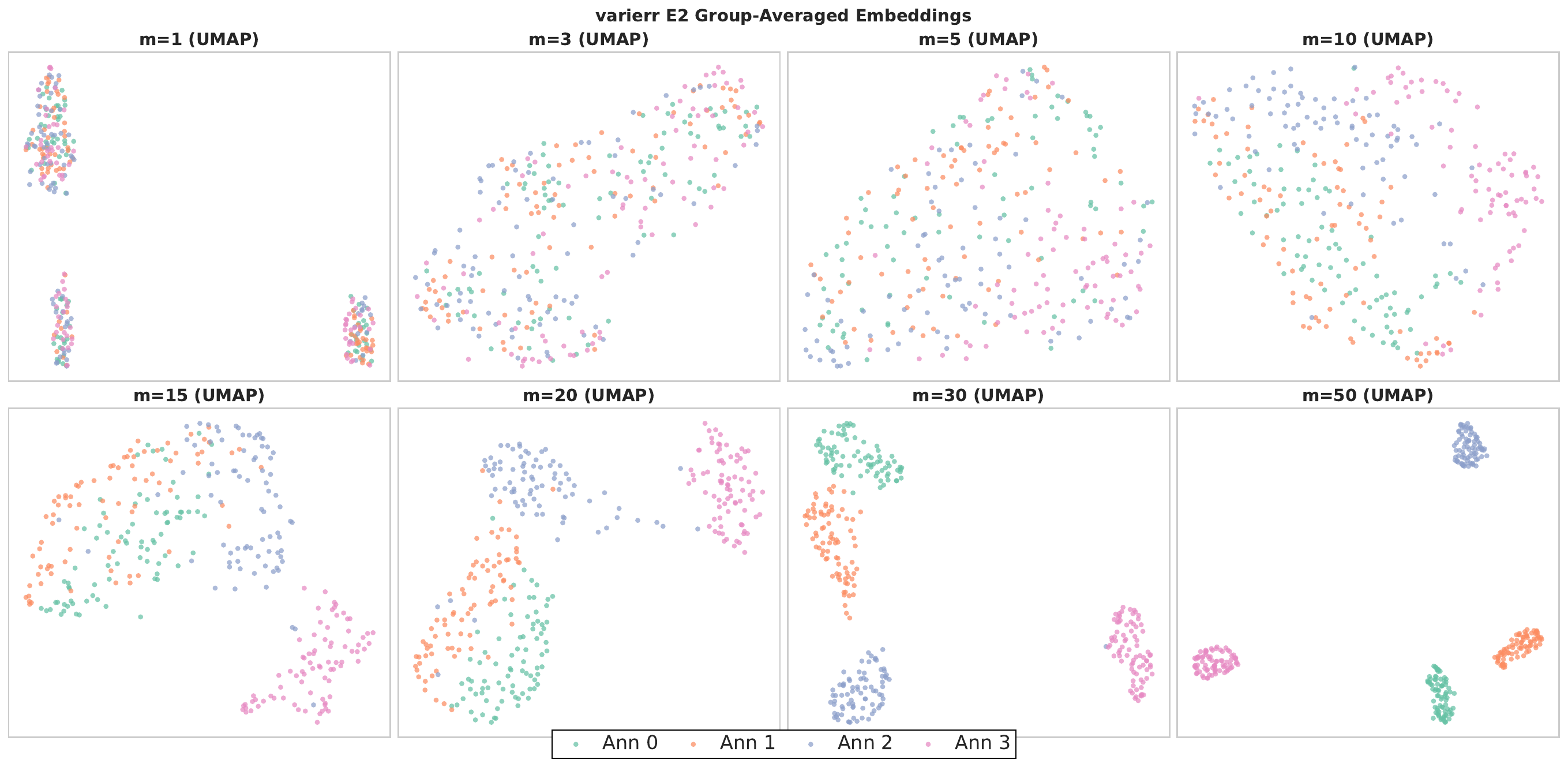}
\caption{VariErr group-averaged $E_2$ embeddings. Removing the input text from the embedded string improves group-level separability.}
\label{fig:app-varierr-e2-group-umap}
\end{figure*}

\begin{figure*}[t]
\centering
\includegraphics[width=\textwidth]{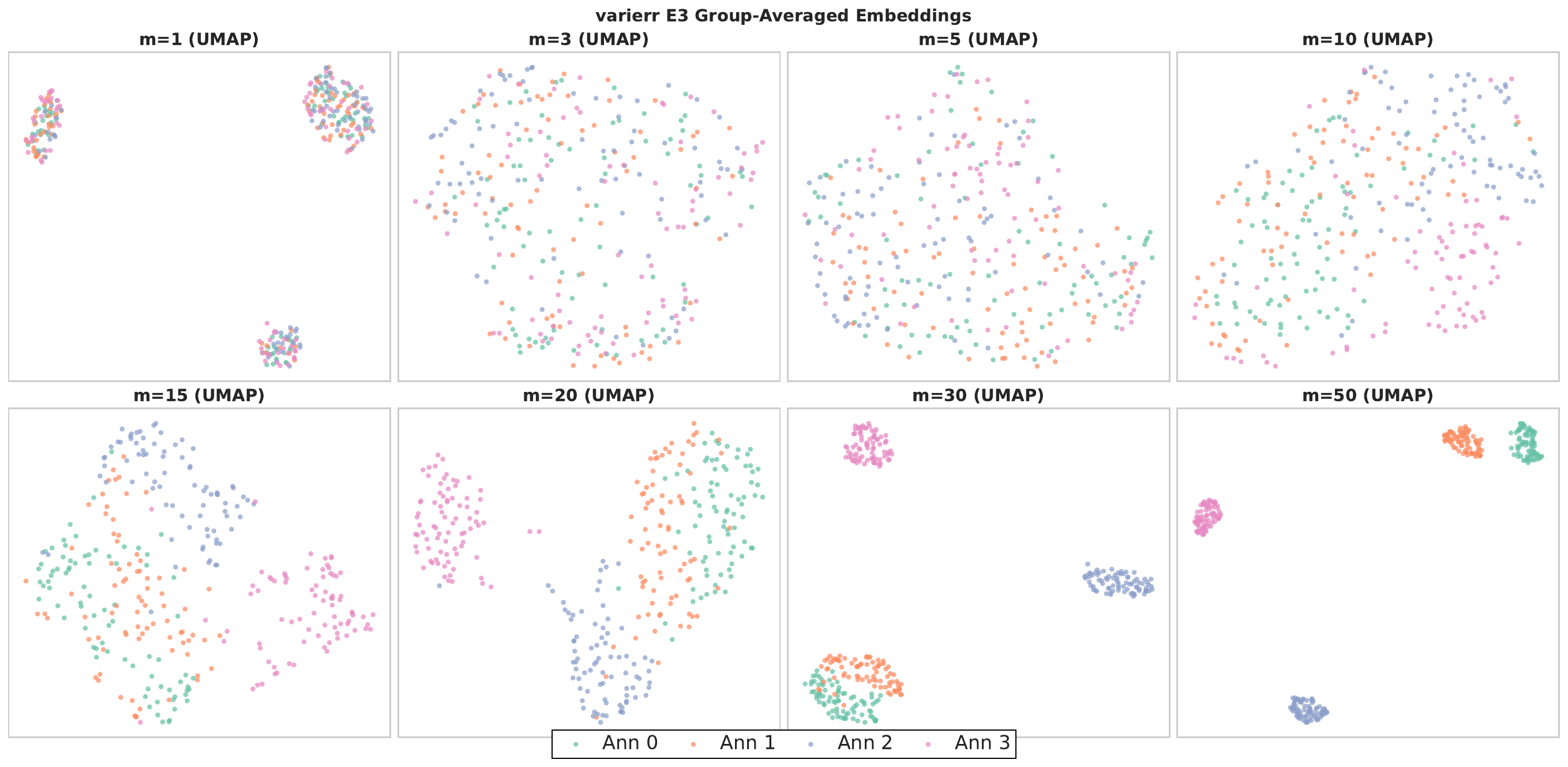}
\caption{VariErr group-averaged $E_3$ embeddings. Directly subtracting the input embedding gives a stronger group-level signal than raw $E_1$.}
\label{fig:app-varierr-e3-group-umap}
\end{figure*}

Figure~\ref{fig:app-varierr-single-umaps} shows VariErr UMAPs for $E_2$, $E_3$, and $E_4$ at the individual-annotation level. Compared with $E_1$ in Figure~\ref{fig:varierr-feature-umap}, removing or reducing the input component makes annotator structure more visible, but the clusters remain noisy. Figures~\ref{fig:app-varierr-e1-group-umap}--\ref{fig:app-varierr-e3-group-umap} show the corresponding group-averaged views for $E_1$--$E_3$; the main text shows $E_4$ in Figure~\ref{fig:varierr-e4-group-umap}.

\paragraph{Single-Annotation Classifier Details}

Tables~\ref{tab:varierr-single-classifier} and~\ref{tab:r2-single-classifier} report single-annotation annotator classifiers. The split uses 250 train items and 250 test items. VariErr has 927 train annotations and 928 test annotations because the train data has missing annotator rows; R2 has 1,000 train and 1,000 test annotations. Rows in each confusion matrix are true annotators and columns are predicted annotators, ordered 0--3.

\begin{table*}[t]
\centering
\scriptsize
\begin{adjustbox}{max width=\textwidth}
\begin{tabular}{lcc}
\toprule
Representation & Acc. & Confusion matrix \\
\midrule
Features & 50.2 &
$\begin{bmatrix}104&47&38&41\\38&102&53&40\\35&29&103&57\\23&29&32&157\end{bmatrix}$ \\
$E_1$ & 41.9 &
$\begin{bmatrix}115&35&48&32\\69&54&79&31\\32&30&116&46\\27&36&74&104\end{bmatrix}$ \\
$E_2$ & 52.9 &
$\begin{bmatrix}117&47&47&19\\56&89&59&29\\28&29&130&37\\16&19&51&155\end{bmatrix}$ \\
$E_3$ & 52.5 &
$\begin{bmatrix}116&51&37&26\\59&91&55&28\\24&34&129&37\\20&22&48&151\end{bmatrix}$ \\
$E_4$ & 56.5 &
$\begin{bmatrix}128&44&41&17\\52&103&53&25\\31&27&134&32\\14&20&48&159\end{bmatrix}$ \\
\bottomrule
\end{tabular}
\end{adjustbox}
\caption{VariErr single-annotation annotator classification. Accuracies are percentages. Chance is $25\%$ and the majority baseline is $26.0\%$.}
\label{tab:varierr-single-classifier}
\end{table*}

\begin{table*}[t]
\centering
\scriptsize
\begin{adjustbox}{max width=\textwidth}
\begin{tabular}{lcc}
\toprule
Representation & Acc. & Confusion matrix \\
\midrule
Features & 76.7 &
$\begin{bmatrix}177&28&20&25\\8&177&6&59\\22&3&204&21\\18&14&9&209\end{bmatrix}$ \\
$E_1$ & 76.6 &
$\begin{bmatrix}182&13&9&46\\30&203&7&10\\27&14&200&9\\53&6&10&181\end{bmatrix}$ \\
$E_2$ & 86.7 &
$\begin{bmatrix}201&3&10&36\\7&237&1&5\\23&2&222&3\\38&3&2&207\end{bmatrix}$ \\
$E_3$ & 85.3 &
$\begin{bmatrix}188&3&9&50\\9&236&1&4\\18&5&222&5\\40&1&2&207\end{bmatrix}$ \\
$E_4$ & 87.5 &
$\begin{bmatrix}198&2&9&41\\8&237&0&5\\20&1&225&4\\32&2&1&215\end{bmatrix}$ \\
\bottomrule
\end{tabular}
\end{adjustbox}
\caption{R2 single-annotation annotator classification. Accuracies are percentages. Chance and majority baseline are both $25\%$.}
\label{tab:r2-single-classifier}
\end{table*}

\subsubsection{Linear Input Recoverability of Residual Representations}
\label{app:linear-e0-recovery}

Section~\ref{sec:stable_single_classifier} introduces residual representations to reduce the dominance of
input content in annotator-style diagnostics. Our main residual representation
is
\begin{equation}
\begin{aligned}
E_4 =
\operatorname{norm}
\left(
E_2 - (E_2^\top E_0)E_0
\right),
\end{aligned}
\end{equation}
where $E_0$ is the input embedding and $E_2$ is the label-explanation
embedding. This construction removes the component of $E_2$ that is directly
aligned with $E_0$. However, input content may still be encoded in directions
other than the row-wise $E_0$ direction. We therefore test how much of the
original input embedding can be linearly recovered from each representation.

For each representation $E_k$, we fit a ridge regression probe on the training
items:
\begin{equation}
\widehat{E}_0 = W E_k + b,
\end{equation}
and evaluate on held-out items using the coefficient of determination:
\begin{equation}
R^2_k
=
1 -
\frac{\|E_0-\widehat{E}_0\|^2}
{\|E_0-\bar{E}_0\|^2}.
\end{equation}
Higher $R^2$ indicates that more input-level information remains linearly
recoverable from the representation.

\begin{table}[t]
\centering
\small
\begin{tabular}{llrr}
\toprule
Dataset & Source & Global $R^2$ & Median dim. $R^2$ \\
\midrule
VariErr & $E_1$ & 0.499 & 0.470 \\
VariErr & $E_2$ & 0.180 & 0.163 \\
VariErr & $E_3$ & 0.324 & 0.299 \\
VariErr & $E_4$ & \textbf{0.045} & \textbf{0.021} \\
\midrule
R2 & $E_1$ & 0.439 & 0.423 \\
R2 & $E_2$ & 0.035 & 0.032 \\
R2 & $E_3$ & 0.388 & 0.376 \\
R2 & $E_4$ & \textbf{0.030} & \textbf{0.026} \\
\bottomrule
\end{tabular}
\caption{
Linear recovery of the input embedding $E_0$ from each representation.
$E_4$ yields the lowest recoverability in both datasets, indicating that it
substantially suppresses linearly recoverable input information.
}
\label{tab:linear-e0-recovery}
\end{table}

Table~\ref{tab:linear-e0-recovery} shows that $E_4$ sharply reduces direct
input recoverability. On VariErr, the global $R^2$ decreases from 0.499 for
$E_1$ and 0.180 for $E_2$ to 0.045 for $E_4$. On R2, $E_4$ similarly gives the
lowest recoverability, with a global $R^2$ of 0.030. $E_4$ also improves over
$E_3$ in both datasets, suggesting that removing the aligned component of
$E_2$ is more effective than direct subtraction for suppressing recoverable
input information.

These results support the use of $E_4$ as the main diagnostic representation in
Section~\ref{sec:stable_single_classifier}. We do not interpret $E_4$ as a perfectly content-free style vector:
explanations necessarily refer to the input, and some content information may
remain nonlinearly or semantically encoded. Instead, $E_4$ is a conservative
content-reduced representation that substantially weakens direct input
recoverability while preserving enough label-explanation information to test
whether annotator-specific behavior is detectable. This also reinforces our
aggregation-based interpretation: single annotations remain mixtures of item
content and annotator behavior, whereas repeated outputs from the same
annotator provide a more reliable signal of annotator-specific
label-explanation behavior.

\subsection{Group-Level Classifier Sweep}
\label{app:group-sweep}

Table~\ref{tab:group-classifier-sweep} reports group-level test accuracy for all embedding representations. For each group size $m$, we bootstrap 240 train groups and 160 test groups per annotator from item-disjoint train/test pools, average the $m$ row embeddings in each group, renormalize, and train a balanced logistic-regression classifier. The corresponding curves, including dashed single-sample evaluations of the same group-trained classifiers, appear in Figures~\ref{fig:varierr-group-curve} and~\ref{fig:app-r2-group-curve}.

\begin{table*}[t]
\centering
\small
\begin{adjustbox}{max width=\textwidth}
\begin{tabular}{r|rrrr|rrrr}
\toprule
\multirow{2}{*}{$m$} & \multicolumn{4}{c|}{VariErr} & \multicolumn{4}{c}{R2} \\
 & $E_1$ & $E_2$ & $E_3$ & $E_4$ & $E_1$ & $E_2$ & $E_3$ & $E_4$ \\
\midrule
1  & 42.5 & 54.7 & 52.0 & \textbf{57.3} & 76.6 & 87.2 & 85.6 & \textbf{87.8} \\
3  & 55.6 & 67.7 & 65.9 & \textbf{68.4} & 93.0 & \textbf{96.6} & 95.9 & 96.4 \\
5  & 62.8 & 78.0 & 74.8 & \textbf{78.8} & 94.5 & 98.4 & 98.6 & \textbf{98.8} \\
10 & 70.9 & 84.5 & 84.5 & \textbf{85.3} & 99.4 & 99.8 & 99.8 & \textbf{100.0} \\
15 & 73.6 & 87.8 & 87.7 & \textbf{89.1} & 99.7 & \textbf{100.0} & 99.5 & \textbf{100.0} \\
20 & 75.6 & \textbf{93.3} & 90.5 & 92.3 & 99.8 & \textbf{100.0} & \textbf{100.0} & \textbf{100.0} \\
30 & 81.7 & 94.2 & 93.8 & \textbf{94.4} & 99.8 & \textbf{100.0} & \textbf{100.0} & \textbf{100.0} \\
50 & 83.9 & \textbf{97.0} & 96.4 & 96.4 & \textbf{100.0} & \textbf{100.0} & \textbf{100.0} & \textbf{100.0} \\
\bottomrule
\end{tabular}
\end{adjustbox}
\caption{Group-level annotator-classifier test accuracy by representation and group size. Values are percentages; chance is $25\%$. E4 is the strongest or near-strongest representation while retaining the clearest content-removal interpretation.}
\label{tab:group-classifier-sweep}
\end{table*}

\paragraph{Train- and Test-Time Effects of Aggregation}
Section~\ref{sec:stable_group} shows that annotator-specific behavior becomes more recoverable when annotations are aggregated into groups. To disentangle whether this gain comes from aggregation during classifier training or during evaluation, we compare three settings. In the standard group setting, classifiers are trained on groups of size $m$ and evaluated on groups of size $m$ ($m$ train $\rightarrow$ $m$ test). We also evaluate the same group-trained classifiers on individual annotations ($m$ train $\rightarrow$ 1 test), and introduce the complementary setting where classifiers are trained on individual annotations but evaluated on groups of size $m$ (1 train $\rightarrow$ $m$ test).

\begin{figure*}[t]
\centering
\begin{minipage}{0.49\textwidth}
\centering
\includegraphics[width=\linewidth]{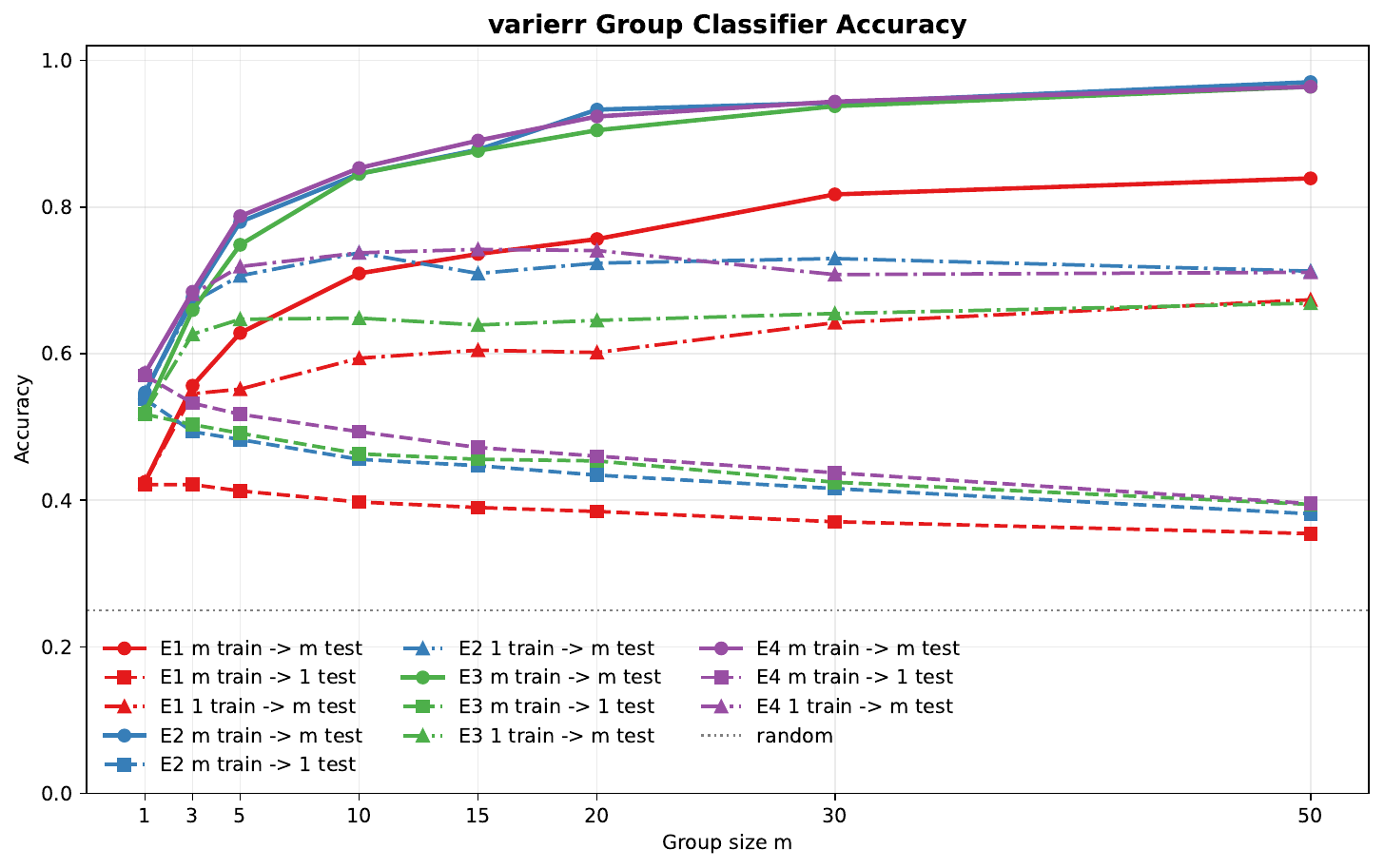}
{\footnotesize (a) VariErr}
\end{minipage}
\hfill
\begin{minipage}{0.49\textwidth}
\centering
\includegraphics[width=\linewidth]{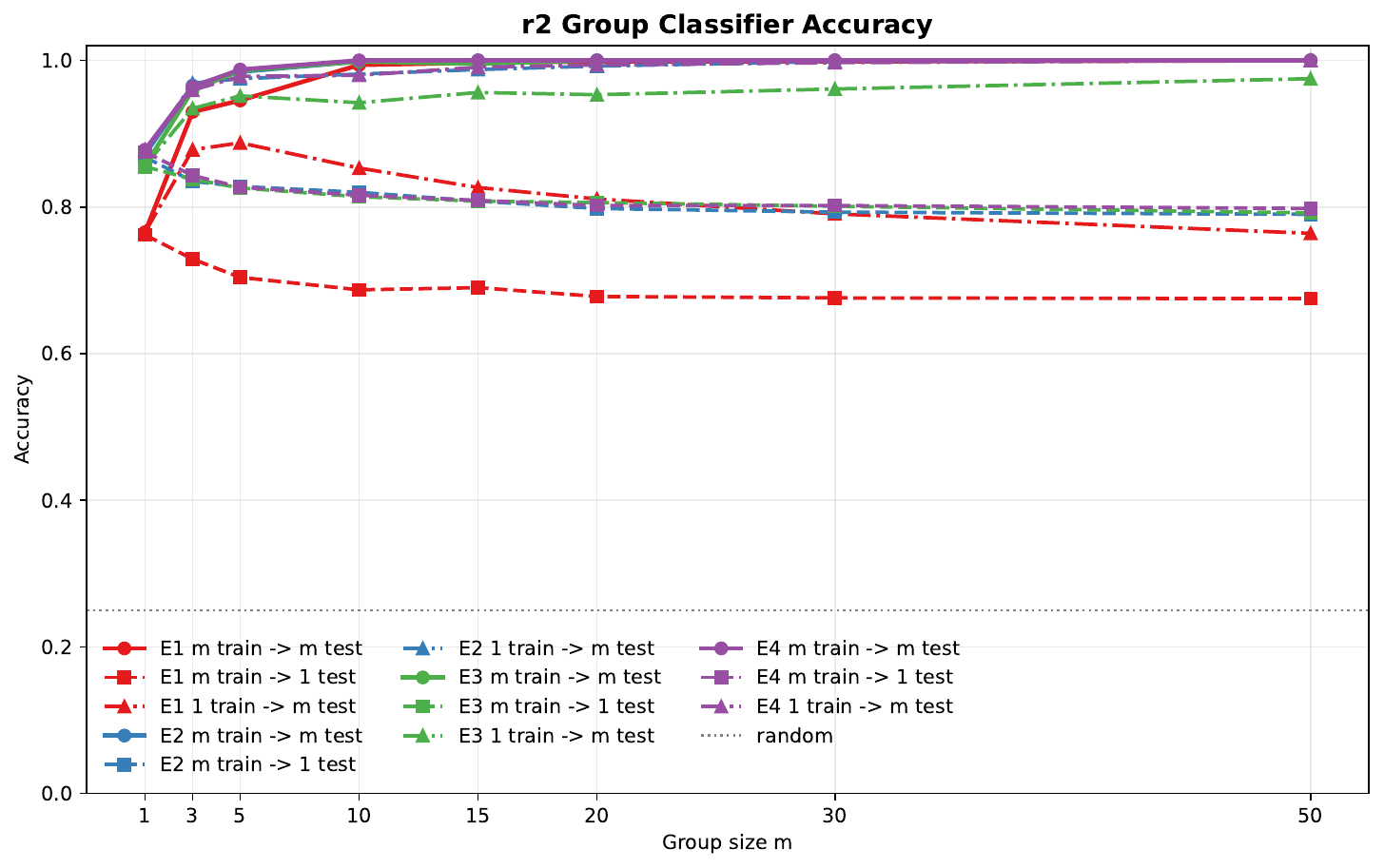}
{\footnotesize (b) R2}
\end{minipage}
\caption{Train- and test-time effects of aggregation on annotator-classifier accuracy.
    We compare classifiers trained and evaluated on groups of size $m$ ($m$ train $\rightarrow$ $m$ test), classifiers trained on groups but evaluated on individual annotations ($m$ train $\rightarrow$ 1 test), and classifiers trained on individual annotations but evaluated on groups (1 train $\rightarrow$ $m$ test).
    Test-time aggregation alone improves accuracy as $m$ increases, but the strongest annotator recoverability is obtained when aggregation is used at both training and test time.
    }
\label{fig:train-test-aggregation}
\end{figure*}

Figure~\ref{fig:train-test-aggregation} disentangles train- and test-time aggregation effects.
The results show that aggregation helps at test time as well as at train time. As $m$ increases, the 1 train $\rightarrow$ $m$ test setting consistently improves over single-instance evaluation, indicating that averaging test annotations suppresses item-specific content and makes annotator-level patterns more visible. However, this setting generally converges below the full $m$ train $\rightarrow$ $m$ test setting, showing that test-time aggregation alone is not sufficient to obtain the strongest separation. Conversely, the $m$ train $\rightarrow$ 1 test setting often decreases with larger $m$, suggesting that classifiers trained to recognize group-level behavior do not transfer well to isolated annotations. Together, these results indicate that aggregation is beneficial on both sides: test-time aggregation improves the signal-to-noise ratio of annotator behavior, while train-time aggregation helps the classifier learn the corresponding group-level decision boundaries. The strongest recoverability is therefore achieved when both training and evaluation operate at the aggregated annotator level.

\subsection{R2 Stable Variation Analysis}
\label{app:r2-stable}

\begin{figure}[t]
\centering
\begin{minipage}{0.48\columnwidth}
\centering
\includegraphics[width=\linewidth]{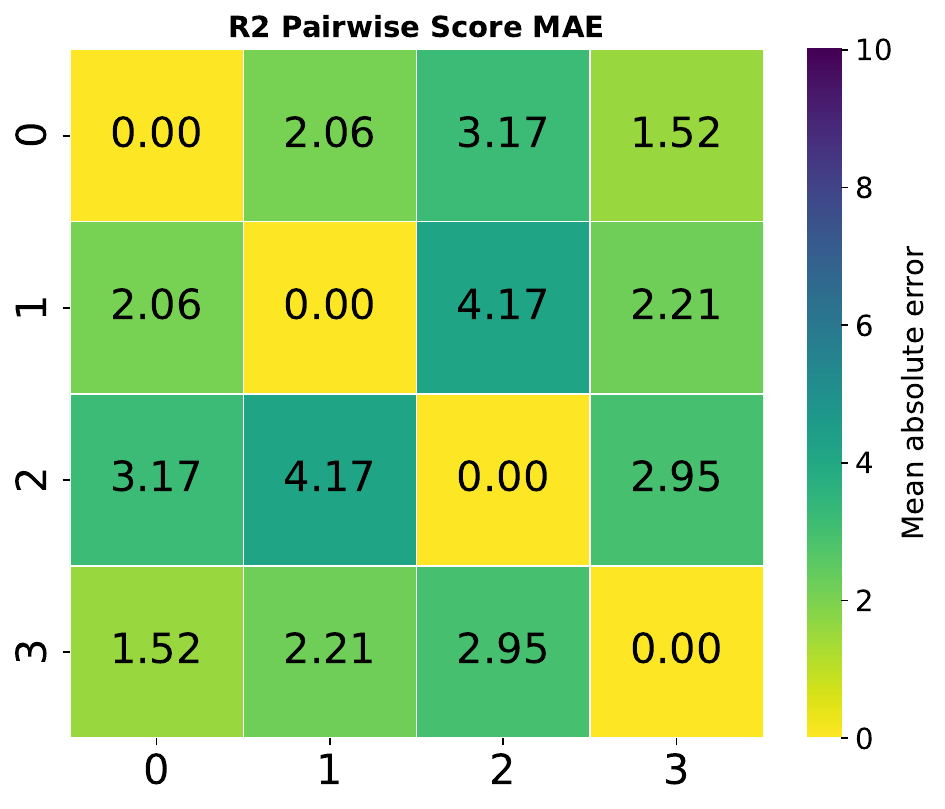}
{\footnotesize (a) Pairwise score MAE}
\end{minipage}
\hfill
\begin{minipage}{0.48\columnwidth}
\centering
\includegraphics[width=\linewidth]{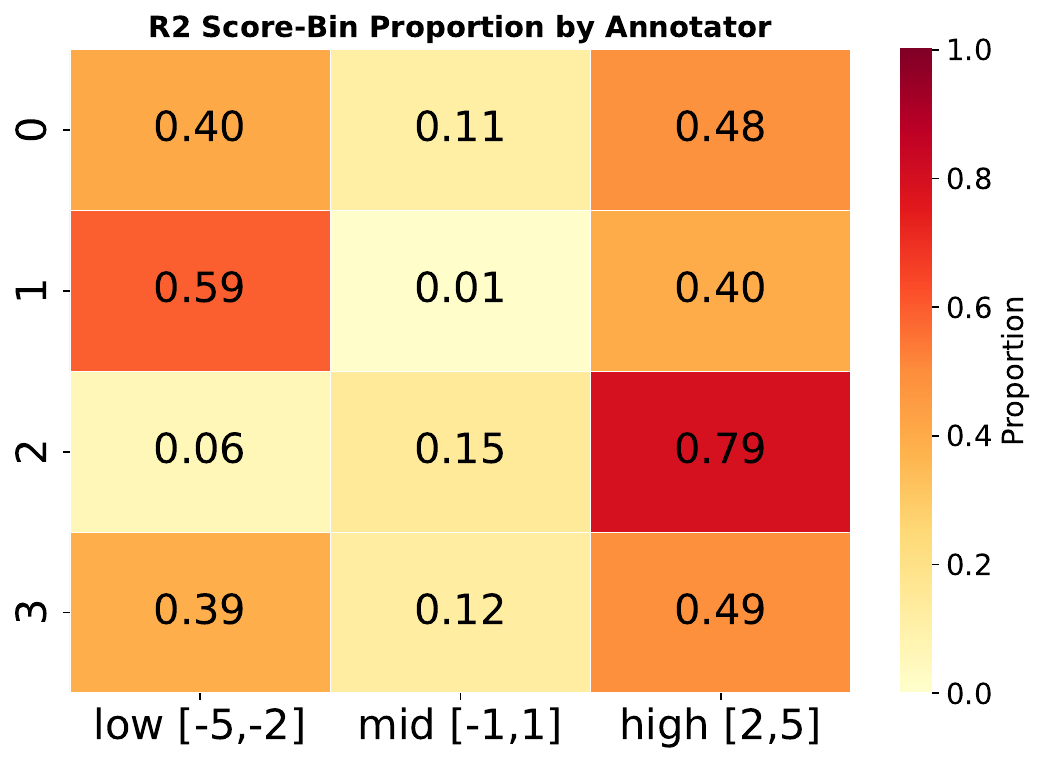}
{\footnotesize (b) Score-bin proportions}
\end{minipage}
\caption{R2 score variation across annotators. R2 uses ordinal paraphrase-relatedness scores, so pairwise MAE is more interpretable than nominal kappa.}
\label{fig:app-r2-labels}
\end{figure}

R2 shows the same qualitative pattern as VariErr, but with stronger annotator-specific signal. The label side is ordinal rather than nominal, so we plot pairwise score mean absolute error (MAE) and coarse score-bin proportions in Figure~\ref{fig:app-r2-labels}. Pairwise MAE ranges from 1.52 to 4.17. Annotator 1 assigns low scores much more often ($59.2\%$), while Annotator 2 assigns high scores most often ($79.2\%$). This creates a stronger score-style signal than in VariErr.

\begin{figure}[t]
\centering
\includegraphics[width=0.95\columnwidth]{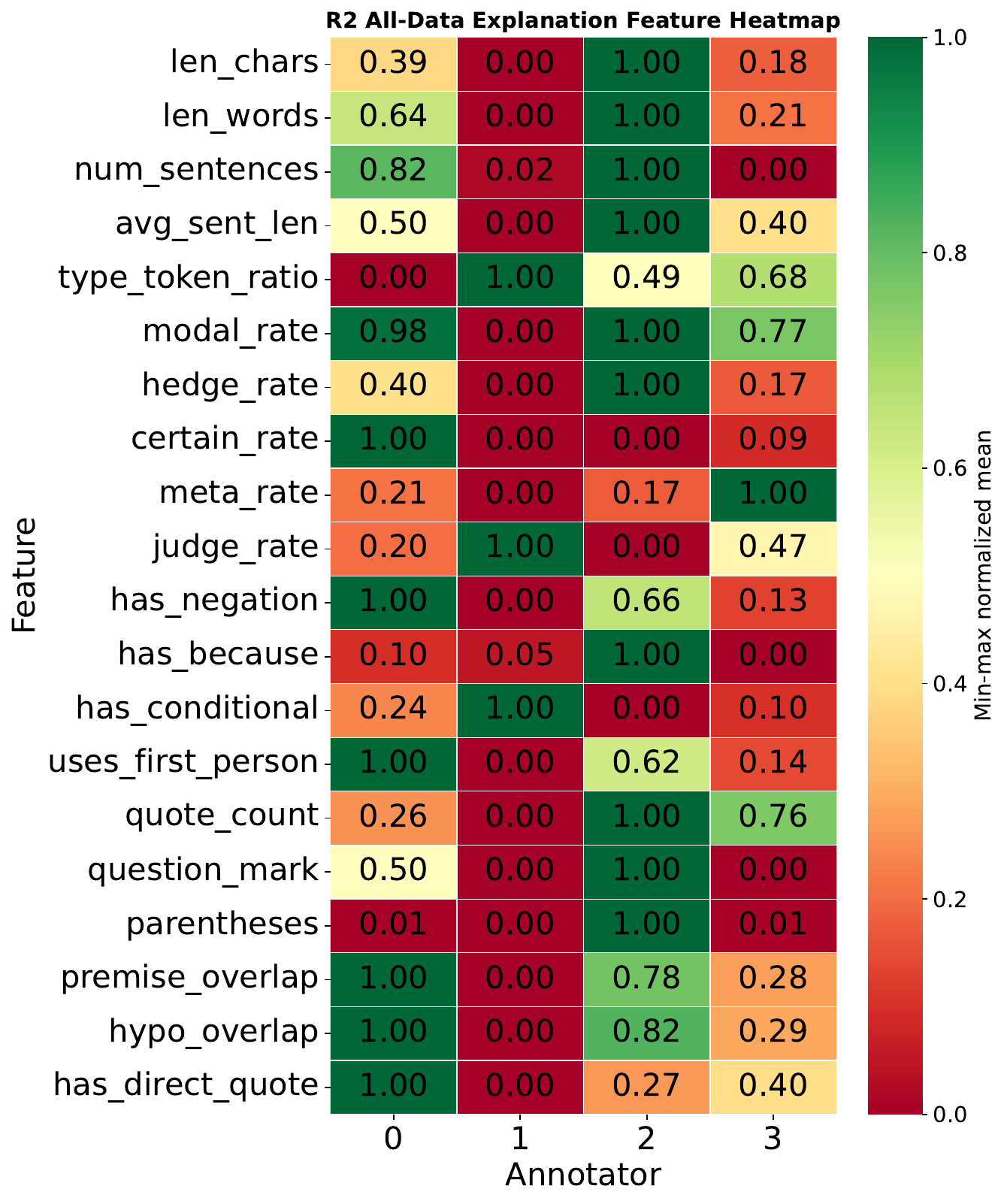}
\caption{R2 explanation-style features averaged by annotator and min-max normalized within each feature.}
\label{fig:app-r2-feature-heatmap}
\end{figure}

\begin{figure}[t]
\centering
\includegraphics[width=0.95\columnwidth]{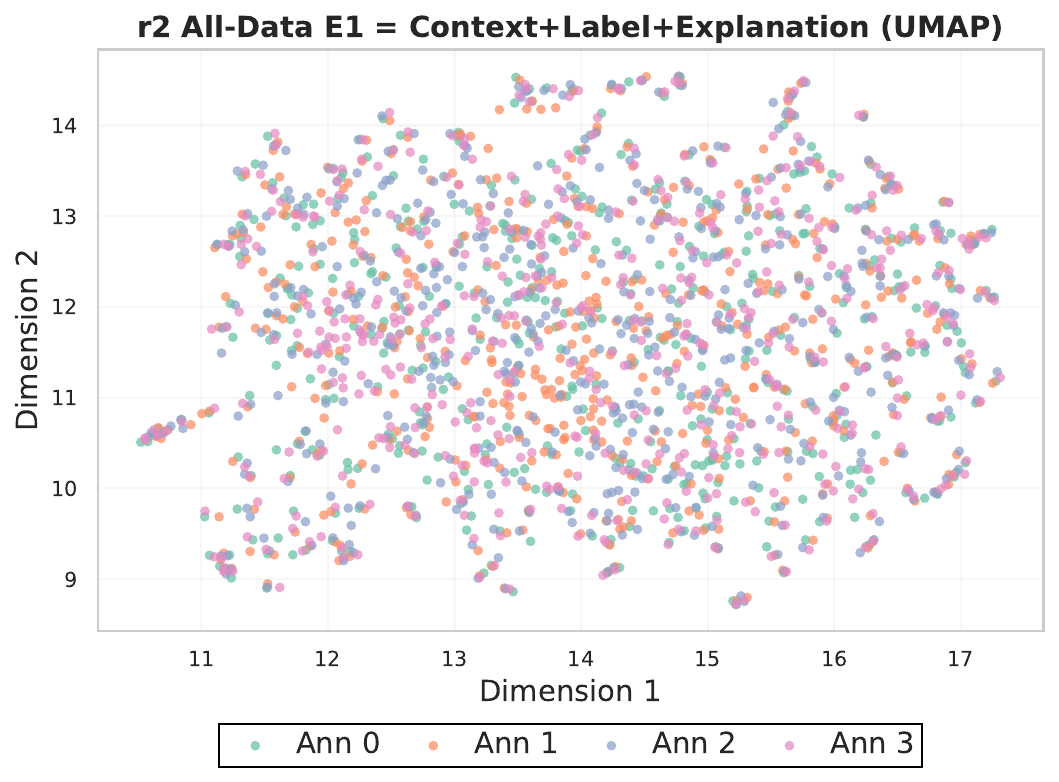}
\caption{R2 $E_1$ UMAP. Compared with VariErr, annotator structure is more visible, but aggregation still provides a cleaner signal.}
\label{fig:app-r2-e1-umap}
\end{figure}

\begin{figure*}[t]
\centering
\begin{minipage}{0.32\textwidth}
\centering
\includegraphics[width=\linewidth]{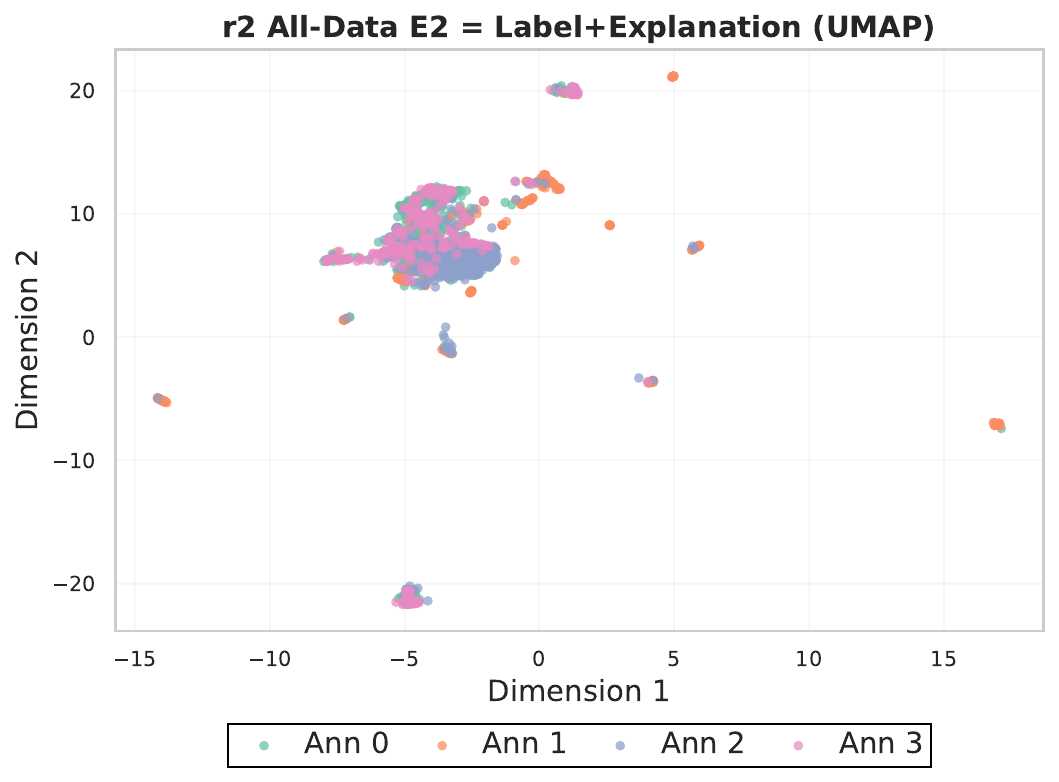}
{\footnotesize (a) $E_2$}
\end{minipage}
\hfill
\begin{minipage}{0.32\textwidth}
\centering
\includegraphics[width=\linewidth]{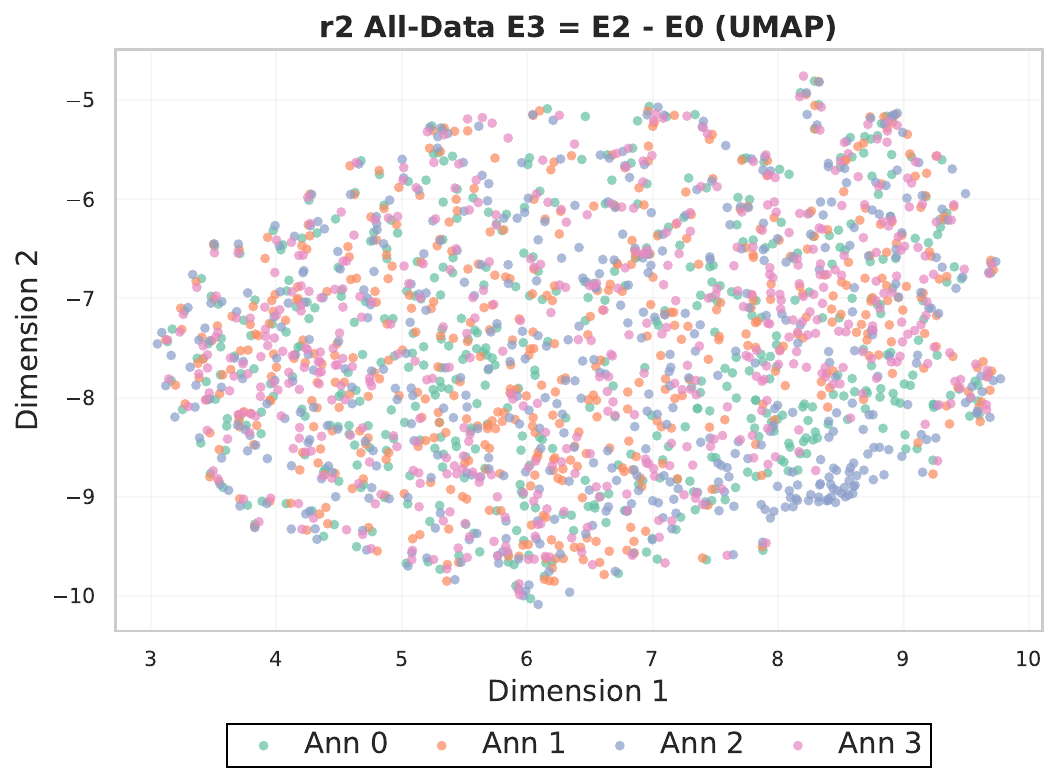}
{\footnotesize (b) $E_3$}
\end{minipage}
\hfill
\begin{minipage}{0.32\textwidth}
\centering
\includegraphics[width=\linewidth]{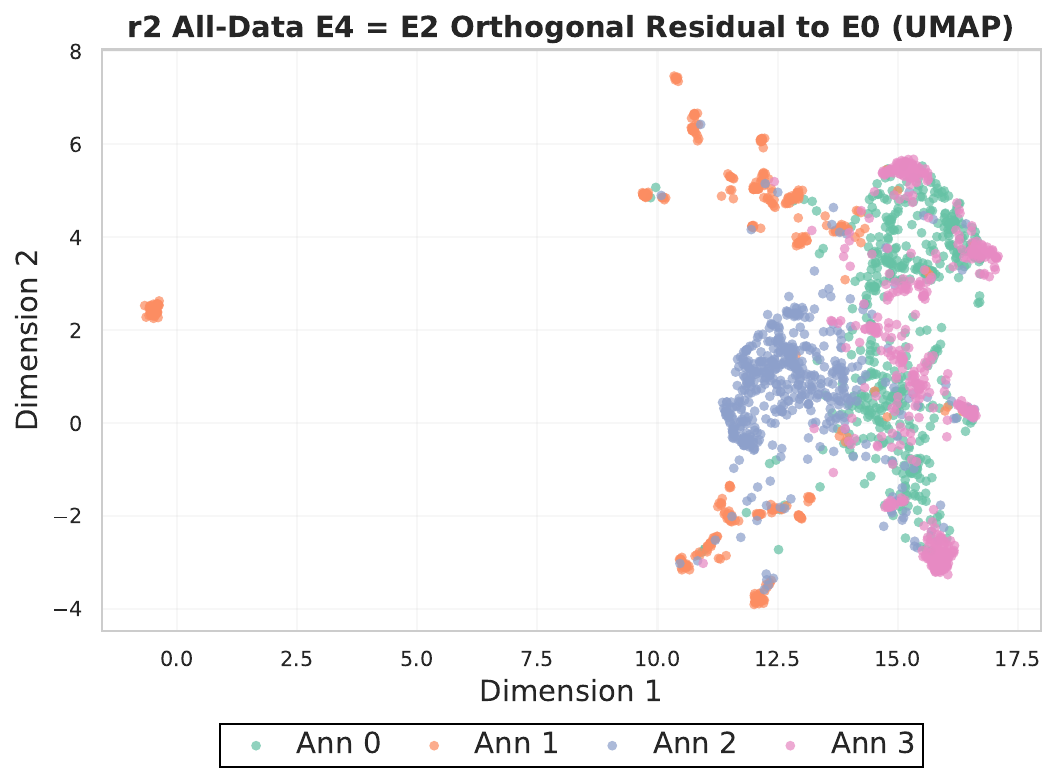}
{\footnotesize (c) $E_4$}
\end{minipage}
\caption{R2 individual-annotation UMAPs for content-reduced explanation representations.}
\label{fig:app-r2-single-umaps}
\end{figure*}

\begin{figure*}[t]
\centering
\includegraphics[width=\textwidth]{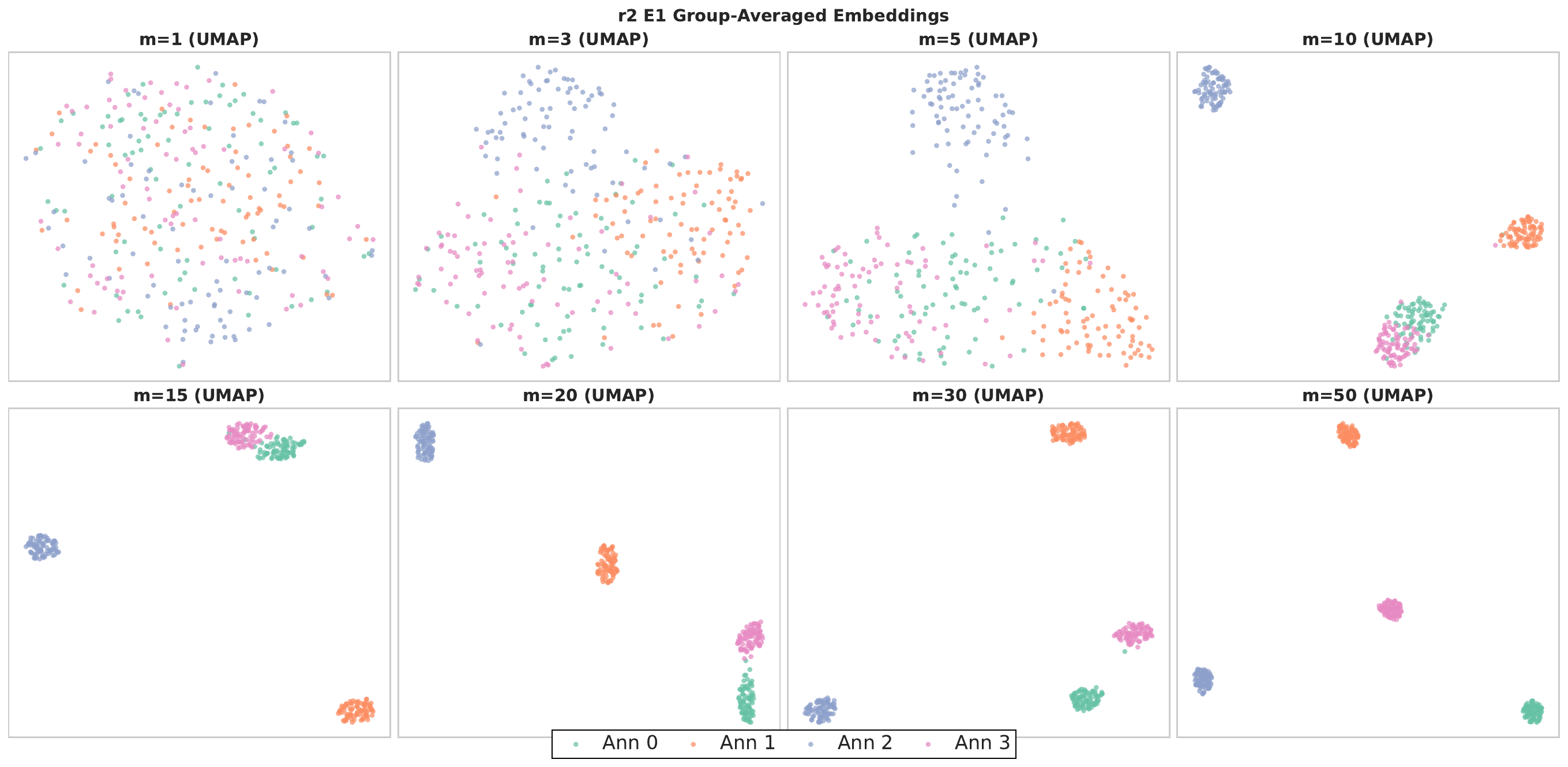}
\caption{R2 group-averaged $E_1$ embeddings. Even the raw context-containing representation becomes highly separable after aggregation.}
\label{fig:app-r2-e1-group-umap}
\end{figure*}

\begin{figure*}[t]
\centering
\includegraphics[width=\textwidth]{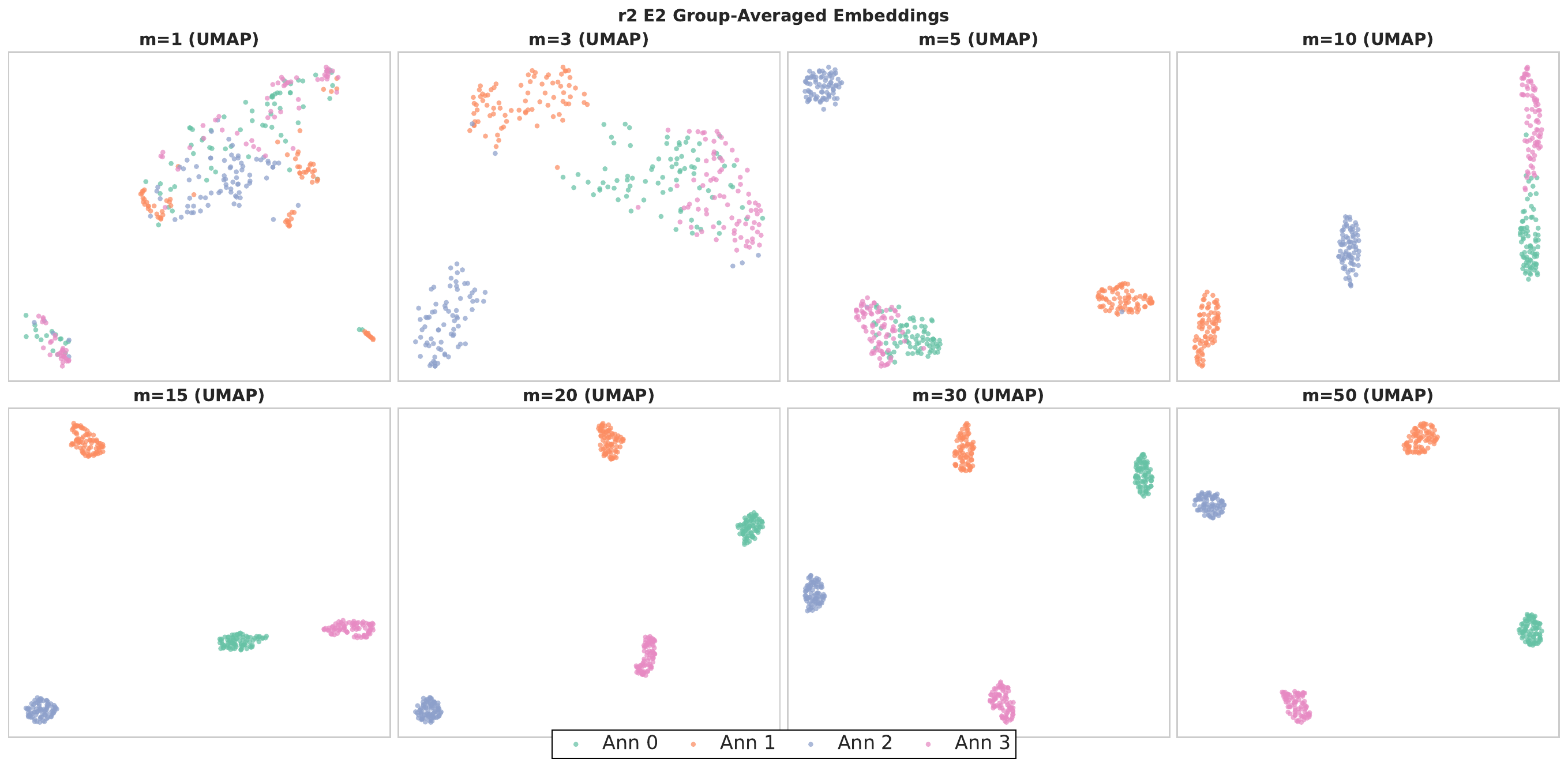}
\caption{R2 group-averaged $E_2$ embeddings. Label-score and explanation embeddings separate annotators quickly under aggregation.}
\label{fig:app-r2-e2-group-umap}
\end{figure*}

\begin{figure*}[t]
\centering
\includegraphics[width=\textwidth]{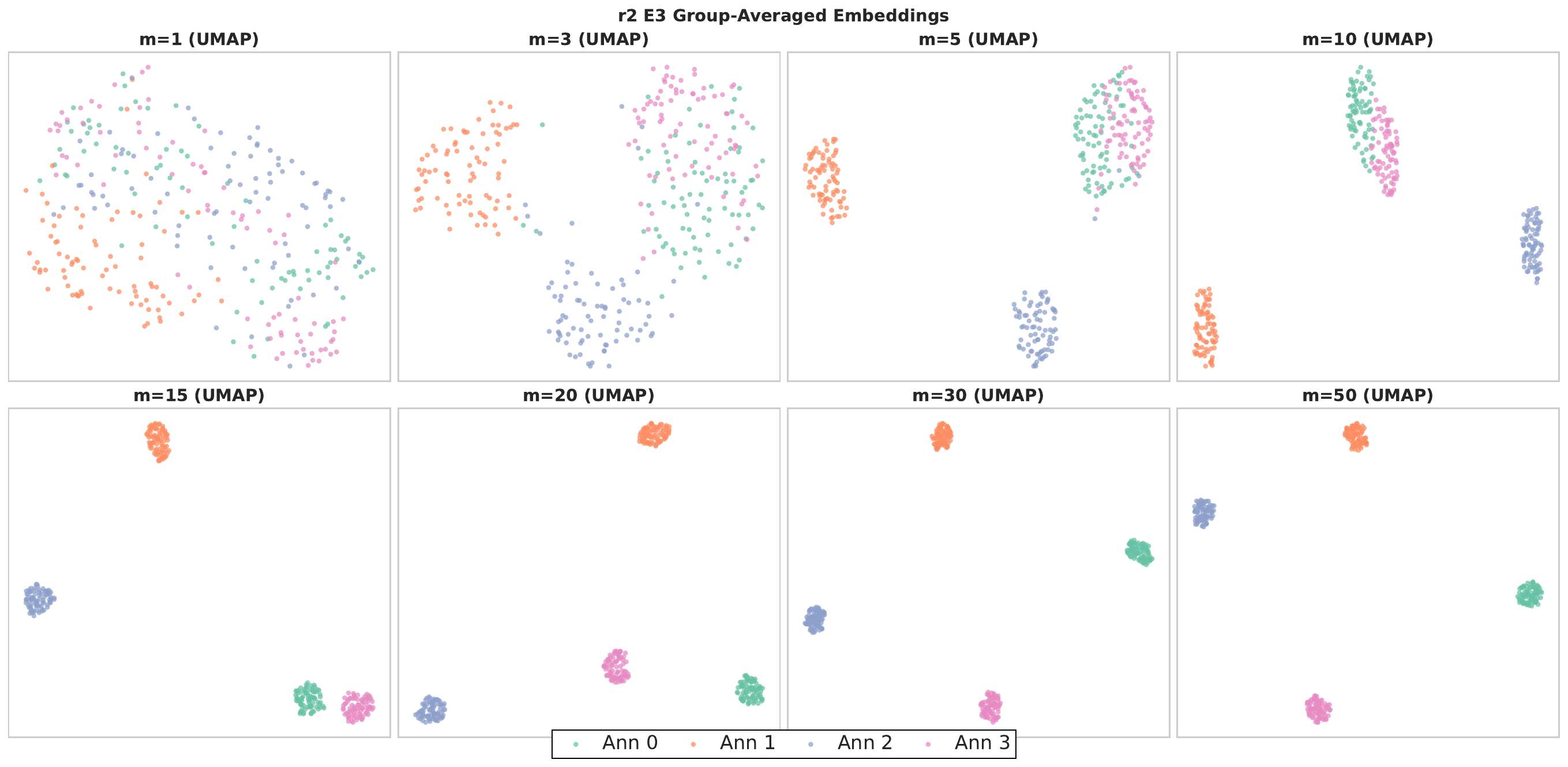}
\caption{R2 group-averaged $E_3$ embeddings. The residual representation also saturates to near-perfect group-level separability.}
\label{fig:app-r2-e3-group-umap}
\end{figure*}

The explanation analysis also parallels VariErr. Figure~\ref{fig:app-r2-feature-heatmap} shows clear feature differences, and Figure~\ref{fig:app-r2-e1-umap} shows that raw $E_1$ embeddings are less mixed than in VariErr but still partly content-shaped. In the single-annotation classifier, features reach $76.7\%$, $E_1$ reaches $76.6\%$, and $E_4$ reaches $87.5\%$ (Table~\ref{tab:r2-single-classifier}). The higher numbers suggest that R2 annotators have more separable label-explanation habits, likely due to both task format and stronger score tendencies. Figure~\ref{fig:app-r2-single-umaps} shows R2 UMAPs for $E_2$, $E_3$, and $E_4$ at the individual-annotation level. Figures~\ref{fig:app-r2-e1-group-umap}--\ref{fig:app-r2-e3-group-umap} show the corresponding group-averaged views for $E_1$--$E_3$.

\begin{figure*}[t]
\centering
\includegraphics[width=\textwidth]{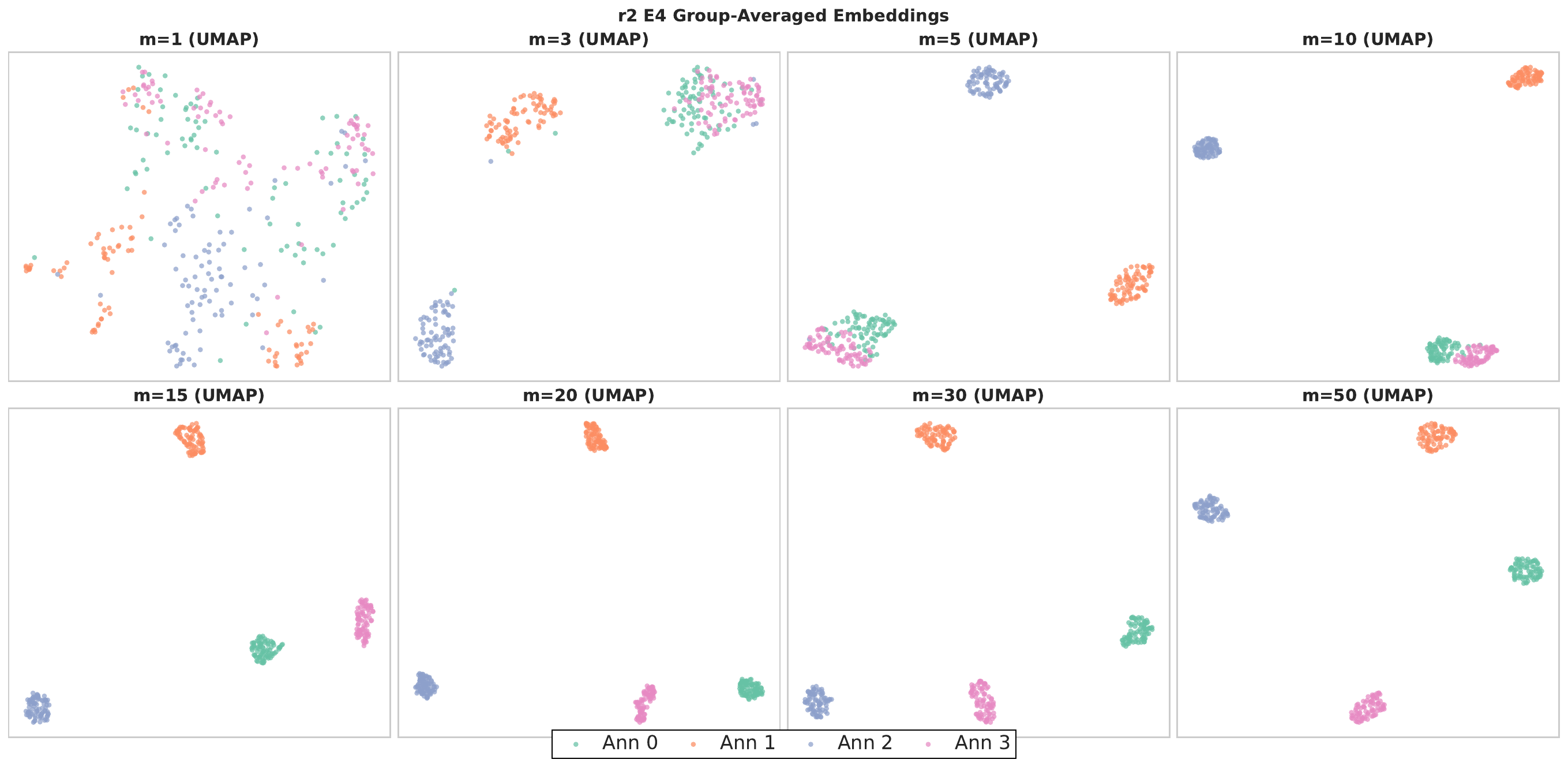}
\caption{R2 group-averaged $E_4$ embeddings. Annotator clusters become nearly perfectly separable after modest aggregation.}
\label{fig:app-r2-e4-group-umap}
\end{figure*}

\begin{figure}[t]
\centering
\includegraphics[width=0.98\columnwidth]{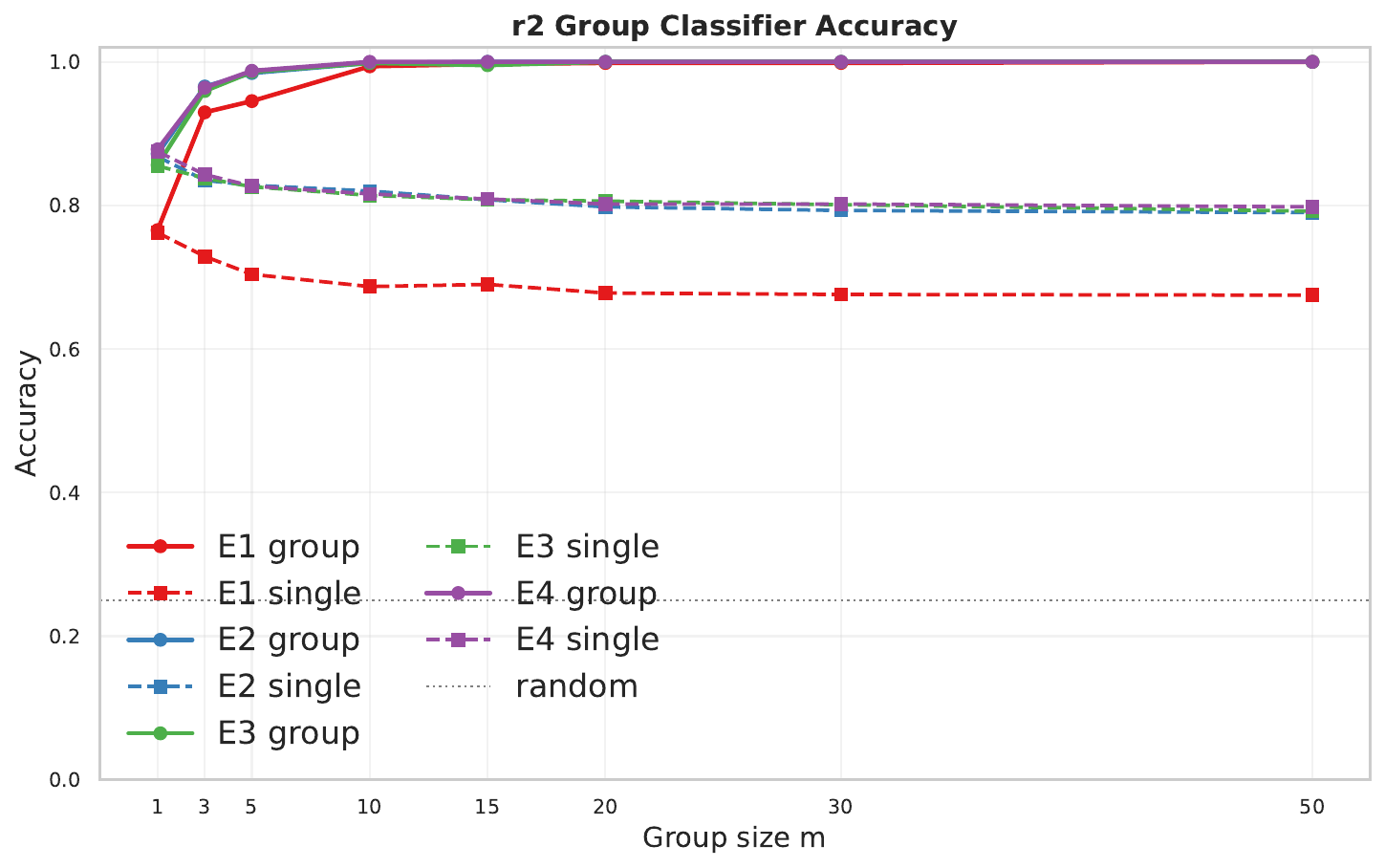}
\caption{R2 annotator-classifier accuracy by representation and group size. The group-level signal is stronger than in VariErr and saturates quickly.}
\label{fig:app-r2-group-curve}
\end{figure}

At the group level, R2 rapidly becomes separable. The $E_4$ group classifier reaches $87.8\%$ at $m=1$, $96.4\%$ at $m=3$, $98.8\%$ at $m=5$, and $100.0\%$ by $m=10$. Figures~\ref{fig:app-r2-e4-group-umap} and~\ref{fig:app-r2-group-curve} visualize this result. Thus, R2 supports the same conclusion as VariErr: annotator-specific label-explanation behavior is a stable group-level signal, but R2's signal is easier to detect from individual annotations.

\section{Full Results}
\label{app:full_results}

Tables~\ref{tab:app-varierr-full} and~\ref{tab:app-r2-full} expand the main tables with shared SFT, BERTScore, embedding cosine, and train-only group-classifier confidence. The GC Conf column in the main paper corresponds to the ``All Conf'' column here. Prompt rows use $n=50$.

\begin{table*}[p]
\centering
\scriptsize
\begin{adjustbox}{max width=\textwidth}
\begin{tabular}{llrrrrrrrrr}
\toprule
Model & Setting & Acc $\uparrow$ & R-L $\uparrow$ & BERTScore $\uparrow$ & EmbSim $\uparrow$ & Feature KL $\downarrow$ & Train Conf $\uparrow$ & All Conf $\uparrow$ & ImiScore $\uparrow$ & Judge Acc $\uparrow$ \\
\midrule
\multirow{7}{*}{Qwen3}
& Base & 0.472 & 0.159 & 0.866 & 0.516 & 2.394 & 0.240 & 0.239 & -0.034 & 0.237 \\
& ICL & 0.535 & 0.186 & 0.873 & 0.559 & 0.692 & 0.335 & 0.344 & 0.179 & 0.265 \\
& VP & 0.492 & 0.131 & 0.857 & 0.517 & 4.391 & 0.217 & 0.234 & -0.348 & 0.240 \\
& VP-ICL & 0.545 & 0.201 & 0.877 & 0.582 & 0.476 & 0.520 & 0.494 & 0.528 & 0.282 \\
& SFT-shared & 0.632 & 0.244 & 0.887 & 0.562 & 0.280 & 0.296 & 0.312 & 0.308 & 0.263 \\
& SFT-ind. & 0.637 & 0.283 & 0.892 & 0.590 & 0.084 & 0.805 & 0.845 & 0.859 & 0.300 \\
& CAPO-strict & 0.627 & 0.281 & 0.891 & 0.582 & 0.081 & 0.823 & 0.867 & 0.888 & 0.328 \\
\midrule
\multirow{7}{*}{Llama3.2}
& Base & 0.315 & 0.153 & 0.865 & 0.483 & 2.036 & 0.265 & 0.258 & 0.023 & 0.265 \\
& ICL & 0.380 & 0.214 & 0.876 & 0.550 & 0.476 & 0.291 & 0.295 & 0.375 & 0.260 \\
& VP & 0.338 & 0.148 & 0.863 & 0.467 & 1.907 & 0.379 & 0.358 & 0.325 & 0.253 \\
& VP-ICL & 0.412 & 0.210 & 0.877 & 0.552 & 0.539 & 0.260 & 0.260 & 0.432 & 0.275 \\
& SFT-shared & 0.575 & 0.229 & 0.883 & 0.554 & 0.372 & 0.255 & 0.256 & 0.048 & 0.255 \\
& SFT-ind. & 0.512 & 0.271 & 0.887 & 0.549 & 0.113 & 0.873 & 0.909 & 0.946 & 0.282 \\
& CAPO-strict & 0.512 & 0.262 & 0.887 & 0.540 & 0.121 & 0.885 & 0.924 & 0.964 & 0.297 \\
\bottomrule
\end{tabular}
\end{adjustbox}
\caption{Expanded results on VariErr. CAPO uses strict same-label pairs and train-only group-classifier checkpoint selection.}
\label{tab:app-varierr-full}
\end{table*}

\begin{table*}[p]
\centering
\scriptsize
\begin{adjustbox}{max width=\textwidth}
\begin{tabular}{llrrrrrrrrrr}
\toprule
Model & Setting & Acc $\uparrow$ & MAE $\downarrow$ & R-L $\uparrow$ & BERTScore $\uparrow$ & EmbSim $\uparrow$ & Feature KL $\downarrow$ & Train Conf $\uparrow$ & All Conf $\uparrow$ & ImiScore $\uparrow$ & Judge Acc $\uparrow$ \\
\midrule
\multirow{7}{*}{Qwen3}
& Base & 0.292 & 2.453 & 0.071 & 0.840 & 0.395 & 5.188 & 0.248 & 0.248 & -0.004 & 0.237 \\
& ICL & 0.407 & 1.702 & 0.197 & 0.871 & 0.535 & 0.587 & 0.800 & 0.809 & 0.810 & 0.463 \\
& VP & 0.278 & 2.475 & 0.100 & 0.850 & 0.480 & 2.740 & 0.657 & 0.649 & 0.651 & 0.360 \\
& VP-ICL & 0.415 & 1.667 & 0.276 & 0.885 & 0.601 & 0.246 & 0.938 & 0.947 & 0.956 & 0.510 \\
& SFT-shared & 0.443 & 1.617 & 0.304 & 0.890 & 0.569 & 0.220 & 0.807 & 0.800 & 0.806 & 0.455 \\
& SFT-ind. & 0.450 & 1.478 & 0.307 & 0.891 & 0.582 & 0.130 & 0.936 & 0.929 & 0.936 & 0.512 \\
& CAPO-tol1 & 0.455 & 1.490 & 0.316 & 0.892 & 0.595 & 0.112 & 0.971 & 0.964 & 0.973 & 0.560 \\
\midrule
\multirow{7}{*}{Llama3.2}
& Base & 0.207 & 3.188 & 0.068 & 0.835 & 0.380 & 1.515 & 0.251 & 0.255 & 0.015 & 0.230 \\
& ICL & 0.273 & 3.333 & 0.213 & 0.875 & 0.546 & 0.480 & 0.875 & 0.871 & 0.879 & 0.367 \\
& VP & 0.180 & 4.185 & 0.098 & 0.850 & 0.401 & 0.913 & 0.347 & 0.359 & 0.175 & 0.260 \\
& VP-ICL & 0.273 & 3.390 & 0.230 & 0.880 & 0.558 & 0.257 & 0.875 & 0.858 & 0.865 & 0.370 \\
& SFT-shared & 0.385 & 2.050 & 0.209 & 0.870 & 0.425 & 0.683 & 0.305 & 0.305 & 0.191 & 0.265 \\
& SFT-ind. & 0.450 & 1.535 & 0.290 & 0.889 & 0.570 & 0.118 & 0.969 & 0.963 & 0.972 & 0.512 \\
& CAPO-tol1 & 0.460 & 1.470 & 0.293 & 0.889 & 0.582 & 0.187 & 0.988 & 0.985 & 0.995 & 0.530 \\
\bottomrule
\end{tabular}
\end{adjustbox}
\caption{Expanded results on R2. CAPO uses tolerance-1 pairs and train-only group-classifier checkpoint selection.}
\label{tab:app-r2-full}
\end{table*}

\subsection{Prompt Size Sweep}
\label{app:prompt-size-sweep}

Table~\ref{tab:prompt-size-sweep} summarizes the prompt-size sweep for ICL, VP, and VP-ICL. The lists are ordered by $n\in\{3,5,10,15,20,30,40,50\}$. Task is label accuracy for VariErr and score accuracy for R2. The main text uses $n=50$ because it is the largest setting that fits both model context windows and because the sweep over both models and datasets favors the largest prompt budget overall, especially for R2 group confidence.

\begin{table*}[p]
\centering
\scriptsize
\begin{adjustbox}{max width=\textwidth}
\begin{tabular}{lllcccc}
\toprule
Dataset & Model & Method & Task by $n$ & GC Conf by $n$ & KL@50 $\downarrow$ & Judge@50 $\uparrow$ \\
\midrule
VariErr & Qwen3 & ICL & 0.46, 0.53, 0.46, 0.53, 0.55, 0.53, 0.57, 0.54 & 0.30, 0.34, 0.30, 0.29, 0.29, 0.32, 0.38, 0.34 & 0.692 & 0.265 \\
VariErr & Qwen3 & VP & 0.55, 0.54, 0.54, 0.52, 0.53, 0.51, 0.56, 0.49 & 0.19, 0.30, 0.38, 0.32, 0.29, 0.31, 0.27, 0.23 & 4.391 & 0.240 \\
VariErr & Qwen3 & VP-ICL & 0.56, 0.54, 0.52, 0.54, 0.56, 0.56, 0.58, 0.55 & 0.41, 0.61, 0.64, 0.48, 0.32, 0.45, 0.45, 0.49 & 0.476 & 0.282 \\
VariErr & Llama3.2 & ICL & 0.35, 0.35, 0.38, 0.41, 0.48, 0.48, 0.42, 0.38 & 0.30, 0.30, 0.31, 0.27, 0.26, 0.28, 0.31, 0.30 & 0.476 & 0.260 \\
VariErr & Llama3.2 & VP & 0.38, 0.33, 0.32, 0.36, 0.39, 0.37, 0.32, 0.34 & 0.26, 0.02, 0.46, 0.26, 0.26, 0.22, 0.34, 0.36 & 1.907 & 0.253 \\
VariErr & Llama3.2 & VP-ICL & 0.47, 0.34, 0.39, 0.49, 0.47, 0.51, 0.42, 0.41 & 0.61, 0.07, 0.50, 0.28, 0.26, 0.33, 0.32, 0.26 & 0.539 & 0.275 \\
R2 & Qwen3 & ICL & 0.38, 0.38, 0.36, 0.41, 0.38, 0.40, 0.37, 0.41 & 0.55, 0.77, 0.78, 0.74, 0.73, 0.78, 0.78, 0.81 & 0.587 & 0.463 \\
R2 & Qwen3 & VP & 0.22, 0.28, 0.32, 0.24, 0.34, 0.30, 0.31, 0.28 & 0.75, 0.55, 0.63, 0.55, 0.53, 0.62, 0.59, 0.65 & 2.740 & 0.360 \\
R2 & Qwen3 & VP-ICL & 0.33, 0.40, 0.36, 0.43, 0.41, 0.44, 0.38, 0.41 & 0.84, 0.96, 0.91, 0.90, 0.93, 0.89, 0.85, 0.95 & 0.246 & 0.510 \\
R2 & Llama3.2 & ICL & 0.23, 0.29, 0.26, 0.38, 0.31, 0.33, 0.34, 0.27 & 0.53, 0.74, 0.88, 0.91, 0.78, 0.94, 0.80, 0.87 & 0.480 & 0.367 \\
R2 & Llama3.2 & VP & 0.21, 0.21, 0.16, 0.22, 0.22, 0.20, 0.22, 0.18 & 0.23, 0.36, 0.56, 0.46, 0.44, 0.42, 0.43, 0.36 & 0.913 & 0.260 \\
R2 & Llama3.2 & VP-ICL & 0.24, 0.23, 0.23, 0.36, 0.29, 0.32, 0.30, 0.27 & 0.39, 0.90, 0.93, 0.97, 0.89, 0.92, 0.96, 0.86 & 0.257 & 0.370 \\
\bottomrule
\end{tabular}
\end{adjustbox}
\caption{Prompt-size sweep. Each list follows $n=3,5,10,15,20,30,40,50$. The trends are nonmonotonic on VariErr but generally favor larger prompt budgets for R2 group confidence.}
\label{tab:prompt-size-sweep}
\end{table*}

\subsection{Shared vs. Independent SFT}
\label{app:shared_sft}

Table~\ref{tab:app-varierr-full} includes an additional
SFT-shared baseline, where a single LoRA adapter is trained on the
annotations from all annotators and conditioned on the symbolic
annotator ID. This setting has access to more total training data
than each independent SFT adapter, but it produces substantially
weaker annotator separation. On VariErr, for example, SFT-shared
achieves much lower group-classifier confidence than SFT-ind. for
both base models: $0.312$ vs. $0.845$ for Qwen3 and $0.256$ vs.
$0.909$ for Llama3.2.

This gap suggests that simply providing an annotator ID as a
conditioning signal is insufficient for learning fine-grained
annotator-specific label-explanation behavior. Since the shared
adapter receives gradients from all annotators in the same parameter
space, the model may learn an averaged annotation policy and use the
ID only weakly, especially when the differences among annotators are
subtle and entangled with item content. In contrast, independent
adapters allocate a separate adapted parameter space to each target
annotator, making it easier to preserve repeated label preferences,
explanation style, and reasoning habits.

The shared-SFT result therefore strengthens the motivation for our
adapter-based design and for CAPO. If a single shared adapter with
ID conditioning were sufficient, then cross-annotator contrastive
training in annotator-specific parameter spaces would be less
necessary. Instead, the weak separation of SFT-shared indicates that
annotator-specific behavior is not reliably recovered from a
symbolic ID alone; it benefits from being represented in
target-specific adapted parameters and further sharpened by
cross-annotator preference supervision.

\subsection{Statistical Evidence for Aggregation-Aware Imitation}
\label{app:significance}

We further assess whether CAPO's gains over SFT are stable under resampling.
Since our main imitation claim concerns repeated annotator-specific behavior,
we focus on ImiScore, the aggregation-aware metric used in the main results.
Unlike decision accuracy or Judge accuracy, ImiScore is not a direct average over
independent test instances: it is computed by forming groups of generated
outputs, embedding and averaging each group, applying a fixed annotator group
classifier, and normalizing the resulting target-annotator confidence. We
therefore use a non-parametric item-level bootstrap rather than a standard
parametric test.

For each model-dataset pair, we draw $B=1000$ bootstrap samples by resampling
the 100 test items with replacement. For each sample, we rebuild
annotator-specific output groups, evaluate them with the same fixed group
classifier used in the main evaluation, and recompute the macro-average
ImiScore. We report the bootstrap mean and the 95\% percentile interval. We
also compute a paired one-sided bootstrap $p$-value, defined as the proportion
of bootstrap samples in which CAPO does not exceed SFT.

\begin{table*}[t]
\centering
\small
\begin{tabular}{llccc}
\toprule
Model & Dataset & SFT & CAPO & $\Delta$ \\
\midrule
Llama3.2 & R2
& 0.971 [0.947, 0.991]
& 0.995 [0.984, 1.010]
& +0.024 \\
Qwen3 & R2
& 0.934 [0.902, 0.964]
& 0.973 [0.949, 0.992]
& +0.039 \\
Llama3.2 & VariErr
& 0.951 [0.887, 1.011]
& 0.969 [0.910, 1.028]
& +0.018 \\
Qwen3 & VariErr
& 0.864 [0.795, 0.936]
& 0.893 [0.823, 0.969]
& +0.029 \\
\bottomrule
\end{tabular}
\caption{
Bootstrap confidence intervals for ImiScore in the main SFT and CAPO settings.
Each cell reports the bootstrap mean with the 95\% percentile interval in
brackets. CAPO obtains higher bootstrap mean ImiScore than SFT in all four
model-dataset settings.
}
\label{tab:imiscore-bootstrap}
\end{table*}

The bootstrap results support the main conclusion that CAPO improves
aggregation-aware annotator imitation. As shown in
Table~\ref{tab:imiscore-bootstrap}, CAPO achieves a higher bootstrap mean
ImiScore than SFT in all four model-dataset settings. The gains are especially
clear on R2, where CAPO improves Llama3.2 from 0.971 to 0.995 and Qwen3 from
0.934 to 0.973. The paired one-sided bootstrap tests provide additional evidence
for these R2 improvements, with $p=0.017$ for Llama3.2 and $p=0.024$ for Qwen3.
The VariErr gains are smaller but directionally consistent, suggesting that
CAPO's benefit is most reliable when annotator-specific behavior is more
separable at the group level.

Overall, this analysis strengthens the interpretation of the main results:
CAPO's advantage over SFT is most visible in aggregation-aware evaluation, where
annotator identity is assessed through repeated behavior rather than isolated
single-instance judgments.

\subsection{Large-Model Prompting Robustness}
\label{app:large_prompting}

The main controlled comparison uses Qwen3-4B and Llama3.2-3B because SFT and CAPO are trained from the same corresponding backbones.
To test whether the weaker prompting results under these models reflect prompting itself or model scale, we additionally evaluate Llama-3.1-70B and Qwen3-235B-A22B-Instruct-2507-FP8 under the same Base, ICL-50, and VP-ICL-50 prompting protocols.
No parameter adaptation is performed for these large-model experiments.

\begin{table*}[t]
\centering
\scriptsize
\setlength{\tabcolsep}{3.0pt}
\begin{tabular}{llccccccccc}
\toprule
& &
\multicolumn{4}{c}{VariErr} &
\multicolumn{5}{c}{R2} \\
\cmidrule(lr){3-6}
\cmidrule(lr){7-11}
Model & Method &
Label Acc $\uparrow$ &
GC Conf $\uparrow$ &
ImiScore $\uparrow$ &
Judge Acc $\uparrow$ &
Score Acc $\uparrow$ &
MAE $\downarrow$ &
GC Conf $\uparrow$ &
ImiScore $\uparrow$ &
Judge Acc $\uparrow$ \\
\midrule

Llama-3.1-70B
& Base
& 0.495 & 0.240 & -0.042 & 0.268
& 0.325 & 2.403 & 0.253 & 0.005 & 0.243 \\

& ICL-50
& 0.598 & 0.605 & 0.591 & 0.295
& 0.408 & 1.653 & 0.982 & 0.991 & 0.490 \\

& VP-ICL-50
& 0.610 & 0.725 & 0.687 & 0.305
& 0.398 & 1.685 & 0.953 & 0.961 & 0.465 \\

\midrule

Qwen3-235B
& Base
& 0.533 & 0.232 & -0.062 & 0.250
& 0.310 & 2.078 & 0.245 & -0.022 & 0.238 \\

& ICL-50
& 0.635 & 0.472 & 0.486 & 0.293
& 0.408 & 1.478 & 0.932 & 0.939 & 0.483 \\

& VP-ICL-50
& 0.655 & 0.542 & 0.566 & 0.283
& 0.430 & 1.423 & 0.971 & 0.981 & 0.515 \\

\bottomrule
\end{tabular}

\caption{
Large-model prompting robustness.
Llama-3.1-70B and Qwen3-235B are evaluated only with prompting and are not used for the controlled SFT/CAPO comparison.
Model scale substantially improves prompting and can outperform the adapted 3--4B models on some task metrics, but does not uniformly yield stronger annotator-specific imitation.
}
\label{tab:large_prompting}
\end{table*}

Results are in Table \ref{tab:large_prompting}.
Model scale materially improves prompting.
In particular, Qwen3-235B VP-ICL obtains VariErr label accuracy of $0.655$ and R2 MAE of $1.423$, outperforming the smaller adapted models on these task-level metrics.
However, on VariErr its GC Confidence ($0.542$), ImiScore ($0.566$), and Judge Accuracy ($0.283$) remain below the strongest adapted models.
On R2, large-model prompting closes much more of the imitation gap, with Qwen3-235B VP-ICL reaching GC Confidence $0.971$ and ImiScore $0.981$.
Its exact score accuracy ($0.430$) and Judge Accuracy ($0.515$), however, remain below the corresponding Qwen3-4B CAPO values ($0.455$ and $0.560$).

These results show that general task competence and faithful reproduction of a target annotator's repeated behavior are related but not identical.
They also motivate calibrating our prompting conclusion by model scale: sufficiently large prompted models can match or surpass adaptation along some dimensions, but do not do so uniformly across datasets and evaluation views.

\subsection{Target Alignment and Non-Target Leakage}
\label{app:target_leakage}

CAPO explicitly contrasts the target annotator with other valid annotators.
Its improvement could therefore arise from stronger alignment with the intended target, reduced confusion with alternatives, or both.
We decompose the group-classifier signal into the confidence assigned to the intended annotator on target-conditioned generations and the leakage toward that annotator from generations targeting the other annotators.

\begin{table*}[t]
\centering
\small
\begin{tabular}{lccccc}
\toprule
Setting &
\multicolumn{2}{c}{Target confidence} &
\multicolumn{2}{c}{Non-target leakage} &
$\Delta$ ImiScore \\
\cmidrule(lr){2-3}
\cmidrule(lr){4-5}
& SFT & CAPO & SFT & CAPO & \\
\midrule
Llama3.2--R2      & 0.873 & 0.916 & 0.091 & 0.075 & +0.050 \\
Qwen3--R2         & 0.840 & 0.882 & 0.091 & 0.086 & +0.048 \\
Llama3.2--VariErr & 0.699 & 0.697 & 0.138 & 0.137 & +0.008 \\
Qwen3--VariErr    & 0.703 & 0.729 & 0.138 & 0.130 & +0.050 \\
\bottomrule
\end{tabular}
\caption{
Decomposition of target alignment and non-target leakage.
CAPO reduces non-target leakage in all four settings and increases target confidence in three of four settings.
The resulting ImiScore change is positive in every setting.
}
\label{tab:target_leakage}
\end{table*}

As shown in Table \ref{tab:target_leakage},
CAPO therefore does not improve ImiScore solely by suppressing other annotators.
Target confidence increases in three of the four settings, while non-target leakage decreases in all four.
The exception is Llama3.2--VariErr, where target confidence is essentially unchanged and the overall improvement is correspondingly small.
These results support interpreting CAPO as a target-relative refinement that generally combines stronger target alignment with reduced confusion among plausible annotators.

\section{CAPO Ablations}
\label{app:capo-policy-ablation}

This appendix provides additional details for the CAPO ablations summarized in
Figure~\ref{fig:capo-policy-selection}. We study two design choices: the pair
policy used to construct CAPO preference pairs and the checkpoint-selection
criterion used after preference optimization.

\subsection{Pair-Policy Ablations}
\label{app:capo-pair-policy-ablation}

CAPO constructs preference pairs by comparing responses targeted at different
annotators. The pair policy controls how similar the paired examples must be in
their task-level labels or scores. This is important because overly permissive
pairs provide more contrastive supervision, but may also allow the model to learn
coarse label or score preferences instead of explanation-grounded,
annotator-specific behavior.

For VariErr, which uses categorical NLI labels, we consider three policies:
\begin{itemize}
    \item \textbf{strict}: the chosen and rejected responses must have the same
    task label. This is the main VariErr setting.
    \item \textbf{near-label}: the paired responses may differ by a small
    label-level discrepancy. This increases the number of available pairs, but
    introduces a weaker label-matching constraint than strict pairing.
    \item \textbf{no-restriction}: no label-matching constraint is applied when
    forming pairs. This provides the largest and strongest contrastive signal,
    but is also the most likely to introduce label-driven confounds.
\end{itemize}

Table~\ref{tab:app-varierr-capo-ablation} reports the VariErr pair-policy
ablation. The strict policy is used in the main experiments because it provides
the cleanest test of whether CAPO improves target-specific explanation framing
without relying on label differences.

\begin{table*}[ht]
\centering
\small
\begin{adjustbox}{max width=\textwidth}
\begin{tabular}{llrrrrrr}
\toprule
Model & Policy & Acc $\uparrow$ & R-L $\uparrow$ & Feature KL $\downarrow$ & GC Conf $\uparrow$ & ImiScore $\uparrow$ & Judge Acc $\uparrow$ \\
\midrule
\multirow{3}{*}{Qwen3}
& CAPO-strict & 0.627 & 0.281 & 0.081 & 0.867 & 0.888 & 0.328 \\
& CAPO-near-label & 0.580 & 0.272 & 0.068 & 0.907 & 0.940 & 0.302 \\
& CAPO-no-restriction & 0.575 & 0.269 & 0.064 & 0.918 & 0.955 & 0.343 \\
\midrule
\multirow{3}{*}{Llama3.2}
& CAPO-strict & 0.512 & 0.262 & 0.121 & 0.924 & 0.964 & 0.297 \\
& CAPO-near-label & 0.500 & 0.273 & 0.149 & 0.957 & 1.006 & 0.297 \\
& CAPO-no-restriction & 0.517 & 0.272 & 0.102 & 0.959 & 1.008 & 0.310 \\
\bottomrule
\end{tabular}
\end{adjustbox}
\caption{VariErr CAPO pair-policy ablation with train-only group-classifier
checkpoint selection. The main setting uses \textsc{CAPO-strict}, which restricts
pairs to the same task label and therefore minimizes label-driven contrast.}
\label{tab:app-varierr-capo-ablation}
\end{table*}

For R2, whose targets are ordinal scores rather than categorical NLI labels, we
use tolerance-based policies. Let $s_i$ and $s_j$ denote the two target scores in
a candidate pair. A tolerance-$k$ policy allows the pair only if
$|s_i - s_j| \leq k$:
\begin{itemize}
    \item \textbf{strict}: only exactly matched scores are paired, i.e.,
    $|s_i - s_j| = 0$.
    \item \textbf{tol1}: scores may differ by at most one point, i.e.,
    $|s_i - s_j| \leq 1$. This is the main R2 setting.
    \item \textbf{tol2}: scores may differ by at most two points, i.e.,
    $|s_i - s_j| \leq 2$.
    \item \textbf{tol3}: scores may differ by at most three points, i.e.,
    $|s_i - s_j| \leq 3$.
    \item \textbf{no-restriction}: no score-distance constraint is applied.
\end{itemize}

Table~\ref{tab:app-r2-capo-ablation} reports the R2 tolerance-policy ablation.
We use \textsc{CAPO-tol1} in the main experiments because exact score matching
can be too restrictive, while tolerance-1 remains the closest practical analogue
to same-label pairing.

\begin{table*}[ht]
\centering
\small
\begin{adjustbox}{max width=\textwidth}
\begin{tabular}{llrrrrrrr}
\toprule
Model & Policy & Acc $\uparrow$ & MAE $\downarrow$ & R-L $\uparrow$ & Feature KL $\downarrow$ & GC Conf $\uparrow$ & ImiScore $\uparrow$ & Judge Acc $\uparrow$ \\
\midrule
\multirow{5}{*}{Qwen3}
& CAPO-strict & 0.463 & 1.482 & 0.308 & 0.156 & 0.958 & 0.967 & 0.535 \\
& CAPO-tol1 & 0.455 & 1.490 & 0.316 & 0.112 & 0.964 & 0.973 & 0.560 \\
& CAPO-tol2 & 0.450 & 1.490 & 0.312 & 0.112 & 0.974 & 0.983 & 0.517 \\
& CAPO-tol3 & 0.453 & 1.488 & 0.309 & 0.113 & 0.979 & 0.988 & 0.532 \\
& CAPO-no-restriction & 0.448 & 1.532 & 0.310 & 0.151 & 0.992 & 1.002 & 0.545 \\
\midrule
\multirow{5}{*}{Llama3.2}
& CAPO-strict & 0.450 & 1.542 & 0.284 & 0.136 & 0.976 & 0.985 & 0.502 \\
& CAPO-tol1 & 0.460 & 1.470 & 0.293 & 0.187 & 0.985 & 0.995 & 0.530 \\
& CAPO-tol2 & 0.448 & 1.470 & 0.294 & 0.153 & 0.991 & 1.001 & 0.530 \\
& CAPO-tol3 & 0.438 & 1.480 & 0.303 & 0.150 & 0.990 & 1.000 & 0.527 \\
& CAPO-no-restriction & 0.415 & 1.790 & 0.283 & 0.149 & 0.996 & 1.006 & 0.520 \\
\bottomrule
\end{tabular}
\end{adjustbox}
\caption{R2 CAPO tolerance-policy ablation with train-only group-classifier
checkpoint selection. \textsc{CAPO-tol1} is the main setting: it allows target
scores to differ by at most one point and balances pair coverage with a relatively
tight score-matching constraint.}
\label{tab:app-r2-capo-ablation}
\end{table*}

Overall, Tables~\ref{tab:app-varierr-capo-ablation} and
\ref{tab:app-r2-capo-ablation} show the expected accuracy--imitation trade-off:
looser policies often increase GC Conf or ImiScore, but their interpretation is
less clean because the contrast may include larger task-level differences. We
therefore use the conservative policies in the main experiments.

\subsection{CAPO Checkpoint-Selection Ablations}
\label{app:capo-checkpoint-ablation}

CAPO produces multiple candidate checkpoints during preference optimization. For
the main protocol, we select checkpoints using a group classifier trained only on
human training annotations. For each candidate checkpoint, we generate
development outputs, group them by target annotator, and select the checkpoint
whose groups receive the highest target-annotator confidence. This follows the
aggregation-aware motivation in Section~\ref{sec:stable}: annotator-specific
behavior can be weak at the single-output level but more reliable when several
outputs are evaluated together. The selector does not use test outputs or any
classifier trained on test annotations. Additional details on group classifiers
are provided in Appendix~\ref{app:group-size-sweep}.

To check whether this selection criterion overly favors classifier-style
evaluation, we compare three checkpoint choices:
\begin{itemize}
    \item \textbf{classifier-selected}: the checkpoint selected by train-only
    group-classifier confidence. This is the main CAPO protocol.
    \item \textbf{dev-loss-selected}: the checkpoint with the best development
    loss during CAPO training.
    \item \textbf{last checkpoint}: the final checkpoint after CAPO training,
    without intermediate selection.
\end{itemize}

Table~\ref{tab:app-varierr-capo-checkpoint} reports the checkpoint-selection
ablation for VariErr under the main strict pair policy.

\begin{table*}[ht]
\centering
\small
\begin{adjustbox}{max width=\textwidth}
\begin{tabular}{llrrrrrr}
\toprule
Model & Selection & Acc $\uparrow$ & R-L $\uparrow$ & Feature KL $\downarrow$ & GC Conf $\uparrow$ & ImiScore $\uparrow$ & Judge Acc $\uparrow$ \\
\midrule
\multirow{3}{*}{Qwen3}
& classifier-selected & 0.627 & 0.281 & 0.081 & 0.867 & 0.888 & 0.328 \\
& dev-loss-selected & 0.625 & 0.282 & 0.066 & 0.877 & 0.902 & 0.312 \\
& last checkpoint & 0.627 & 0.281 & 0.081 & 0.867 & 0.888 & 0.320 \\
\midrule
\multirow{3}{*}{Llama3.2}
& classifier-selected & 0.512 & 0.262 & 0.121 & 0.924 & 0.964 & 0.297 \\
& dev-loss-selected & 0.507 & 0.259 & 0.098 & 0.944 & 0.989 & 0.273 \\
& last checkpoint & 0.507 & 0.260 & 0.113 & 0.942 & 0.988 & 0.292 \\
\bottomrule
\end{tabular}
\end{adjustbox}
\caption{VariErr checkpoint-selection ablation for the main strict CAPO policy.
The classifier-selected checkpoint is chosen using train-only group-classifier
confidence; dev-loss-selected and last checkpoints are included as
selection-sanity checks.}
\label{tab:app-varierr-capo-checkpoint}
\end{table*}

Table~\ref{tab:app-r2-capo-checkpoint} reports the corresponding
checkpoint-selection ablation for R2 under the main tolerance-1 pair policy.

\begin{table*}[ht]
\centering
\small
\begin{adjustbox}{max width=\textwidth}
\begin{tabular}{llrrrrrrr}
\toprule
Model & Selection & Acc $\uparrow$ & MAE $\downarrow$ & R-L $\uparrow$ & Feature KL $\downarrow$ & GC Conf $\uparrow$ & ImiScore $\uparrow$ & Judge Acc $\uparrow$ \\
\midrule
\multirow{3}{*}{Qwen3}
& classifier-selected & 0.455 & 1.490 & 0.316 & 0.112 & 0.964 & 0.973 & 0.560 \\
& dev-loss-selected & 0.453 & 1.490 & 0.312 & 0.134 & 0.967 & 0.976 & 0.520 \\
& last checkpoint & 0.455 & 1.490 & 0.316 & 0.112 & 0.964 & 0.973 & 0.530 \\
\midrule
\multirow{3}{*}{Llama3.2}
& classifier-selected & 0.460 & 1.470 & 0.293 & 0.187 & 0.985 & 0.995 & 0.530 \\
& dev-loss-selected & 0.465 & 1.438 & 0.296 & 0.149 & 0.981 & 0.991 & 0.520 \\
& last checkpoint & 0.460 & 1.470 & 0.293 & 0.187 & 0.985 & 0.995 & 0.527 \\
\bottomrule
\end{tabular}
\end{adjustbox}
\caption{R2 checkpoint-selection ablation for the main tolerance-1 CAPO policy.
The tolerance-1 policy allows paired target scores to differ by at most one
point.}
\label{tab:app-r2-capo-checkpoint}
\end{table*}

Tables~\ref{tab:app-varierr-capo-checkpoint} and
\ref{tab:app-r2-capo-checkpoint} show that classifier-selected checkpoints are
generally competitive with dev-loss-selected and last checkpoints, including
under judge-based evaluation. These results serve as sanity checks for the
selection protocol, while the main experiments use classifier-selected
checkpoints because the training objective is aggregation-aware
annotator-specific imitation.
\section{Additional Feature Shift Analysis}
\label{app:feature-shift}

\paragraph{Feature KL.}
We use Feature KL as a diagnostic measure of distributional alignment between human and generated explanations. 
For each feature $f$, we estimate the human feature distribution $p_f$ and the model feature distribution $q_f$ on the test set, and compute
\begin{equation}
    D_{\mathrm{KL}}(p_f \| q_f)
    =
    \sum_{b \in \mathcal{B}_f}
    p_f(b)
    \log \frac{p_f(b)}{q_f(b)} ,
\end{equation}
where $\mathcal{B}_f$ denotes the bins or support values for feature $f$. 
Continuous features, such as explanation length or overlap rate, are discretized into fixed bins, while binary features are computed over their two support values. 
We apply additive smoothing to avoid zero-probability bins. 
The reported mean Feature KL is the unweighted average over all features. 
Lower values indicate that generated explanations are closer to human explanations with respect to the corresponding feature distribution.

\paragraph{Overall results.}
Table~\ref{tab:appendix-feature-kl-overall} reports mean Feature KL for SFT and CAPO. 
CAPO improves mean Feature KL in the Qwen settings, especially on R2, but its effect is mixed for Llama. 
This suggests that CAPO does not simply optimize a generic notion of surface similarity; rather, its effect depends on the base model, dataset, and which annotator-specific properties are easiest to control.

\begin{table}[t]
\centering
\small
\begin{adjustbox}{max width=\columnwidth}
\begin{tabular}{llccc}
\toprule
Model & Dataset & SFT & CAPO & Rel. Change \\
\midrule
Llama & VariErr & 0.1131 & 0.1214 & $+7.3\%$ \\
Qwen  & VariErr & 0.0839 & 0.0806 & $-4.0\%$ \\
Llama & R2      & 0.1177 & 0.1875 & $+59.3\%$ \\
Qwen  & R2      & 0.1302 & 0.1118 & $-14.2\%$ \\
\bottomrule
\end{tabular}
\end{adjustbox}
\caption{Mean Feature KL between human and generated explanations. Lower is better. Negative relative change means CAPO reduces the feature distribution gap relative to SFT.}
\label{tab:appendix-feature-kl-overall}
\end{table}

\paragraph{Feature-wise improvements.}
Table~\ref{tab:appendix-feature-kl-breakdown} shows representative feature-wise improvements. 
The improved features span several interpretable groups: verbosity and structure, such as \texttt{len\_chars} and \texttt{avg\_sent\_len}; annotator or judgment language, such as \texttt{modal\_rate} and \texttt{judge\_rate}; 
and diversity, such as \texttt{type\_token\_ratio}. 
This supports the claim that CAPO can reduce SFT's feature-level distribution shift along multiple human-interpretable dimensions, rather than only matching length.

\begin{table}[t]
\centering
\small
\begin{adjustbox}{max width=\columnwidth}
\begin{tabular}{llccc}
\toprule
Setting & Feature & SFT KL & CAPO KL & Reduction \\
\midrule
Qwen-VariErr & \texttt{avg\_sent\_len}      & 0.1127 & 0.0233 & 0.0895 \\
Qwen-VariErr & \texttt{modal\_rate}         & 0.1795 & 0.1274 & 0.0521 \\
Qwen-VariErr & \texttt{judge\_rate}         & 0.0951 & 0.0524 & 0.0428 \\
Qwen-R2      & \texttt{len\_chars}          & 0.1691 & 0.0256 & 0.1435 \\
Qwen-R2      & \texttt{parentheses}         & 0.1940 & 0.0572 & 0.1368 \\
Qwen-R2      & \texttt{avg\_sent\_len}      & 0.1356 & 0.0479 & 0.0877 \\
Qwen-R2      & \texttt{type\_token\_ratio}  & 0.1394 & 0.1111 & 0.0283 \\
\bottomrule
\end{tabular}
\end{adjustbox}
\caption{Representative feature-wise KL reductions from SFT to CAPO. Lower KL means the generated distribution is closer to the human distribution for that feature.}
\label{tab:appendix-feature-kl-breakdown}
\end{table}

\paragraph{Regressions and limitations.}
Feature KL also reveals that CAPO's feature control is not uniformly positive. 
For example, Llama-R2 shows a worse mean Feature KL despite improvements on several individual features, because some discourse-marker features, such as \texttt{modal\_rate}, become less aligned with the human distribution. 
Similarly, CAPO can over-correct explanation length in some settings, making outputs too short or too long relative to human explanations. 
These regressions are important because they show that CAPO does not trivially optimize the Feature KL metric itself. 
Instead, CAPO mainly optimizes a relative preference objective that encourages target-annotator-specific outputs, while Feature KL serves as an external diagnostic of which explanation properties move closer to or farther from human behavior.

\paragraph{Surface-marker concern.}
A potential concern is that group-level classifier metrics may reward easily detectable markers, such as explanation length, quotation frequency, first-person usage, or modal words, rather than faithful reasoning imitation. 
Our analysis partially addresses this concern in two ways. 
First, the feature-wise improvements are distributed across multiple feature families, including sentence structure, modal and causal language, lexical diversity, and input grounding, rather than being concentrated only in length features. 
Second, the qualitative examples in Table~\ref{tab:case-study} show cases where CAPO better matches the gold annotator's reasoning focus, not merely its surface form.

Nevertheless, we treat GC Conf and ImiScore as diagnostic metrics rather than ground-truth measures of imitation. 
A stronger future test would retrain the group classifier after removing explicit surface markers such as \texttt{len\_chars}, \texttt{len\_words}, first-person indicators, modal words, and direct-quotation features. 
If CAPO continues to improve classifier confidence under this marker-ablated setting, it would provide stronger evidence that the model learns deeper reasoning-style regularities rather than only marker amplification. 
Another useful extension is label-conditioned Feature KL, where human and generated feature distributions are compared within the same gold label. 
This would help distinguish explanation-style alignment from indirect improvements caused by matching label priors.

\paragraph{Fluency and diversity.}
CAPO may also introduce a trade-off between style specificity and generation diversity. 
Because CAPO strengthens target-specific preferences, it could make outputs more templatic or repetitive. 
We therefore recommend reporting diversity diagnostics such as type-token ratio, unique-output ratio, repetition rate, or self-BLEU in addition to GC Conf and ImiScore. 
In our current feature analysis, CAPO improves \texttt{type\_token\_ratio} KL for Qwen-R2, suggesting that the gain in annotator specificity does not necessarily come from reduced lexical diversity in this setting. 
However, this pattern is not universal across all model-dataset combinations, so diversity should be treated as a complementary health check for CAPO-generated explanations.

\paragraph{Summary.}
Overall, Feature KL helps interpret what CAPO changes, while also exposing its failure modes. 
The main-text results show that CAPO improves group-level recognizability; the feature analysis shows which human-interpretable properties shift toward the gold distribution; and the case study verifies that these shifts can correspond to more faithful reasoning focus. 
This combination forms a more reliable evaluation loop than any single metric alone.
\section{Human Validation Protocol}
\label{app:human-validation-protocol}

Two annotators participated in the human validation. Both annotators are trained on the task instructions and paid according to national standards. The guideline explicitly instructs annotators not to judge only whether the predicted label matches the gold label. Instead, they are asked to focus on whether the generated explanation is coherent, grounded in the input pair, compatible with the gold explanation, and similar to the original annotator's style and perspective. A
generated explanation is marked as making sense when it is plausible for that
annotator, even if the predicted label is imperfect; it is left unmarked when it
is generic, contradicts the example, ignores key evidence, or fails to match the
annotator's reasoning style.

Each annotator independently annotated the same 50-sample subset from VariErr. We aggregate the binary ``makes sense'' decisions over samples, settings, and annotator profiles to compute the make-sense rate for each setting. 
On the 50-sample VariErr subset, both annotators produced the same ranking: CAPO $>$ SFT $>$ prompt. The overall raw percent agreement between the two annotators was 0.828, computed over $50 \times 3 \times 4 = 600$ binary decisions.

To assess inter-annotator agreement, we further compute Cohen's $\kappa$ between the two annotators. Overall agreement is substantial ($\kappa=0.650$). Per-setting raw agreement and $\kappa$ values are reported in Table~\ref{tab:human-validation-iaa}.

\begin{table}[t]
\centering
\begin{adjustbox}{max width=\columnwidth}
\begin{tabular}{lcccc}
\toprule
Annotator & CAPO & SFT & Prompt & Overall ranking \\
\midrule
Annotator 1 & 45.5\% & 43.5\% & 27.0\% & CAPO $>$ SFT $>$ Prompt \\
Annotator 2 & 46.0\% & 45.5\% & 45.0\% & CAPO $>$ SFT $>$ Prompt \\
\bottomrule
\end{tabular}
\end{adjustbox}
\caption{Human validation results on the 50-sample VariErr subset. Each cell reports the make-sense rate, computed over $50 \times 4 = 200$ binary decisions for each setting.}
\label{tab:human-validation-makesense}
\end{table}

\begin{table}[t]
\centering
\begin{adjustbox}{max width=\columnwidth}
\begin{tabular}{lccc}
\toprule
Setting & Agreement/Total decisions & Agreement & Cohen's $\kappa$ \\
\midrule
CAPO & 167/200 & 83.5\% & 0.668 \\
SFT & 174/200 & 87.0\% & 0.737 \\
Prompt & 156/200 & 78.0\% & 0.539 \\
\midrule
Overall & 497/600 & 82.8\% & 0.650 \\
\bottomrule
\end{tabular}
\end{adjustbox}
\caption{Inter-annotator agreement for the human validation study. Agreement and Cohen's $\kappa$ are computed over binary make-sense decisions. The overall score aggregates across 50 samples, three settings, and four original annotators.}
\label{tab:human-validation-iaa}
\end{table}

\subsection{Case study on R2}
\label{app:case_study_r2}

To accompany the VariErr examples shown in the main text, Table~\ref{tab:case-study-r2} presents two additional examples from R2. These examples illustrate the same pattern in a semantic similarity setting. The gold explanations often rely on fine-grained similarity rationales: two questions may share a broad topic or object, while differing in intent, specificity, or the aspect being queried. SFT tends to collapse such cases into generic judgments such as ``different questions'' or ``conceptually unrelated,'' which misses the partial overlap emphasized by the original annotator. In contrast, CAPO more often preserves both parts of the rationale: it identifies the shared domain (e.g., death-related questions or iPhone-related questions) while also noting the relevant distinction between the two queries. 

\begin{table*}[t]
\centering
\small
\renewcommand{\arraystretch}{1.25}
\begin{adjustbox}{max width=\textwidth}
\begin{tabular}{p{0.02\textwidth} p{0.13\textwidth} p{0.13\textwidth} p{0.20\textwidth} p{0.25\textwidth} p{0.25\textwidth}}
\toprule
Case & Q1 & Q2 & Gold Explanation & SFT Explanation & CAPO Explanation \\
\midrule

\textbf{1}
& How much time it takes to die by hanging?
& Can you die by forcing yourself to stay awake for too long?
& Both Q deal with reasons for death and are domain specific; syntax differ.
& The queries are conceptually unrelated. The first query is a request for a time estimate, the second is a request for a yes/no answer.
& Q1 is a specific query, Q2 is a general query. Both are related to death. \\
\multicolumn{6}{p{\textwidth}}{
\textcolor{gray}{\textit{\textbf{Observation:} SFT over-separates the pair by treating the questions as unrelated and focusing on surface question type. CAPO better matches the gold explanation by preserving the shared death-related domain while still noting their different specificity.}}
} \\
\midrule

\textbf{2}
& How accurate is the iPhone as a pedometer?
& How good is the iPhone 6?
& Q2 is a more general question. Q1 is more specificly asking about iphone as pedometer.
& Two different questions.
& Q1 is about the iPhone's pedometer, while Q2 is about the iPhone 6. \\

\multicolumn{6}{p{\textwidth}}{
\textcolor{gray}{\textit{\textbf{Observation:} SFT reduces the comparison to a generic ``different questions'' judgment. CAPO better captures the gold distinction: both questions concern the iPhone, but Q1
targets a specific function while Q2 asks about the product more generally.}}
} \\
\bottomrule
\end{tabular}
\end{adjustbox}
\caption{
Case study comparing SFT and CAPO on R2. CAPO better preserves the gold annotator's similarity rationale by capturing shared domains and fine-grained differences, while SFT tends to over-separate question pairs or collapse them into generic difference judgments.
}
\label{tab:case-study-r2}
\end{table*}

\begin{table*}[t]
\centering
\small
\begin{adjustbox}{max width=0.98\textwidth}
\begin{tabular}{p{0.07\textwidth} p{0.29\textwidth} p{0.28\textwidth} p{0.30\textwidth}}
\toprule
Annotator & Behavioral interpretation & Empirical cues & Representative examples \\
\midrule
\rowcolor{gray!15}
\multicolumn{4}{>{\centering\arraybackslash}p{\textwidth}}{\textbf{VariErr NLI}} \\

0 &
Evidence-checking and context-grounded: treats the task as verifying whether the hypothesis is explicitly licensed by the context. &
Longer explanations; frequent meta-level NLI terms such as ``context,'' ``statement,'' and ``mention.'' &
- Both the \textit{context} and the \textit{statement} suggest that the speaker believes what was said.
\par
- The \textit{context mentions} that activities with a longer time frame than 12 months will be included, so some activities could be shorter than 12 months. \\
\midrule
1 &
Conservative under underspecification: withholds commitment when a specific detail is not stated. &
Concise absence-of-evidence formulations, especially ``not clear whether'' and ``does not say anything about.'' &
- \textit{It is not clear whether} the speaker wants the taxes to be lower.
\par
- \textit{The context does not say anything about} lepers or where they could work. \\
\midrule
2 &
Possibility-oriented and ambiguity-preserving: keeps multiple compatible readings open instead of forcing a deterministic inference. &
High modal/first-person tendency; frequent use of ``could,'' ``maybe,'' and ``we do not know.'' &
- The man \textit{could} be grabbed by Jon, \textit{but also could} slip from him.
\par
- \textit{We do not know} how often a Global Expo is held. \\
\midrule
3 &
Compressed and diagnostic: uses explanations as short decision cues rather than full rationales. &
Short explanations; repeated minimal phrases such as ``irrelevant,'' ``paraphrases,'' and ``no info about \ldots''. &
- No info about the murder.
\par
- Irrelevant. \\

\rowcolor{gray!15}
\multicolumn{4}{>{\centering\arraybackslash}p{\textwidth}}{\textbf{R2}} \\

0 &
Aspect-level question comparison: contrasts what Q1 and Q2 ask about, especially conditions, degree, or requested information. &
Explicit Q1--Q2 contrast, often using ``Q1 asks \ldots whereas Q2 asks \ldots''. &
- \textit{Q1 asks} whether sex is important, \textit{whereas Q2 asks} how important it is.
\par
- \textit{In Q2}, the prerequisite, knowing C and C++, is given, \textit{whereas in Q1} there is no prerequisite. \\
\midrule
1 &
Coarse topic-relation categorization: reduces the comparison to topic sameness, topic difference, or semantic-field overlap. &
Shortest and most formulaic explanations; recurrent labels such as ``same topic,'' ``different topics,'' and ``same semantic field.'' &
- \textit{Different topics}, although both are in the same semantic field, namely MIT.
\par
- \textit{The same topic}. \\
\midrule
2 &
Analytical query-relevance comparison: weighs phrase overlap, domain, template effects, and fine-grained differences. &
Frequent input reuse and quotation; markers such as ``common\_phrase,'' ``Difference in,'' and ``boilerplate template.'' &
- \textit{common\_phrase}: sleep cycle; Difference in: fix--calculate can be neglected.
\par
- Calculation request for both Qs where context is specified for the problem. The problems themselves differ. No penalty since \textit{``boilerplate template''}. \\
\midrule
3 &
Answer-equivalence and keyword-sensitive reasoning: judges whether two questions would lead to the same answer. &
Broad semantic judgments such as ``similar meaning'' or ``different meaning,'' combined with answer-changing keyword contrasts. &
- They have a \textit{similar meaning}.
\par
- \textit{Different keywords}, ``goherenext.com'' and ``thedartmouth.com,'' can lead to entirely different answers. \\

\bottomrule
\end{tabular}
\end{adjustbox}
\caption{Qualitative characterization of annotator-level explanation behavior. Behavioral interpretations describe recurring annotation strategies, not demographic or psychological attributes. Empirical cues are summarized from feature statistics, value-profile summaries, and manual inspection of human explanations; examples are lightly normalized for readability.}
\label{tab:annotator-behavioral-profiles}
\end{table*}

\subsection{Profile Study Details}
\label{app:profile_study_details}

Table~\ref{tab:annotator-behavioral-profiles} summarizes the behavioral profiles for the eight annotators across the two datasets. These profiles connect closely to the handcrafted explanation-style features shown in Table~\ref{tab:feature-kl-breakdown}, and the LLM-generated profiles in Appendix~\ref{app:value-profiles}, which help us to better understand and characterize annotator behavior preferences.

It is recognizable that different annotators reuse different levels of detail from the input, different lexical choices, and also different ways of grounding their reasoning. In VariErr for example, annotator 0 gives the longest explanation in average with relatively detailed, context-grounded explanations using vocabulary such as ``context,'' ``statement,'' and ``mention.'' In contrast, annotator 3 is highly compressed, often giving short diagnostic phrases such as ``irrelevant,'' ``paraphrases,'' or ``no info about.''.

In R2, the four annotator-profiles reflect different ways of explaining similarity and relations between the two given sentences. For instance, annotator 0 often writes explicit comparisons of Q1 and Q2, contrasting what each question asks. Annotator 2, on the other hand, gives more analytic-style explanations, frequently mentioning ``domain,'' ``overlap,'' ``specific,'' or exact shared phrases.

These differences further motivate our use of feature-level diagnostics: a model should not only provide explanations that are reasonable and plausible when imitating human annotators, but also focus on the target annotator's characteristic explanation style and reasoning focus. The stronger alignment of CAPO with these feature profiles suggests that it better captures annotator-specific explanatory behavior, rather than producing simply more fluent or generically ``high quality'' rationales.

\section{Value Profiles}
\label{app:value-profiles}

This appendix records the value/style profiles used by the VP and VP-ICL baselines. Each profile was generated from 50 target-annotator training examples with seed 42 and the direct profile-construction method described in Appendix~\ref{app:prompts}.

\subsection{VariErr Value Profiles}

\subsubsection{Llama3.2 profiles}

These profiles correspond to profile model family: \texttt{llama32}, sample size: 50, seed: 42.

\begin{promptbox}[VariErr Llama3.2 Annotator 0 value profile]{PromptBlueFrame}
Based on the provided examples, here is a compact profile of Annotator_0's annotation behavior:

**Label/Score Bias:**

* Tends to label as "Contradiction" when the context and statement have opposing meanings.
* Labels as "Entailment" when the context implies a logical connection between the premise and hypothesis.
* Labels as "Neutral" when the context is unclear or lacks sufficient information to make a conclusion.

**Value Judgment Criteria:**

* Evaluates statements for implicit value judgments, such as "beautiful" or "hardworking".
* May not always recognize value judgments, especially if they are implicit or nuanced.

**Evidence Usage:**

* Requires explicit evidence to support a conclusion, such as quotes or specific examples.
* May not consider implicit evidence, such as tone or context.

**Explanation Style:**

* Provides concise explanations that summarize the reasoning behind the label.
* May use phrases like "the context only mentions" or "it is unclear if" to indicate uncertainty.

**Typical Wording:**

* Uses phrases like "the context only mentions" or "it is unclear if" to indicate uncertainty.
* May use phrases like "the statement implies" or "the context suggests" to indicate logical connections.

**Strictness:**

* Can be quite strict in its labeling, especially when it comes to "Contradiction" and "Entailment".
* May be more lenient when it comes to "Neutral" labels.

**Uncertainty:**

* Recognizes uncertainty when the context is unclear or lacks sufficient information.
* May use phrases like "it is unclear if" or "the context only mentions" to indicate uncertainty.

**Quirks:**

* May not always recognize implicit evidence, such as tone or context.
* Can be quite strict in its labeling, especially when it comes to "Contradiction" and "Entailment".
* May use phrases like "the statement implies" or "the context suggests" to indicate logical connections.

To imitate Annotator_0, follow these guidelines:

1. Be cautious when labeling as "Contradiction" or "Entailment", as these labels require explicit evidence and logical connections.
2. Use phrases like "the context only mentions" or "it is unclear if" to indicate uncertainty.
3. Provide concise explanations that summarize the reasoning behind the label.
4. Be strict in your labeling, especially when it comes to "Contradiction" and "Entailment".
5. Recognize implicit evidence, such as tone or context, when possible.
6. Use phrases like "the statement implies" or "the context suggests" to indicate logical connections.
\end{promptbox}

\begin{promptbox}[VariErr Llama3.2 Annotator 1 value profile]{PromptBlueFrame}
**Annotation Style Profile: Annotator_1**

**Label/Score Bias:**

* Tends to avoid making definitive judgments, often labeling as "Neutral" when the context suggests a stronger stance.
* May not always provide a clear label, especially when the context is ambiguous or open to multiple interpretations.

**Value Judgment Criteria:**

* Focuses on literal meaning and context, rather than making value judgments about the statements.
* Avoids making assumptions or inferences that may not be supported by the text.

**Evidence Usage:**

* Typically provides a concise explanation to support the label, but may not always provide explicit evidence.
* May use paraphrasing or restatement of the context to support the label.

**Explanation Style:**

* Uses a formal, objective tone, avoiding emotive language or personal opinions.
* Provides a clear and concise explanation, but may not always elaborate on the reasoning.

**Typical Wording:**

* Uses straightforward language, avoiding jargon or technical terms unless necessary.
* May use phrases like "it's not clear" or "it's not clear whether" to indicate uncertainty.

**Strictness:**

* Tends to be neutral or cautious in their labeling, avoiding strong or absolute statements.
* May not always provide a clear label, especially when the context is ambiguous or open to multiple interpretations.

**Uncertainty:**

* Often indicates uncertainty or ambiguity in the context, using phrases like "it's not clear" or "it's not clear whether".
* May not always provide a clear label, especially when the context is ambiguous or open to multiple interpretations.

**Quirks:**

* May use phrases like "it's not clear" or "it's not clear whether" to indicate uncertainty.
* Tends to avoid making value judgments or assumptions, focusing on literal meaning and context.
* May not always provide explicit evidence to support the label, relying on paraphrasing or restatement of the context instead.

**Imitation Guidelines:**

To imitate Annotator_1's style, follow these guidelines:

1. Use a formal, objective tone and avoid emotive language or personal opinions.
2. Focus on literal meaning and context, avoiding value judgments or assumptions.
3. Provide a clear and concise explanation to support the label, but avoid elaborating on the reasoning unless necessary.
4. Use straightforward language and avoid jargon or technical terms unless necessary.
5. Be cautious in labeling, avoiding strong or absolute statements.
6. Indicate uncertainty or ambiguity in the context using phrases like "it's not clear" or "it's not clear whether".
7. Avoid making value judgments or assumptions, focusing on literal meaning and context instead.
\end{promptbox}

\begin{promptbox}[VariErr Llama3.2 Annotator 2 value profile]{PromptBlueFrame}
**Annotator_2's Annotation Behavior Profile**

**Label/Score Bias:**

* Tends to label as "Contradiction" when the premise and hypothesis are mutually exclusive, but not necessarily contradictory in a strict sense.
* Labels as "Entailment" when the premise implies a logical consequence, but may not be a direct entailment.
* Labels as "Neutral" when the premise and hypothesis are ambiguous or lack clear implications.

**Value Judgment Criteria:**

* Evaluates the strength of the premise and hypothesis based on logical coherence and implications.
* May consider the context and tone of the text when making judgments.

**Evidence Usage:**

* Relys on explicit statements and logical implications to support judgments.
* May consider implicit assumptions or unstated premises.

**Explanation Style:**

* Provides concise and clear explanations for judgments.
* May use phrases like "usually," "typically," or "in context" to clarify reasoning.

**Typical Wording:**

* Uses phrases like "usually," "typically," or "in context" to clarify reasoning.
* May use simple, direct language to explain judgments.

**Strictness:**

* May be lenient in cases where the premise and hypothesis are ambiguous or lack clear implications.
* May be more strict in cases where the premise and hypothesis are mutually exclusive.

**Uncertainty:**

* May indicate uncertainty or ambiguity in judgments, especially when the premise and hypothesis are unclear or lack clear implications.

**Quirks:**

* May over-interpret or over-analyze text, leading to overly complex or nuanced judgments.
* May be overly cautious in cases where the premise and hypothesis are ambiguous or lack clear implications.

**Imitation Guidelines:**

To imitate Annotator_2's annotation behavior, follow these guidelines:

1. Evaluate the strength of the premise and hypothesis based on logical coherence and implications.
2. Consider the context and tone of the text when making judgments.
3. Use explicit statements and logical implications to support judgments.
4. Be concise and clear in explanations, using phrases like "usually," "typically," or "in context" to clarify reasoning.
5. Be lenient in cases where the premise and hypothesis are ambiguous or lack clear implications.
6. Indicate uncertainty or ambiguity in judgments when the premise and hypothesis are unclear or lack clear implications.
7. Avoid over-interpreting or over-analyzing text, and be cautious in cases where the premise and hypothesis are ambiguous or lack clear implications.
\end{promptbox}

\begin{promptbox}[VariErr Llama3.2 Annotator 3 value profile]{PromptBlueFrame}
Based on the provided examples, here is a compact profile of Annotator_3's annotation behavior:

**Label/Score Bias:**

* Tends to label as "Entailment" when the context implies a positive relationship between two statements.
* Labels as "Contradiction" when the context implies a direct opposition between two statements.
* Neutral labels are used when the context is unclear or when the annotator is unsure of the relationship between the statements.

**Value Judgment Criteria:**

* Evaluates statements based on their logical coherence and implied relationships.
* Does not make explicit value judgments, but rather focuses on identifying the relationships between statements.

**Evidence Usage:**

* Uses context to infer relationships between statements.
* Does not require explicit evidence to support the annotation.

**Explanation Style:**

* Provides concise explanations that summarize the relationship between the statements.
* Uses simple language and avoids complex explanations.

**Typical Wording:**

* Uses phrases like "entails", "contradicts", and "neutral" to describe the relationships between statements.
* Avoids using technical jargon or complex terminology.

**Strictness:**

* Tends to be lenient in its annotation, allowing for some ambiguity in the context.
* Does not require absolute certainty in the relationships between statements.

**Uncertainty:**

* Indicates uncertainty when the context is unclear or when the annotator is unsure of the relationship between the statements.
* Uses phrases like "No info" or "irrelevant" to indicate uncertainty.

**Quirks:**

* May misinterpret certain phrases or idioms, leading to incorrect annotations.
* Can be overly literal in its interpretation of statements, leading to incorrect conclusions.
* Does not account for nuances in language, such as sarcasm or irony.

By following this profile, another LLM can attempt to imitate Annotator_3's annotation behavior on new items. However, it is essential to note that this profile is based on a limited dataset and may not capture the full complexity of Annotator_3's annotation behavior.
\end{promptbox}

\subsubsection{Qwen3 profiles}

These profiles correspond to profile model family: \texttt{qwen3}, sample size: 50, seed: 42.

\begin{promptbox}[VariErr Qwen3 Annotator 0 value profile]{PromptGreenFrame}
**Annotator_0 Profile (Imitation Guide)**

- **Label/Score Bias**: Strongly avoids *Entailment* when evidence is weak or indirect; favors *Contradiction* when the hypothesis introduces a false or incompatible claim (e.g., misreading negation, overreach, or false inference). *Neutral* is default when information is missing, ambiguous, or contextually unrelated.

- **Value Judgment Criteria**: Does **not** apply value judgments independently. Labels are based strictly on **literal content**, **semantic consistency**, and **explicit or implied factual alignment**. No moral, emotional, or ideological evaluation is introduced.

- **Evidence Usage**:
  - Requires **explicit or directly implied** support for *Entailment* or *Contradiction*.
  - Rejects assumptions (e.g., "people might say", "it could be", "it seems like").
  - Flags **missing elements** (e.g., "no mention of", "not stated", "unclear") as grounds for *Neutral*.
  - Uses **semantic precision** (e.g., "interim rules" vs "general notices") to detect contradictions.

- **Explanation Style**:
  - Concise, direct, and **fact-based**.
  - Uses **logical reasoning** to trace from premise to label.
  - Often includes **explicit negation** or **contrast** (e.g., "not mentioned", "means X, not Y").
  - Avoids speculation or hedging beyond necessary qualifiers.

- **Typical Wording**:
  - "The context only mentions X, has no information about Y."
  - "There is no mention of Z."
  - "It is unclear if A or B."
  - "The context states that X, but the statement says Y."
  - "Means X, so Y is false."
  - "The statement suggests Z, but the context says W."
  - "It is not clear if A is true."

- **Strictness**:
  - Very strict on **semantic precision** (e.g., "interim" vs "general notices", "national" vs "regional").
  - Rejects **overgeneralization** or **inference beyond the text**.
  - Treats **modalities** (e.g., "might", "could", "possibly") as insufficient for entailment.

- **Uncertainty Handling**:
  - Uses *Neutral* when:
    - The hypothesis introduces new claims not in the premise.
    - The premise lacks sufficient detail to support entailment.
    - The meaning is ambiguous (e.g., "protest", "hardworking", "crazy").
  - Avoids *Uncertain* or *Possibly* labels -- only *Neutral* or *Contradiction/Entailment*.

- **Quirks**:
  - Frequently **rephrases the premise** in the explanation to highlight mismatch.
  - Uses **self-reflective language** ("I am asking myself", "it is unclear") to signal hesitation.
  - Treats **questions** or **statements of inquiry** as *Neutral* unless clearly entailed.
  - Strongly **rejects inference from tone or implication** (e.g., "splendid" -> excited) unless explicitly tied
\end{promptbox}

\begin{promptbox}[VariErr Qwen3 Annotator 1 value profile]{PromptGreenFrame}
**Annotator_1 Profile (Imitation Guide)**

- **Label/Score Bias**:
  Strongly avoids over-entailment; defaults to *Neutral* when meaning is ambiguous, incomplete, or requires inference beyond direct textual support. *Entailment* is only assigned when the hypothesis is a clear paraphrase, logical consequence, or direct restatement. *Contradiction* is used when explicit negation or factual incompatibility exists. *Neutral* is the default for uncertainty or missing information.

- **Value Judgment Criteria**:
  Does **not** assign value judgments based on sentiment or moral stance. Instead, evaluates logical consistency and textual support. If a statement asserts a value (e.g., "stealing is bad"), it is only labeled *Entailment* if the premise explicitly supports it. Otherwise, it is *Neutral*.

- **Evidence Usage**:
  Relies **strictly** on explicit or directly inferable content. Does **not** introduce external knowledge, assumptions, or interpretive leaps. If a claim is not directly supported or contradicted, it is labeled *Neutral*. Even plausible inferences (e.g., "wallop" ~= "hit with a spoon") are only accepted if the similarity is clearly grounded in the text.

- **Explanation Style**:
  Concise, neutral tone. Uses **factual, observational language**. Avoids emotional or evaluative phrasing. Structure:
  - Label first.
  - Explanation: "It is not clear..." / "The context does not say..." / "That indicates..." / "It is a paraphrase of..."
  - Avoids speculative or hypothetical language.

- **Typical Wording**:
  - "It is not clear whether..."
  - "The context does not say..."
  - "That indicates..." / "It implies..."
  - "It is a paraphrase of..."
  - "It's clearly stated that..."
  - "It's not clear how..."
  - "The speaker says that..."

- **Strictness**:
  High. Rejects entailment when the connection is indirect, inferential, or based on interpretation. Requires **direct textual alignment** or **explicit contradiction**. Even subtle differences (e.g., "best example" vs. "good example") are treated as entailment only if clearly supported.

- **Uncertainty Handling**:
  Explicitly acknowledges uncertainty using phrases like "It is not clear", "It's not clear whether", or "It's not clear how". Never assumes missing information is implied or can be reasonably inferred.

- **Quirks**:
  - Treats **paraphrasing** as *Entailment* if it's a direct rewording.
  - Treats **negation** or **contradiction** as *Contradiction* when explicitly stated.
  - Uses **"it is not clear"** as a default when meaning is ambiguous.
  - Occasionally interprets **intensity** (e.g., "wallop" vs. "hit with spoon") as equivalent if contextually plausible.
  - Avoids over-interpretation of emotional or rhetorical tone.
  - Labels *Neutral* when the premise and hypothesis are unrelated or only partially aligned.



**Imitation Rule Summary**:
When in doubt, label **Neutral**. Only label
\end{promptbox}

\begin{promptbox}[VariErr Qwen3 Annotator 2 value profile]{PromptGreenFrame}
**Annotator_2 Profile (Imitation Guide)**

- **Label/Score Bias**:
  Strongly favors *Contradiction* when the hypothesis directly conflicts with explicit or implied facts (e.g., mismatched terms like "national" vs. "regional", or logical impossibility). Favors *Entailment* when the hypothesis is a plausible, direct inference (e.g., "made beautiful" -> "positive view of violence"). *Neutral* is default when uncertainty, missing context, or multiple interpretations exist.

- **Value Judgment Criteria**:
  Rejects value-laden or emotionally charged inferences (e.g., "crazy", "great", "flawed") unless directly supported. Avoids assuming intent, emotion, or moral stance. Labels like "neutral" are used when the premise doesn't justify a positive/negative evaluation.

- **Evidence Usage**:
  Relies on **literal, contextual, and grammatical alignment**. Prioritizes **explicit terms** (e.g., "national" vs. "regional") and **logical consistency** over implied or metaphorical meanings. Ignores out-of-scope or unmentioned details (e.g., cost, location, actor vs. character).

- **Explanation Style**:
  Concise, direct, and **cautious**. Uses **"we don't know"**, **"maybe"**, **"could be"**, or **"not definitely"** to express uncertainty. Avoids speculative or emotional language. Explains *why* a label is chosen through **logical or factual mismatch**, not emotional or philosophical reasoning.

- **Typical Wording**:
  - "We don't know..."
  - "Maybe it is..." / "could be"
  - "No, because..." / "It is not true because..."
  - "In context, there is nothing about..."
  - "He could be sure of X, or Y"
  - "It is possible, because..."
  - "No, the rules were issued as interim rules."

- **Strictness**:
  High. Rejects entailment when the connection is not direct or unambiguous. Requires **explicit support** or **clear logical necessity**. Even plausible inferences are labeled *Neutral* if the premise doesn't fully justify the conclusion.

- **Uncertainty Handling**:
  Systematically uses *Neutral* when:
  - Context lacks information (e.g., location, cost, intent)
  - Multiple interpretations are possible
  - The hypothesis introduces new elements not in the premise
  - Ambiguity in pronouns, references, or emotional tone

- **Quirks**:
  - Frequently uses **"No"** to reject contradictions, even in neutral contexts.
  - Treats **"only"**, **"best"**, **"must"**, **"flawed"**, or **"crazy"** as red flags for contradiction or neutrality.
  - Misinterprets **character vs. actor** (e.g., "George Clooney" as character -> neutral).
  - Assumes **"positive" or "negative" connotations** only when explicitly stated.
  - Over-uses **"we don't know"** or **"maybe"** even when the premise is clear.
  - Labels **"entailment"**
\end{promptbox}

\begin{promptbox}[VariErr Qwen3 Annotator 3 value profile]{PromptGreenFrame}
**Annotator_3 Profile (Imitation Guide)**

- **Label/Score Bias**:
  Strongly avoids over-entailment or over-contradiction. Prefers *Neutral* when information is missing, ambiguous, or contextually insufficient. Only assigns *Entailment* or *Contradiction* when clear logical or semantic alignment/disagreement exists. Avoids assuming intent, emotion, or value judgments.

- **Value Judgment Criteria**:
  Does **not** apply value judgments (e.g., "crazy," "flawed," "good," "bad") in labels. Labels are strictly based on factual, logical, or semantic consistency. Any value-laden language in explanation is used only to clarify meaning, not to justify a label.

- **Evidence Usage**:
  Relies **exclusively** on explicit or directly inferable information from the premise. Rejects assumptions, inferences beyond the text, or external knowledge. Flags gaps with "No info," "irrelevant," or "not mentioned."

- **Explanation Style**:
  Concise, direct, and grounded in linguistic or logical analysis. Uses plain language, avoids jargon. Often includes **paraphrase recognition** ("paraphrases"), **semantic equivalence**, or **word-level reasoning** (e.g., "budge entails that the door wouldn't move").
  Frequently includes **explicit negation** or **contradiction of implication** (e.g., "cannot be outlawed" = "cannot be made illegal").

- **Typical Wording**:
  - "No info about..."
  - "Not mentioned"
  - "It is not known"
  - "Paraphrases"
  - "Entails X because Y"
  - "Contradiction because Z"
  - "No info about where he is hardworking or not"
  - "Can not conclude that..."
  - "The context is talking about..."

- **Strictness**:
  Very strict. Rejects entailment when the hypothesis introduces new claims not supported by the premise. Requires **direct logical or semantic connection**. Even subtle mismatches (e.g., "only" vs. "also") trigger *Contradiction* or *Neutral*.

- **Uncertainty Handling**:
  Explicitly acknowledges uncertainty with phrases like:
  - "No info"
  - "Not sure"
  - "Can not conclude"
  - "The context is talking about..."
  Never uses vague or probabilistic language (e.g., "possibly," "might").

- **Quirks & Idiosyncrasies**:
  - **High sensitivity to word precision**: e.g., "only" vs. "also," "best way" vs. "the only way."
  - **Strong rejection of over-interpretation**: e.g., "crazy" -> "no info about beauty," "excited" -> "entails excitement" only if clearly implied.
  - **Frequent use of synonym mapping** (e.g., "generally accepted" = "large acknowledged").
  - **Rejects emotional or subjective inferences** (e.g., "he is hardworking" -> no info).
  - **Treats questions as neutral** (e.g., "is it true?" -> neutral).
  - **
\end{promptbox}

\subsection{R2 Value Profiles}

\subsubsection{Llama3.2 profiles}

These profiles correspond to profile model family: \texttt{llama32}, sample size: 50, seed: 42.

\begin{promptbox}[R2 Llama3.2 Annotator 0 value profile]{PromptPurpleFrame}
**Label/Score Bias:**
Annotator_0 tends to assign high scores (4-5) when the questions are paraphrases of each other, indicating a strong bias towards recognizing similar questions. Conversely, when the questions are not paraphrases (scores -5), Annotator_0 tends to assign lower scores, indicating a lack of understanding or recognition of the similarity between the questions.

**Value Judgment Criteria:**
Annotator_0's value judgment criteria seem to be based on the similarity in wording and structure between the questions. They tend to prioritize questions that have similar phrasing, sentence structure, and context.

**Evidence Usage:**
Annotator_0's evidence usage is limited to recognizing the similarity between the questions. They do not provide additional context or evidence to support their answers.

**Explanation Style:**
Annotator_0's explanation style is concise and to the point, often focusing on the similarity between the questions. They tend to avoid providing additional context or background information.

**Typical Wording:**
Annotator_0's typical wording is straightforward and lacks embellishments. They tend to use simple language to convey their answers.

**Strictness:**
Annotator_0's strictness is evident in their tendency to assign high scores only when the questions are paraphrases. They seem to be quite strict in their evaluation of question similarity.

**Uncertainty:**
Annotator_0's uncertainty is evident in their tendency to assign lower scores when the questions are not paraphrases. However, they also seem to be uncertain when the questions are similar but not identical.

**Quirks:**
Annotator_0's quirks include:

* A tendency to assign high scores only when the questions are paraphrases
* A focus on the similarity in wording and structure between the questions
* A lack of evidence usage beyond recognizing question similarity
* A concise explanation style that lacks additional context or background information
* A strict evaluation of question similarity
* Uncertainty when questions are similar but not identical

**Profile:**
Based on the analysis, Annotator_0's annotation behavior can be summarized as follows:

* Strong bias towards recognizing paraphrases (scores 4-5)
* Focus on similarity in wording and structure between questions
* Limited evidence usage beyond recognizing question similarity
* Concise explanation style that lacks additional context or background information
* Strict evaluation of question similarity
* Uncertainty when questions are similar but not identical

To imitate Annotator_0's annotation behavior, it is recommended to:

* Focus on the similarity in wording and structure between questions
* Assign high scores only when the questions are paraphrases
* Use a concise explanation style that lacks additional context or background information
* Evaluate question similarity strictly
* Be uncertain when questions are similar but not identical
\end{promptbox}

\begin{promptbox}[R2 Llama3.2 Annotator 1 value profile]{PromptPurpleFrame}
Based on the provided examples, here is a compact profile of Annotator_1's annotation behavior:

**Label/Score Bias:**
Annotator_1 tends to assign high scores (4-5) when the questions are paraphrases of each other, indicating a strong understanding of the topic and a clear connection between the questions. They also assign low scores (-4 to -5) when the questions are not paraphrases, indicating a lack of understanding or a different topic.

**Value Judgment Criteria:**
Annotator_1's explanations often reflect their personal opinions and biases. For example, in Sample 4, they mention that the two questions are "the same topic and subject," which implies a positive bias towards the topic.

**Evidence Usage:**
Annotator_1 rarely provides explicit evidence to support their answers. Instead, they rely on their understanding of the topic and make inferences based on the context.

**Explanation Style:**
Annotator_1's explanations are concise and to the point, often focusing on the main topic or subject. They use simple language and avoid complex explanations.

**Typical Wording:**
Annotator_1's typical wording is straightforward and lacks jargon. They use simple sentences and avoid technical terms unless necessary.

**Strictness:**
Annotator_1's annotations are generally strict, with a focus on identifying paraphrases and distinguishing them from non-paraphrases.

**Uncertainty:**
Annotator_1 rarely expresses uncertainty or doubt in their annotations. When they do, it's usually in the form of a low score (-4 to -5) indicating a lack of understanding or a different topic.

**Quirks:**
Annotator_1 has a few quirks in their annotation style:

* They often use similar wording in their explanations, such as "the same topic" or "the same subject."
* They tend to focus on the main topic or subject, rather than providing a comprehensive explanation.
* They rarely provide explicit evidence to support their answers.
* They have a bias towards positive language, using words like "good" and "best" to describe topics and subjects.

Overall, Annotator_1's annotation style is straightforward and focused on identifying paraphrases. They have a bias towards positive language and rarely express uncertainty or doubt in their annotations.
\end{promptbox}

\begin{promptbox}[R2 Llama3.2 Annotator 2 value profile]{PromptPurpleFrame}
**Label/Score Bias:**
The annotator tends to assign higher scores (closer to 5) to questions that are more concise, have a clear and specific query, and are related to a specific domain or entity. Conversely, questions that are more open-ended, vague, or unrelated to a specific domain are assigned lower scores.

**Value Judgment Criteria:**
The annotator's value judgment criteria seem to be based on the clarity and specificity of the query, as well as the domain or entity being queried. Questions that are more direct and concise are given higher scores, while questions that are more open-ended or vague are given lower scores.

**Evidence Usage:**
The annotator tends to rely on the context and the specific words used in the query to determine the relevance and specificity of the question. They also seem to consider the domain or entity being queried, as well as the presence of any qualifiers or modifiers.

**Explanation Style:**
The annotator's explanations are concise and to the point, often focusing on the key differences between the two questions or the specific criteria that led to the assigned score.

**Typical Wording:**
The annotator's typical wording is formal and objective, avoiding any emotive or subjective language. They tend to use technical terms and domain-specific jargon when relevant.

**Strictness:**
The annotator's strictness seems to be moderate, with a focus on clarity and specificity in the query. They are willing to consider context and domain-specific knowledge when assigning scores, but are not overly lenient with vague or open-ended questions.

**Uncertainty:**
The annotator tends to be uncertain when faced with questions that are highly ambiguous or unrelated to a specific domain. In such cases, they may assign a lower score or indicate that the question is not relevant to the task.

**Quirks:**
Some quirks in the annotator's behavior include:

* A tendency to overlook minor differences in wording or phrasing when the overall meaning and intent are clear.
* A focus on domain-specific knowledge and context when assigning scores.
* A willingness to consider technical terms and jargon when relevant.
* A moderate level of strictness, with a focus on clarity and specificity in the query.
* A tendency to be uncertain when faced with highly ambiguous or unrelated questions.

**Profile:**
Based on the provided examples, the annotator's profile can be summarized as follows:

* Label/Score Bias: Moderate, with a focus on clarity and specificity in the query.
* Value Judgment Criteria: Clarity and specificity of the query, as well as domain or entity being queried.
* Evidence Usage: Context and specific words used in the query, as well as domain or entity being queried.
* Explanation Style: Concise and objective, focusing on key differences between questions or specific criteria.
* Typical Wording: Formal and objective, using technical terms and domain-specific jargon when relevant.
* Strictness: Moderate, with a focus on clarity and specificity in the query.
* Uncertainty: Uncertain when faced with highly ambiguous or unrelated questions.
* Quirks: Tendency to overlook minor differences in wording or phrasing, focus on domain-specific knowledge and context, willingness to consider technical terms and jargon, moderate level of strictness, and tendency to be uncertain when faced with highly ambiguous or unrelated questions.

This profile can guide another LLM to imitate the annotator's behavior and assign scores based on the provided criteria.
\end{promptbox}

\begin{promptbox}[R2 Llama3.2 Annotator 3 value profile]{PromptPurpleFrame}
**Label/Score Bias:**
Annotator_3 tends to assign high scores (4-5) when the questions are paraphrases of each other, indicating a strong bias towards recognizing similar meaning. However, when the questions are not paraphrases (score -5), the annotator acknowledges the difference in meaning.

**Value Judgment Criteria:**
Annotator_3 does not explicitly express value judgments in their explanations. They focus on the literal meaning of the questions and provide concise explanations.

**Evidence Usage:**
Annotator_3 does not provide explicit evidence to support their answers. Their explanations are based on the literal meaning of the questions.

**Explanation Style:**
Annotator_3's explanations are concise and to the point, providing a clear indication of whether the questions are paraphrases or not.

**Typical Wording:**
Annotator_3's explanations often use simple and straightforward language, avoiding complex sentences or jargon.

**Strictness:**
Annotator_3 is relatively strict in their classification of paraphrases, assigning low scores (-5) when the questions are not similar in meaning.

**Uncertainty:**
Annotator_3 does not express uncertainty in their answers, providing a clear classification of paraphrases or not.

**Quirks:**

1. Annotator_3 tends to emphasize specific keywords or phrases in the questions, which can lead to different answers.
2. They do not provide explicit explanations for their answers, relying on the literal meaning of the questions.
3. Annotator_3's explanations are often brief, providing only a concise indication of whether the questions are paraphrases or not.
4. They do not acknowledge the context or nuances of the questions, focusing solely on the literal meaning.

**Profile:**
To imitate Annotator_3, follow these guidelines:

1. Focus on the literal meaning of the questions.
2. Emphasize specific keywords or phrases in the questions.
3. Provide concise explanations, avoiding complex sentences or jargon.
4. Classify questions as paraphrases or not based on their similarity in meaning.
5. Avoid expressing value judgments or providing explicit evidence.
6. Be strict in your classification of paraphrases, assigning low scores (-5) when the questions are not similar in meaning.
7. Do not express uncertainty in your answers.
8. Use simple and straightforward language in your explanations.
\end{promptbox}

\subsubsection{Qwen3 profiles}

These profiles correspond to profile model family: \texttt{qwen3}, sample size: 50, seed: 42.

\begin{promptbox}[R2 Qwen3 Annotator 0 value profile]{PromptOrangeFrame}
**Annotator_0 Profile (Imitation Guide)**

- **Label/Score Bias**:
  Strongly penalizes minor differences (e.g., word order, quantifiers like "some" vs. "a few") unless they alter core meaning. Assigns **-5** for completely unrelated topics, **-3 to -1** for subtle mismatches (e.g., scope, specificity, context), **0** only when ambiguity remains unresolved. **5** is given only when meaning is identical or trivially different (e.g., synonyms, rephrasing, minor syntax).

- **Value Judgment Criteria**:
  Prioritizes **semantic precision** over surface similarity. Rejects paraphrases that differ in:
  - Subject (e.g., "all questions" vs. "some questions")
  - Object (e.g., "steak" vs. "tomahawk steak")
  - Scope (e.g., country, time, field, entity)
  - Intent (e.g., definition vs. background, action vs. condition)
  - Detail (e.g., "why" vs. "what", "how" with or without reasons)

- **Evidence Usage**:
  Relies **exclusively on explicit content** in the questions. Does **not** infer meaning from context, assumptions, or external knowledge. Compares **literal content** and **core intent**. Ignores extraneous details (e.g., URLs, names, dates) unless they change the subject or scope.

- **Explanation Style**:
  Concise, direct, and **diagnostic**. Uses **clear comparative language** ("Q1 asks for X, Q2 asks for Y") to highlight differences. Avoids speculation. Uses **neutral, factual phrasing** without emotional tone.

- **Typical Wording**:
  - "Q1 asks for X. Q2 asks for Y."
  - "The words to be translated are different."
  - "Time is different."
  - "Two different questions."
  - "Q2 adds a reason..."
  - "Scope is different."
  - "The field of professorship is different."
  - "They could be different trailer."
  - "The issues on the webpage are different."

- **Strictness**:
  **Highly strict** on semantic fidelity. Even slight shifts in focus (e.g., "after 20" vs. "after 20 years") or entity (e.g., country, type of steak) lead to negative scores. Requires **exact alignment** in subject, object, and intent.

- **Uncertainty Handling**:
  Assigns **0 only when ambiguity is genuine** (e.g., unclear reference, missing context). Otherwise, defaults to **-1 or higher** if differences are detectable. Avoids neutral scores unless truly uncertain.

- **Quirks**:
  - **Sensitive to quantifiers and scope**: "some" vs. "all", "best" vs. "reliable" (assumes implied meaning).
  - **Rejects implied or contextual inferences** (e.g., "why" implies reasons -> not equivalent to "what").
  - **Finds differences in specificity** (e.g., "Dwarf DPS" vs. "Blood Elf DPS").
  - **Ignores
\end{promptbox}

\begin{promptbox}[R2 Qwen3 Annotator 1 value profile]{PromptOrangeFrame}
**Annotator_1 Profile (Imitation Guide)**

- **Label/Score Bias**:
  Tends to assign **high scores (4-5)** when questions share the **same core topic, object, and intent**, even with minor differences in phrasing or condition. Assigns **-4 or -5** when topics diverge, even if semantically related (e.g., same field but different focus, object, or condition). Avoids 0 unless ambiguity is clear.

- **Value Judgment Criteria**:
  Does **not** evaluate the correctness or quality of content. Focuses solely on **semantic alignment** and **topic coherence**. Does not judge tone, sentiment, or personal opinion.

- **Evidence Usage**:
  Uses **only explicit linguistic features** (topic, object, condition, subject, phrasing) to determine paraphrase status. Does **not** rely on external knowledge, context, or implied meaning.

- **Explanation Style**:
  Concise, direct, and **structured**. Always begins with a clear statement of the score, followed by a brief justification using phrases like:
  - "the same topic", "same topic and object", "similar topic but in different conditions", "different topics", "although refer to the same thing"
  - Avoids vague or emotional language.

- **Typical Wording**:
  - "the same topic and object of discussion"
  - "similar topic but in different conditions"
  - "different topics, although refer to the same thing"
  - "the same topic, subject and object..."
  - "different topics" (used as a strong negative signal)
  - "similar topic, although s2 is more specific" or "s1 refers to personal opinion"

- **Strictness**:
  **Highly strict** about topic boundaries. Even slight shifts in object (e.g., "maids" vs. "maids office"), condition (e.g., time, location), or focus (e.g., feeling vs. action) lead to negative scores.

- **Uncertainty**:
  Rarely assigns 0. When uncertainty exists, it is explicitly stated (e.g., "uncertain or mixed") only in edge cases, and even then, the explanation justifies the choice.

- **Quirks**:
  - **Sensitive to object specificity**: A shift in object (e.g., "Dwarf DPS" vs. "Blood Elf DPS") is treated as a condition change, not a topic shift.
  - **Ignores minor syntactic differences** (e.g., "what is" vs. "how is") if the core topic remains.
  - **Downgrades questions with added conditions or personal context** (e.g., "I'm worried about my boards") unless the core topic is identical.
  - **Fails to treat semantic overlap as sufficient**--e.g., "what is the best" vs. "what are the best" is not enough to justify a 5.
  - **Consistently penalizes location/time shifts** (e.g., August 2014 vs. September 2016) as "different conditions" rather than topic differences.

**Imitation Rules for New Items**:
1. Compare core topic, object, and intent.
2.
\end{promptbox}

\begin{promptbox}[R2 Qwen3 Annotator 2 value profile]{PromptOrangeFrame}
**Annotator_2 Profile (Imitation Guide)**

- **Label/Score Bias**:
  Tends to assign **high scores (4-5)** when questions share core domain, key phrase, and intent, even with minor differences (e.g., word order, redundancy, or minor qualifiers). **Low scores (-3 to -5)** only when concepts are unrelated or domains diverge significantly. **Score 0** is used for "unrelated" or "folklore" queries with shared style but no semantic overlap.

- **Value Judgment Criteria**:
  Prioritizes **semantic similarity over syntactic form**. Ignores trivial differences like capitalization, "some" vs. "the", or "you" vs. "people" if the core intent and domain match. Values **domain specificity** and **shared entity/attribute** (e.g., "mount kailash", "gaming PC", "Wuthering Heights") as primary anchors.

- **Evidence Usage**:
  Relies on **lexical overlap**, **shared domain**, and **core query structure**. Explicitly notes **omitted or redundant phrases** (e.g., "and strong", "second trailer") as negligible. Uses **contextual cues** (e.g., "context-rich", "colloquial") to assess similarity, especially when one query is more specific or structured.

- **Explanation Style**:
  Concise, direct, and **diagnostic**. Uses **structured reasoning** with clear justifications:
  - Identifies overlap (e.g., "common_phrase", "domain specific")
  - Notes differences (e.g., "penalty for different entities", "redundancy", "contextual richness")
  - Explicitly states what is **neglected** or **ignored**
  - Avoids emotional or speculative language

- **Typical Wording**:
  - "Overlap: X; penalty: Y"
  - "Q1 and Q2 share domain, but differ in Z"
  - "Redundancy can be neglected"
  - "Key term overlap: X"
  - "Conceptually unrelated"
  - "Colloquial, low signal"
  - "Query context differs"
  - "Omission of X can be neglected"

- **Strictness**:
  **Moderately strict** -- allows minor syntactic or lexical variation if semantic intent and domain match. **Strict** when domains or concepts differ (e.g., "gym" vs. "hacking", "cancer" vs. "asthma"). Rejects paraphrase claims when the **core object or concept** is different.

- **Uncertainty**:
  Uses **score 0** only when queries are **conceptually unrelated** or **stylistically similar but semantically distinct** (e.g., "folklore", "humorous", "context-rich"). Avoids vague or ambiguous explanations.

- **Quirks**:
  - **Ignores grammatical errors** as long as domain and intent align (e.g., "I like to learn rhetoric is their any institute" -> scored -3 due to conceptual mismatch, not grammar).
  - **Values colloquialism** as a signal of shared context (e.g., "like a maniac", "plz help").
  - **Pen
\end{promptbox}

\begin{promptbox}[R2 Qwen3 Annotator 3 value profile]{PromptOrangeFrame}
**Annotator_3 Profile (Imitation Guide)**

- **Label/Score Bias**:
  Tends to assign **high scores (4-5)** when questions are structurally and semantically similar, especially when core intent, topic, and phrasing align closely. **Low scores (-3 to -5)** when key differences in focus, subject, or implied context create distinct meanings. Avoids assigning 0 unless there's a clear, significant divergence in intent or scope.

- **Value Judgment Criteria**:
  Prioritizes **semantic precision** over surface similarity. Judges based on whether the questions target the same concept, object, or outcome. Values **specificity** (e.g., "Dwarf DPS" vs. "Blood Elf DPS") and **contextual nuance** (e.g., "UK" vs. "global", "in Mexico" vs. "in Germany") as decisive differentiators.

- **Evidence Usage**:
  Relies **exclusively on textual content** and **explicit differences in keywords, focus, or implied scope**. Does not use external knowledge or assumptions. Uses **contrastive analysis** (e.g., "emphasizes X", "focuses on Y") to justify scores.

- **Explanation Style**:
  Concise, direct, and **diagnostic**. Always includes a **specific observation** about what differs (e.g., "emphasizes X", "different keywords", "different time/place", "entirely different answers"). Avoids vague or interpretive language.

- **Typical Wording**:
  - "They have the same meaning."
  - "They have a similar meaning."
  - "They have almost the same meaning."
  - "Different keywords that are asked can lead to entirely different answers."
  - "The second question emphasises X."
  - "They do not mean the same thing, even though both are about Y."
  - "The focus of the questions is completely different."
  - "These two sentences express very different things."

- **Strictness**:
  **Highly strict**. Requires **near-identical semantic intent** for a score of 5. Minor differences in phrasing, focus, or context (e.g., "UK" vs. "global", "Dwarf" vs. "Blood Elf") are sufficient to reduce the score. Rejects overlap without semantic alignment.

- **Uncertainty**:
  Only assigns **0 or 1** when the difference is subtle but meaningful (e.g., "emphasises 'at home'") or when intent is ambiguous. Avoids uncertainty unless clearly warranted by divergent focus or context.

- **Quirks**:
  - **Sensitive to specific terms** (e.g., "Dwarf DPS", "Tomahawk steak", "apple cider vinegar" vs. "apple cider").
  - **Highlights contextual differences** (time, location, object, scope) as primary reasons for scoring.
  - **Consistently uses "emphasises" or "focuses on"** to describe differences.
  - **Avoids overgeneralization**; never assumes similarity without explicit textual alignment.
  - **Scores -5 only when topics are fundamentally unrelated** (e.g., hacking vs. business names, health vs. history).

**Imitation Rule Summary**:
When two
\end{promptbox}

\end{document}